\documentclass{package/IEEEtran}

\usepackage{graphics} \usepackage{amsmath} \usepackage{mathtools} \usepackage{amssymb}  \usepackage[utf8]{inputenc} 
\usepackage{helvet} \usepackage{courier} \usepackage{type1cm} \usepackage{makeidx} \usepackage{multicol} \usepackage{multirow} \usepackage{mleftright}
\usepackage{colortbl}

\usepackage{algorithm}
\usepackage{amsmath,amssymb,amsfonts,amsthm}
\usepackage{graphicx}
\graphicspath{{figures/}}
\usepackage{epsfig}
\usepackage{algorithm}
\usepackage{color} \usepackage{multirow}
\usepackage{url}
\usepackage{bm} \usepackage{float} \usepackage{tikz}
\usetikzlibrary{positioning}
\usetikzlibrary{shapes,snakes}
\usepackage{empheq} \usepackage{xcolor} \usepackage{arydshln} \usepackage{longtable} 
\usepackage{caption} \usepackage{subcaption} \usepackage{fancyhdr}

\usepackage{hyperref} 
\newcommand\scalemath[2]{\scalebox{#1}{\mbox{\ensuremath{\displaystyle #2}}}}

\newcommand{\ws}[1]{\cellcolor{green!20}\textbf{#1}}
\newcommand{\wi}[1]{\cellcolor{green!20}\textbf{#1}}
\newcommand{\wj}[1]{\cellcolor{green!20}\textbf{#1}}

\usepackage{etoolbox}
\makeatletter
\patchcmd\@combinedblfloats{\box\@outputbox}{\unvbox\@outputbox}{}{   \errmessage{\noexpand\@combinedblfloats could not be patched}}\makeatother

\definecolor{lightgray}{gray}{0.9}

\usepackage{natbib}
\begin{document}

\title{Closed-form Preintegration Methods for Graph-based Visual-Inertial Navigation}

\author{Kevin Eckenhoff, Patrick Geneva, and Guoquan Huang\thanks{The authors are with the Dept. of Mechanical Engineering, and Computer and Information Sciences, University of Delaware, Newark, DE 19716, USA.  E-mail:{\tt\small \{keck, pgeneva, ghuang\}@udel.edu}}}

\maketitle
\thispagestyle{empty}
\pagestyle{empty}

\begin{abstract}
In this paper we propose a new analytical preintegration theory for graph-based sensor fusion  with an inertial measurement unit (IMU) and a camera (or other aiding sensors).
Rather than using discrete sampling of the measurement dynamics as in current methods, 
we derive the closed-form solutions to the preintegration equations, yielding improved accuracy in state estimation.
We advocate two new different inertial models for preintegration: (i) the model that assumes piecewise constant measurements, 
and (ii) the model that assumes piecewise constant local true acceleration.
We show through extensive Monte-Carlo simulations the effect that the choice of preintegration model has on estimation performance.
To validate the proposed preintegration theory, we develop both direct and indirect visual-inertial navigation systems (VINS) that leverage our preintegration.
In the first, within a tightly-coupled, sliding-window optimization framework, we jointly estimate the features in the window and the IMU states while performing marginalization to bound the computational cost.
In the second, we loosely-couple the IMU preintegration with a direct image alignment that estimates relative camera motion by minimizing the photometric errors (i.e., image intensity difference), allowing for efficient and informative loop closures. 
Both systems are extensively validated in real-world experiments and are shown to offer competitive performance to state-of-the-art methods.  
\end{abstract}

 \section{Introduction}

Accurate localization for autonomous systems is a prerequisite in many robotic applications such as planetary exploration \citep*{Mourikis2007RSS}, search and rescue \citep*{Ellekilde2007JFR}, and autonomous driving \citep*{Geiger2012CVPR}. In many of these scenarios, access to global information such as from a Global Positioning System (GPS), motion capture system, or a prior map of the environment is unavailable. Instead, one has to estimate the robot state  and its surroundings based on noisy, local measurements from onboard sensors, by performing simultaneous localization and mapping (SLAM),
which has witnessed significant research efforts in the past three decades~\citep*{Cadena2016TRO}.

Of many possible sensors used in SLAM, micro-electro-mechanical-system (MEMS) inertial measurement units (IMUs) have become ubiquitous. These low-cost and light-weight sensors typically provide local linear acceleration and angular velocity readings, and are well suited for many applications such as micro aerial vehicles (MAVs) \citep*{Ling2016ICRA} and mobile devices \citep*{Wu2015RSS}. IMUs provide information only about the derivatives of the kinematic states, so estimation must be performed by integrating over these noisy measurements. This may lead to large drifts over long periods of time, making the use of a low-cost IMU alone an unreliable solution. However, IMU readings are highly-informative about short-term motion which is ideal for fusion with measurements from {exteroceptive} aiding sensors, such as LiDAR and cameras. These sensors compensate for the drift issue inherent in inertial navigation, while high-rate inertial measurements are useful in tracking aggressive motion which may be difficult for exteroceptive low-rate sensors alone. 

One canonical way of fusing IMU measurements in aided inertial navigation is to use an extended Kalman filter (EKF) (e.g., see \citet*{Mourikis2007ICRA}).
In this method, the inertial measurements are used to predict to the next time instance, while measurements from exteroceptive sensors are used to update the state estimate.
More recently, the development of preintegration has allowed for the efficient inclusion of high-rate IMU measurements in graph-based SLAM
\citep*{Lupton2012TOR,Forster2015RSS,Forster2017TRO}.
In this paper, building upon our prior conference publication~\citep*{Eckenhoff2016WAFR}, 
we investigate in-depth the optimal use of preintegration by providing models and their closed-form solutions for the preintegrated measurement dynamics,
allowing for more accurate computation of the inertial factors for use in graph optimization of visual-inertial navigation systems (VINS).

In particular, the main contributions of this work include:
\begin{itemize}
\item We advocate two new preintegration models (i.e., piecewise constant measurements and piecewise constant local true acceleration, instead of piecewise constant global acceleration as assumed in existing methods) 
to better capture the underlying motion dynamics and offer the analytical solutions to the preintegration equations. 
We have open sourced the proposed preintegration to better contribute to our research community.\footnote{The open source of the proposed closed-form preintegration is available at: \url{https://github.com/rpng/cpi}}
\item Using the proposed closed-form preintegration, we develop an indirect, tightly-coupled, sliding-window optimization based visual-inertial odometry (VIO), which marginalizes out features from the state vector when moving to the next time window to enable real-time performance of bounded computational cost.
\item With the proposed closed-form IMU preintegration, we further develop a loosely-coupled, direct VINS, which fuses preintegrated inertial measurements with direct image alignment results. 
\item We conduct thorough Monte-Carlo simulation analysis of different preintegration models by varying motion dynamics and IMU sampling rates.
We also perform extensive real-world experiments to validate the proposed VINS using our preintegration by comparing with a state-of-the-art method.
\end{itemize}

The reminder of the paper is organized as follows: 
After a brief overview of related work in the next section and estimation preliminaries in Section \ref{sec:prelim}, 
we present in detail the proposed continuous preintegration in Section \ref{sec:continuous-preint}. 
The direct and indirect VINS that use the proposed preintegration are described in Section \ref{sec:vins}.
In Sections \ref{sec:sim} and \ref{sec:exp1}, we validate the proposed VINS algorithms through both simulations and experiments. 
Finally, Section \ref{sec:concl} concludes the work in this paper, as well as the possible future research directions.

\section{Related Work}

\subsection{Visual-Inertial Navigation}

\cite*{Mourikis2007ICRA} proposed one of the earliest successful VINS algorithms, known as the multi-state constraint Kalman filter (MSCKF).
This filtering approach used quaternion-based inertial dynamics \citep*{Trawny2005_Q_TR} for state propagation coupled with a novel EKF update step.
Rather than adding features seen in the camera images to the state vector, their visual measurements were projected onto the nullspace of the feature Jacobian matrix (akin to feature marginalization~\citep*{Yang2017IROS}), thereby retaining motion constraints that only related to the stochastically cloned camera poses in the state vector \citep*{Roumeliotis2002ICRAa}.
While reducing the computational cost by removing the need to co-estimate features, 
this nullspace projection prevents the relinearization of the processed features' nonlinear measurements at later time steps.

The standard MSCKF recently has been extended in various directions.
For example, 
\citet*{Hesch2013TRO,Huang2014ICRA} improved the filter consistency by enforcing the correct observability properties of the linearized EKF VINS.
\cite*{Guo2013iros} showed that the inclusion of plane features increases the estimation accuracy.
\cite*{Guo2014RSS} extended to the case of rolling-shutter cameras with inaccurate time synchronization.
Recently, \cite*{Wu2015RSS} further reformulated the VINS problem within a square-root inverse filtering framework for improved computational efficiency and numerical stability without sacrificing estimation accuracy.
While these MSCKF-based methods have shown to exhibit accurate state estimation, they theoretically suffer from a limitation -- that is, nonlinear measurements must have a \textit{one-time} linearization before processing, possibly introducing large linearization errors into the estimator.

Batch optimization methods, by contrast, solve a nonlinear least-squares or bundle adjustment (BA) problem over a set of measurements, allowing for the reduction of error through relinearization \citep*{Kummerle2011ICRA}.
The incorporation of \textit{tightly-coupled} VINS in batch optimization methods requires overcoming the high frequency nature and computational complexity of the inertial measurements.

\cite*{LeuteneggerIJRR2014} introduced a keyframe-based VINS approach (i.e., OKVIS), whereby a set of non-sequential past camera poses and a series of recent inertial states, connected with inertial measurements, was used in nonlinear optimization for accurate trajectory estimation.
These inertial factors took the form of a state prediction: every time that the linearization point for the starting inertial state threshold, it is required to reintegrate the IMU dynamics.
This presents inefficiencies in the inertial processing, while the authors demonstrated the feasibility of such a scheme for a small number of inertial factors in a sliding window estimator. 
It should be noted that the well-known open-source implementation of OKVIS\footnote{\url{https://github.com/ethz-asl/okvis}} 
in fact employs the method of inertial  preintegration, described in detail later, while only triggering full reintegration if the linearization point changes sufficiently and thus improving the efficiency.

\subsection{Visual Processing}

A key component to any VINS algorithm is the visual processing pipeline,
responsible for transforming dense imagery data to motion constraints 
that can be incorporated into the estimation problem.
Seen as the classical technique, indirect methods of visual SLAM extract and track features in the environment, while using geometric reprojection constraints during estimation.
An example of state-of-the-art indirect visual-SLAM methods is ORB-SLAM2 \citep*{Mur2017TRO}, which performs graph-based optimization of camera poses using information from 3D feature point correspondences.

In contrast, direct methods utilize  pixel intensities in their formulation and allow for inclusion of a larger percentage of the available image information.
LSD-SLAM is an example of  state-of-the-art direct visual-SLAM methods which optimizes the transformation between pairs of camera keyframes based on minimizing their intensity error \citep*{Engel2014ECCV}.
Note that this approach also optimizes a separate graph containing keyframe constraints to allow for the incorporation of highly informative loop-closures to correct drift over long trajectories.
This work was later extended from a monocular sensor to stereo and omnidirectional cameras for improved accuracy \citep*{Engel2015IROS,Caruso2015IROS}.
Other popular direct methods include the work by \cite*{Engel2018TPAMI} and \cite*{Wang2017ICCV} which estimated keyframe depths along with the camera poses in a tightly-coupled manner, offering low-drift performance.

Application of direct methods to the visual-inertial problem has seen recent attention due to their ability to robustly track dynamic motion even in low-texture environments.
For example,
\cite*{Bloesch2015IROS,Bloesch2017IJRR} used a patch-based direct method to provide updates with an iterated EKF;
\cite*{Usenko2016ICRA} introduced a sliding-window VINS based on the discrete preintegration and direct image alignment;
\cite*{Ling2016ICRA} employed loosely-coupled direct alignment with preintegration factors for tracking aggressive quadrotor motions.
While these methods have shown the feasibility of incorporating IMU measurements with direct methods, they employed the \textit{discrete} form of inertial preintegration.

\subsection{Inertial Preintegration}

First introduced by \cite*{Lupton2012TOR}, inertial preintegration is a computationally efficient alternative to the standard inertial measurement integration, 
e.g., as performed in EKF propagation.
The authors employed the discrete integration of the inertial measurement dynamics in a \textit{local} frame of reference, preventing the need to reintegrate the state dynamics at each optimization step.
While this addresses the computational complexity issue, this method suffers from singularities due to the use of Euler angles in the orientation representation.
To improve the stability of this preintegration, an on-manifold representation was introduced by \cite*{Forster2015RSS,Forster2017TRO} 
which presents a singularity-free orientation representation on the $SO(3)$ manifold, incorporating the IMU preintegration into an efficient graph-based VINS algorithm.

While \cite*{Shen2015ICRA} introduced preintegration in the continuous form, they still discretely sampled the measurement dynamics without offering closed-form solutions.
This left a significant gap in the theoretical completeness of preintegration theory from a continuous-time perspective.
Albeit, \citet*{Qin2017VINSMonoAR} later extended to a robust tightly-coupled monocular visual-inertial localization system.
As compared to the discrete approximation of the preintegrated measurement and covariance calculations used in previous methods, 
in our prior work \citep*{Eckenhoff2016WAFR}, we have derived the closed-form solutions to both the  measurement and covariance preintegration equations
and showed that these solutions offer improved accuracy over the discrete methods, especially in the case of highly dynamic motion.

In this work, based on our preliminary results \citep*{Eckenhoff2016WAFR,Eckenhoff2017ICRA}, 
we provide a solid theoretical foundation for closed-form preintegration and show that it can be easily incorporated into different graph-based sensor fusion methods.
We investigate the improved accuracy afforded by two different models of closed-form preintegration and scenarios in which they exhibit superior performance.
We further develop both indirect and direct graph-based VINS and demonstrate their competitive performance to state-of-the-art methods.
 \section{Estimation Preliminaries}
\label{sec:prelim}

The IMU state of an aided inertial navigation system at time step $k$ is given by \citep*{Mourikis2007ICRA}:
\begin{align}
\mathbf{x}_k = \begin{bmatrix} {}_G^k \bar{q}{}^{\top} & \mathbf{b}_{\omega_k}^{\top} & {}^G\mathbf{v}_{k}^{\top} & \mathbf{b}_{a_k}^{\top} & {}^G\mathbf{p}_{k}^{\top}
\end{bmatrix}^{\top}
\label{eq:state}
\end{align}
where ${}_G^k \bar{q}$ is the unit quaternion of JPL form parameterizing the rotation ${}_G^k \mathbf{R}$ from the global frame $\{G\}$ to the current local frame $\{k\}$ \citep*{Trawny2005_Q_TR}, $\mathbf{b}_{\omega_ k}$ and $\mathbf{b}_{a_k}$ are the gyroscope and accelerometer biases, and ${}^G \mathbf{v}_{k}$ and  ${}^G \mathbf{p}_{k}$ are the velocity and position of the IMU expressed in the global frame, respectively.

\begin{figure} \centering
\includegraphics[height=2cm]{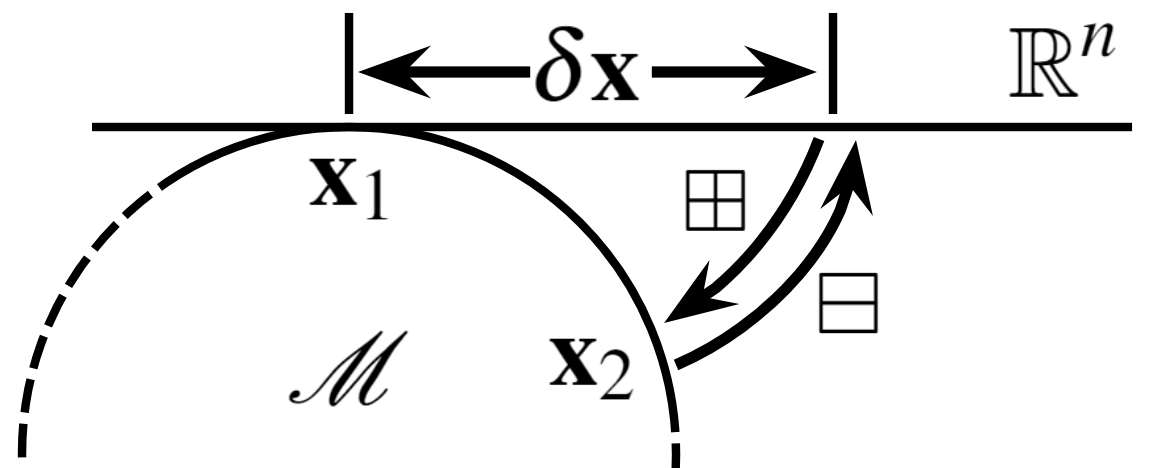}
\caption{Illustration of the state update operations on a manifold.
The $\boxplus$ operation maps $\mathbf{x}_1 \in \mathcal M$ and a vector $\delta \mathbf{x} \in \mathbb R^n$ to a new element $\mathbf{x}_2 \in \mathcal M$, while the $\boxminus$ operation maps $\mathbf{x}_1$ and $\mathbf{x}_2$ to the vector $\delta \mathbf{x}$.
}
\label{fig:manifolddelta}
\end{figure}

Note that while the state vector~\eqref{eq:state} contains 16 variables, there are only 15 degrees of freedom (DOF), due to the constraint that the quaternion ${}_G^k \bar{q}$ must have unit length.
In fact, the state lies on the manifold defined by the product of the unit quaternions $\mathbb H$ with the vector space $\mathbb{R}^{12}$ (i.e., $\mathcal M = {\mathbb{H}} \times \mathbb{R}^{12}$).
In order to represent the estimation problem on manifold, we employ the ``boxplus'' update operation, $\boxplus$, which maps an element from a manifold, $\mathbf{x} \in \mathcal{M}$, and an error vector $\delta \mathbf{x}$ into a new element on  $\mathcal{M}$ \citep*{Hertzberg2011}.
As illustrated in Figure~\ref{fig:manifolddelta}, for a manifold of dimension $n$, we can define the following operation:
\begin{align}
    \boxplus :~ \mathcal{M} \times \mathbb{R}^n &\rightarrow \mathcal{M} \\ 
        \mathbf{x}_1 \boxplus \delta \mathbf{x} &= \mathbf{x}_2
\end{align}
Similarly, the inverse ``boxminus'' operation $\boxminus$ is given by:
\begin{align}
    \boxminus :~ \mathcal{M} \times \mathcal{M} &\rightarrow \mathbb{R}^n \\
    \mathbf{x}_2 \boxminus \mathbf{x}_1 &=   \delta \mathbf{x}
\end{align}
In the case of a state in a vector space, $\mathbf{v} \in \mathbb R^n$, these operations are the standard addition and subtraction:
\begin{alignat}{2}
    \mathbf{v}_1 \boxplus \delta \mathbf{v} &\triangleq \mathbf{v}_1 + \delta \mathbf{v} &&= \mathbf{v}_2 \\
     \mathbf{v}_2 \boxminus \mathbf{v}_1 &\triangleq\mathbf{v}_2- \mathbf{v}_1 &&= \delta \mathbf{v}
\end{alignat}
In the case of a unit quaternion expressed using the JPL convention, $\bar{q}$, we have \citep*{Trawny2005_Q_TR}:
\begin{align}
    \bar{q}_1 \boxplus \delta \bm \theta &\overset{\Delta}{=} \begin{bmatrix} \frac{ \delta \bm \theta}{2} \\ 1 \end{bmatrix} \otimes \bar{q}_1 \simeq \bar{q}_2 \\
    \bar{q}_2 \boxminus \bar{q}_1 &\triangleq 2\mathbf{vec}\left(\bar{q}_2 \otimes \bar{q}_1^{-1}\right)= \delta \bm \theta
\end{align}
where $\mathbf{vec}\left(\bar{q} \right)$ refers to the vector portion of the quaternion argument (i.e., $\mathbf{vec}([ \mathbf{q}^{\top} q_4 ]^{\top})=  \mathbf{q}$).
The quaternion multiplication, $\otimes$, is given by:
\begin{align}
\bar{q} \otimes \bar{p} &\triangleq \mathcal{R}\left(\bar{p}\right)\bar{q} = \mathcal{L}\left(\bar{q}\right)\bar{p}\\
\mathcal{R}\left(\bar{q}\right) &= \begin{bmatrix} q_4\mathbf{I} +\lfloor\mathbf{q}\rfloor & \mathbf{q} \\ -\mathbf{q}^{\top} & q_4 \end{bmatrix} \\
\mathcal{L}\left(\bar{p}\right) & = \begin{bmatrix} p_4\mathbf{I} -\lfloor\mathbf{p}\rfloor & \mathbf{p} \\ -\mathbf{p}^{\top} & p_4 \end{bmatrix}
\end{align}
where for $\mathbf{q}= [q_x ~ q_y ~ q_z ]^{\top}$:
\begin{align}
\lfloor \mathbf{q} \rfloor= \begin{bmatrix} 0 & -q_z & q_y \\ q_z & 0 & -q_x \\ -q_y & q_x & 0 \end{bmatrix}
\end{align}

In state estimation, these operations allow us to model the state on manifold using a Gaussian distribution on its \textit{error state} vector.
In particular, the random variable $\mathbf{x}$ with mean value $\hat{\mathbf{x}}$ takes the form:
\begin{align}
    \mathbf{x} &= \hat{\mathbf{x}} \boxplus \delta \mathbf{x} \\
    \delta \mathbf{x} &\sim \mathcal{N} \left(\mathbf{0}, \bm \Sigma \right)
\end{align}
where $ \bm \Sigma$ is the covariance of the zero-mean error state.
The error state corresponding to \eqref{eq:state} is thus given by:
\begin{align}
\delta \mathbf{x}_k = \begin{bmatrix} {}^k\delta \bm \theta_G^{\top} & \delta\mathbf{b}_{\omega_k}^{\top} & {}^G\delta\mathbf{v}_{k}^{\top} & \delta\mathbf{b}_{a_k}^{\top} & {}^G\delta\mathbf{p}_{k}^{\top}
\end{bmatrix}^{\top}
\label{eq:stateerror}
\end{align}

\subsection{Batch Optimization}

In the case of graph SLAM~\citep*{Grisetti2010ITSM}, the graph nodes can correspond to historical robot states and features in the environment, while the edges represent collected measurements from sensors which relate the incident nodes. As an example, a robot measuring a feature would add an edge between the feature and the robot state node.
Using this graph formulation and under the assumption of independent zero-mean Gaussian noise, we can find a maximum a posteriori (MAP) estimate of all states by solving the following nonlinear least-squares problem \citep*{Kummerle2011ICRA}:
\begin{align}
\hat{\mathbf{x}} = \mathop{\mathrm{argmin}}_{\mathbf{x}} \sum_i \frac{1}{2}\left| \left| \mathbf{e}_i \left(\mathbf{x} \right) \right| \right|_{\bm \Lambda_i}^2 \label{eq::MLE}
\end{align}
where $\mathbf{e}_i$ is the error/residual of the $i$-th measurement, $\bm \Lambda_i$ is the associated information matrix (inverse covariance), and $\left| \left | \mathbf{v} \right| \right|_{\bm \Lambda}^2 =  \mathbf{v}^{\top}{\bm \Lambda}\mathbf{v}$ represents the squared energy norm.
Note that as a common practice, a (Huber or Cauchy) robust cost function of Equation~\eqref{eq::MLE} is often used to compensate for  outliers,  in particular when fusing visual measurements~\citep{Hartley2000}.
Optimization is typically performed iteratively, e.g.,  through a Gauss-Newton or Levenberg–Marquard  method, 
by linearizing the nonlinear measurements about the current estimate, $\hat{\mathbf{x}}$, and defining a new weighted linear least squares problem in terms of the error state $\delta\mathbf{x}$:
\begin{align}
\delta\hat{\mathbf{x}} &= \mathop{\mathrm{argmin}}_{\delta \mathbf{x}} \sum_i \frac{1}{2}\left| \left| \mathbf{e}_i \left(\hat{\mathbf{x}}\right) + \mathbf{J}_i \delta \mathbf{x} \right| \right|_{\bm \Lambda_i}^2 \label{eq:linearizedwls}\\
\mathbf{J}_i &= \frac{\partial \mathbf{e}_i \left(\hat{\mathbf{x}}\boxplus \delta \mathbf{x} \right)}{\partial \delta \mathbf{x}} \Big|_{\delta \mathbf{x}= \mathbf{0}}
\label{eq:linearizedjacobian}
\end{align}
We can see that the original optimization problem has been converted into finding the optimal \textit{correction} vector, $\delta\mathbf{x}$, to the current state estimate. 
The optimal solution can be found by solving the following normal equation:
\begin{align}
\left(  \sum_i \mathbf{J}_i^{\top} {\bm \Lambda_i} \mathbf{J}_i \right)\delta\hat{\mathbf{x}} &= - \sum_i \mathbf{J}_i^{\top}{\bm \Lambda_i} \mathbf{e}_i\left(\hat{\mathbf{x}} \right) \\
\Longleftrightarrow~  \bm\Lambda \delta\hat{\mathbf{x}} &= -\mathbf g
\label{eq:lin_sys}
\end{align}
After obtaining the optimal correction, $\delta\hat{\mathbf{x}}$, 
we update our current estimate at the $k$-th iteration as: $\hat{\mathbf{x}}^{(k+1)}= \hat{\mathbf{x}}^{(k)} \boxplus \delta\hat{\mathbf{x}}$,
and repeat the optimization process.
After convergence, we will be left with the following distribution:
\begin{align}
\mathbf{x} &= \hat{\mathbf{x}} \boxplus \delta \mathbf{x} \\
\delta \mathbf{x} &\sim \mathcal{N} \left(\mathbf{0}, \bm \Sigma \right) \\
\bm \Sigma &= \left(  \sum_i \mathbf{J}_i^{\top} {\bm \Lambda_i} \mathbf{J}_i \right)^{-1}
\end{align}
where the measurement Jacobians, $\mathbf{J}_i$, are evaluated at the final state estimate.

\subsection{Marginalization}

\begin{figure*}[ht]
\centering
\includegraphics[height=4cm]{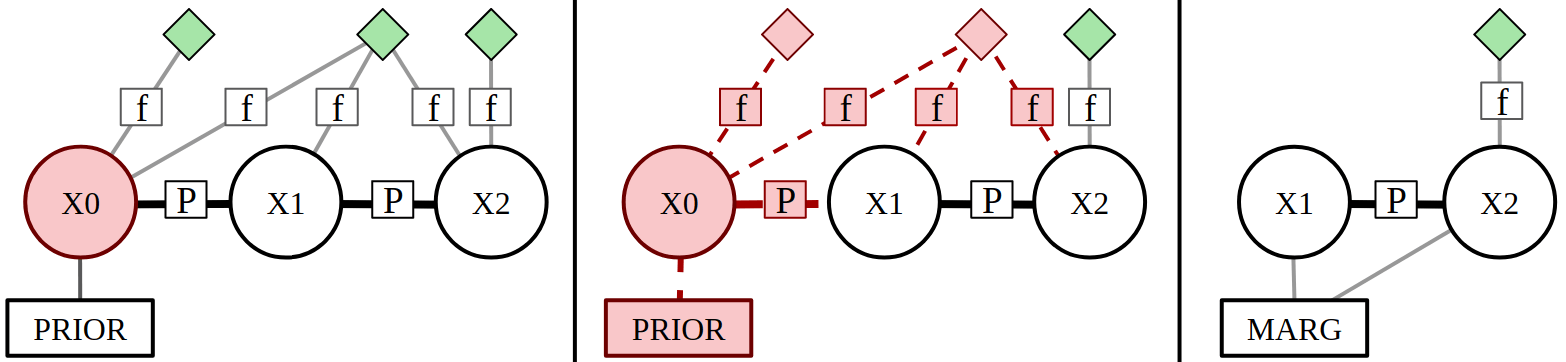}
\caption{During graph optimization of VINS, IMU states (shown in circles) and 3D features (diamonds) are included in the graph.
Image projection measurements connect features and the IMU state corresponding to the time that the image was recorded.
Subsequent IMU states are connected with preintegrated factors, while a prior factor connects to the oldest IMU state.
During marginalization, we first select the states to be marginalized, e.g., the oldest IMU state in the window and its associated features (in red). 
With these measurements we perform marginalization to form a new marginal measurement for future optimization.}
\label{fig:marginalization}
\end{figure*}

In a naive graph SLAM formulation, nodes are continuously added to the graph as time progresses without consideration to the computational burden.
For example, as a robot moves through an unknown environment we would add robot state nodes at \textit{every} measurement time.
This becomes a problem due to the high computational complexity, $O(n^3)$ with $n=\dim(\mathbf x)$, of batch optimization, in the worst case.
In order to bound the computational complexity of the system, marginalization is often performed to remove a set of nodes, called marginalized states, from the graph, while retaining the information contained in their incident edges (see Figure~\ref{fig:marginalization} for an example)~\citep*{Huang2013ECMRa,Eckenhoff2016IROS}.
Partitioning the optimization variables into states remaining after marginalization, $\mathbf{x}_r$, and the to-be marginalized states, $\mathbf{x}_m$, we can write  \eqref{eq::MLE} as the solution of the following minimization \citep*{Huang2011IROS}:
\begin{align}
    \{\hat{\mathbf{x}}_r, \hat{\mathbf{x}}_m\} = \mathop{\mathrm{argmin}}_{\mathbf{x}_r, \mathbf{x}_m} \Big(c_r(\mathbf{x}_r)+ c_m(\mathbf{x}_m, \mathbf{x}_r)  \Big)
\end{align}
The second subcost,  $c_m(\mathbf{x}_m, \mathbf{x}_r)$, is associated with the measurements incident to the marginalized states, and is a function of both these states and the remaining ones.
The first, $c_r(\mathbf{x}_r)$, refers to all other edges in the graph. The optimal estimate for the remaining nodes can be written as:
\begin{align}
    \hat{\mathbf{x}}_r = \mathop{\mathrm{argmin}}_{\mathbf{x}_r} \Big(   c_r(\mathbf{x}_r) + \mathop{\mathrm{min}}_{\mathbf{x}_m} c_m(\mathbf{x}_m, \mathbf{x}_r)  \Big)
\end{align}
That is, minimizing $c_m(\mathbf{x}_m, \mathbf{x}_r)$ with respect to $\mathbf{x}_m$ yields a cost that is a function \textit{only} of the remaining states.
This minimization is performed as in  \eqref{eq:lin_sys}, where we write out the linear system for only the measurements involved in $c_m$:
\begin{align}
\begin{bmatrix} {\bm \Lambda_{rr}} & {\bm \Lambda_{rm}} \\ {\bm \Lambda_{mr}} & {\bm \Lambda_{mm}} \end{bmatrix} \begin{bmatrix} \delta \mathbf{x}_r \\ \delta \mathbf{x}_m \end{bmatrix} = \begin{bmatrix} -\mathbf{g}_r \\ -\mathbf{g}_m \end{bmatrix}
\end{align} 
The optimal subcost $c_m$, up to an irrelevant constant, is given by \citep*{Nerurkar2014ICRA}:\footnote{Throughout the paper, we reserve the symbol $\hat{x}$ to denote the current estimate of state variable $x$ in optimization, while $\breve{x}$ refers to the (inferred) measurement mean value.}
\begin{align} \label{eq:cmarg}
\scalemath{0.95}{
c_{marg}\left(\mathbf{x}_r\right) =  \frac{1}{2}\left| \left|\mathbf{x}_r\boxminus \breve{\mathbf{x}}_r \right| \right|_{\bm \Lambda_{marg}}^2+ \mathbf{g}_{marg}^{\top}\left(\mathbf{x}_r\boxminus \breve{\mathbf{x}}_r\right) 
}
\end{align}
where $\breve{\mathbf{x}}_r$  is the linearization point used to build the system (in practice, the current state estimate at the time of marginalization), and ${\bm \Lambda}_{marg}= {\bm \Lambda}_{rr}- {\bm \Lambda}_{rm}{\bm \Lambda}_{mm}^{-1}{\bm \Lambda}_{mr}$ and ${\mathbf{g}}_{marg}= {\mathbf{g}}_{r}- {\bm \Lambda}_{rm}{\bm \Lambda}_{mm}^{-1}{\mathbf{ g}}_{m}$ are the marginalized Hessian and gradient, respectively.

In future optimization, this marginalization creates both a new quadratic \textit{and} linear cost in terms of the error between the remaining states and their linearization points.
This then replaces the marginalized measurements in the original graph, and we can write this new cost~\eqref{eq:cmarg} up to a constant in the form of  \eqref{eq::MLE}:
\begin{align}
    c_{marg}\left(\mathbf{x}_r\right) &= \frac{1}{2}\left|\left| \mathbf{A}_m\left(\mathbf{x}_r\boxminus \breve{\mathbf{x}}_r\right)+\mathbf{b}_m \right|\right|_{2}^2 \\ 
{\rm with}~    \mathbf{A}_m^{\top}\mathbf{A}_m &= \bm \Lambda_{marg} \\
    \mathbf{A}_m^{\top}\mathbf{b}_m &= \mathbf{g}_{marg}
\end{align}
This cost yields the following residual and Jacobian for use in optimization (see  \eqref{eq:linearizedwls} and \eqref{eq:linearizedjacobian}):
\begin{align}
\mathbf{e}_{marg}(\hat{\mathbf{x}}) &= \mathbf{A}_m\left(\hat{\mathbf{x}}_r\boxminus \breve{\mathbf{x}}_r\right)+\mathbf{b}_m \label{eq:margresidual}\\
\mathbf{J}_{marg} &= \mathbf{A}_m \frac{\partial\left( \left(\hat{\mathbf{x}}_r
\boxplus \delta \mathbf{x}_r \right) \boxminus \breve{\mathbf{x}}_r\right)}{\partial \delta \mathbf{x}_r} \Big|_{\delta \mathbf{x}_r = \mathbf{0} }
\end{align}
where for the Jacobian of a vector (i.e., if $\mathbf x_r =\mathbf{v}$):
\begin{align}
    \frac{\partial \left(\left(\hat{\mathbf{v}}
\boxplus \delta \mathbf{v} \right) \boxminus \breve{\mathbf{v}}\right)}{\partial \delta \mathbf{v}} &= \frac{\partial \left(\hat{\mathbf{v}}
+ \delta \mathbf{v}  - \breve{\mathbf{v}}\right)}{\partial \delta \mathbf{v}} = \mathbf{I}
\end{align}
and for a quaternion $\bar{q}$, with $\tilde{\bar{q}} = \hat{\bar{q}} \otimes \breve{\bar{q}}^{-1}$, we have:
\begin{align}
 \frac{\partial \left(\left(\hat{\bar{q}}
\boxplus \delta \bm \theta \right) \boxminus \breve{\bar{q}}\right)}{\partial \delta \bm \theta} &=  \frac{\partial 2\mathbf{vec}\left( \begin{bmatrix} \frac{\delta \bm \theta}{2} \\ 1 \end{bmatrix} \otimes \hat{\bar{q}}
  \otimes \breve{\bar{q}}^{-1}\right) }{\partial \delta \bm \theta} \notag\\
  &= \frac{\partial 2\mathbf{vec}\left(  \mathcal{R}\left(\tilde{\bar{q}} \right) \begin{bmatrix}\frac{\delta \bm \theta}{2} \\ 1 \end{bmatrix} \right) }{\partial \delta \bm \theta} \notag \\
  &= \tilde{{q}}_4\mathbf{I} + \lfloor \tilde{\mathbf{q}} \rfloor 
\end{align}

 \section{Closed-form Preintegration} \label{sec:continuous-preint}

In this section, we present in detail the proposed closed-form IMU preintegration based on two different realistic inertial models, 
which is expected to be readily used in any graph-based aided inertial navigation, 
thus providing an essential building block for visual-inertial state estimation.

An IMU attached to the robot collects inertial readings of the underlying state dynamics.
In particular, the sensor receives angular velocity $\bm \omega_m$ and local linear acceleration $\mathbf{a}_m$ measurements which relate to the corresponding true values $\bm \omega$ and $\mathbf{a}$ as follows:
\begin{align}
    \bm \omega_m= \bm \omega+\mathbf{b}_{\omega}+\mathbf{n}_{\omega} \\
    \mathbf{a}_m= \mathbf{a}+{}_G^I\mathbf{R}{}^G\mathbf{g}+ \mathbf{b}_{a}+\mathbf{n}_{a}
\end{align}
where ${}^G\mathbf{g} = [ 0 ~ 0 ~ 9.81 ]^{\top}$ is the global gravity\footnote{Note that gravity is slightly different in different parts of the globe.} 
and ${}_G^I\mathbf{R}$ is the rotation from the global frame to the instantaneous local inertial frame.
The measurements are corrupted both by the time-varying biases $\mathbf b_\omega$ and $\mathbf b_a$ (which must be co-estimated with the state), and the zero-mean white Gaussian noises $\mathbf{n}_{\omega}$ and $\mathbf{n}_{a}$.
The standard dynamics of the IMU state is given by \citep*{Chatfield1997}:
\begin{align}
{^I_G \dot{\bar q}} &= \frac{1}{2} \bm\Omega(\bm\omega_m - {\mathbf{b}}_\omega -\mathbf{n}_\omega) {^I_G \bar q}  \label{eq:q_dot} \\
\dot{\mathbf{b}}_{\omega} &= \mathbf{n}_{\omega b} \\
{}^G\dot{\mathbf{v}}_I &= {}_I^G\mathbf{R}\left(\mathbf{a}_m- \mathbf{b}_a- \mathbf{n}_a\right) - {}^G\mathbf{g} \\
\dot{\mathbf{b}}_{a} &= \mathbf{n}_{ab} \\
{}^G\dot{\mathbf{p}}_I &= {}^G\mathbf{v}_I \label{eq:p_dot}
\intertext{where
}\bm \Omega(\mathbf{\bm\omega}) &= \begin{bmatrix} -\lfloor \bm\omega \rfloor  && \bm\omega \\ - \bm\omega^{\top} && 0 \end{bmatrix}
\end{align}

\subsection{Standard IMU Processing}

Given a series of IMU measurements, $\mathcal{I}$, collected over a time interval $[t_k, t_{k+1}]$, 
the standard (graph-based) IMU processing considers the following propagation function:
\begin{align}
    \mathbf{x}_{k+1} = \mathbf{g}\left(\mathbf{x}_{k}, \mathcal{I}, \mathbf{n} \right)
\end{align}
That is, the future state at time step $k+1$ is a function of the current state at step $k$, the IMU measurements $\mathcal{I}$, and the corresponding measurement noise $\mathbf{n}$.
Conditioning on the current state, the expected value of the next state is found by evaluating the propagation function with zero noise:
\begin{align}
    \breve{\mathbf{x}}_{k+1} = \mathbf{g}\left({\mathbf{x}}_{k}, \mathcal{I}, \mathbf{0} \right) \label{eq:standard_imu_res}
\end{align}
which implies that we perform integration of the state dynamics in the absence of noise.

The residual for use in batch optimization of this propagation now constrains the start and end states of the interval and is given by (see Equation~\eqref{eq::MLE}):
\begin{align}
c_{IMU}(\mathbf{x}) &= \frac{1}{2}\left| \left| \mathbf{x}_{k+1} \boxminus \breve{\mathbf{x}}_{k+1} \right| \right|_{\mathbf{Q}_k^{-1}}^2 \\ &= \frac{1}{2} \left| \left| \mathbf{x}_{k+1} \boxminus \mathbf{g}\left(\mathbf{x}_{k}, \mathcal{I}, \mathbf{0} \right)\right| \right|_{\mathbf{Q}_k^{-1}}^2
\end{align}
where $\mathbf{Q}_k$ is the linearized, discrete-time noise covariance computed from the IMU noise characterization and is a {\em function of the state}.
This noise covariance matrix and the propagation function can be found by the integration of Equations \eqref{eq:q_dot}-\eqref{eq:p_dot} and their associated error state dynamics, to which we refer the reader to \citep{Trawny2005_Q_TR,Mourikis2007ICRA}.
It is clear from  \eqref{eq:standard_imu_res} that ideally we need to constantly re-evaluate the propagation function $\mathbf{g}(\cdot)$ and the residual covariance $\mathbf{Q}_k$ whenever the linearization point (state estimate) changes.
However, the high frequency nature of the IMU sensors and the complexity of the propagation function and the noise covariance, 
can make direct incorporation of IMU data in real-time graph-based SLAM  prohibitively expensive.
This motivates the development of inertial preintegration.

\subsection{Model 1: Piecewise Constant Measurements} \label{sec:model1}

IMU preintegration seeks to directly reduce the computational complexity of incorporating inertial measurements by removing the need to re-integrate the propagation function and noise covariance.
This is achieved by processing IMU measurements in a $\textit{local}$ frame of reference, yielding measurements that are, in contrast to Equation \eqref{eq:standard_imu_res}, independent of the state \citep*{Lupton2012TOR}.  

Specifically, 
by denoting $\Delta T = t_{k+1}- t_{k}$, we have the following relationship between a series of IMU measurements, the start state, and the resulting end state~\citep*{Eckenhoff2016WAFR}:
\begingroup
\allowdisplaybreaks
\begin{align}
{}^G \mathbf{p}_{k+1}
&= {}^G \mathbf{p}_{k}+{}^G \mathbf{v}_{k}\Delta T -\frac{1}{2}{}^G \mathbf{g} \Delta T^2 \notag\\ & \hspace{0.2cm}+ {}^{G}_k \mathbf{R}\int_{t_k}^{t_{k+1}} \int_{t_k}^{s} {}^k_{u}\mathbf{R}\left(\mathbf{a}_m- \mathbf{b}_a -\mathbf{n}_a\right) du ds\label{eq:dp}\\
{}^G \mathbf{v}_{k+1} &= {}^G \mathbf{v}_{k}-{}^G \mathbf{g} \Delta T \notag\\
&\hspace{1cm}+ {}^G_k\mathbf{R}\int_{t_k}^{t_{k+1}} {}^k_{u}\mathbf{R}\left(\mathbf{a}_m- \mathbf{b}_a -\mathbf{n}_a\right) du \label{eq:dv} \\
{}^{k+1}_G \mathbf{R} &= {}^{k+1}_k \mathbf{R}~{}^{k}_G \mathbf{R} \\
\mathbf{b}_{\omega_{k+1}} &= \mathbf{b}_{\omega_{k}} + \int_{t_k}^{t_{k+1}} \mathbf{n}_{\omega b} ~du \\
\mathbf{b}_{a_{k+1}} &= \mathbf{b}_{a_{k}} + \int_{t_k}^{t_{k+1}} \mathbf{n}_{a b} ~du
\end{align}
\endgroup
{where $u$ and $s$ are dummy variables in the integration}. From the above, we  define the following preintegrated IMU measurements:\footnote{
Note that  along with the preintegrated inertial measurements in Equations~\eqref{eq:aaa} and \eqref{eq:bbb},
the preintegrated relative-orientation measurement ${}^{k+1}_k \bar{q}$ (or ${}^{k+1}_k \mathbf{R}$)
can  be obtained from the integration of the gyro measurements.}
\begin{align}
{}^k \bm \alpha_{k+1} &= \int_{t_k}^{t_{k+1}} \int_{t_k}^{s} {}^k_{u}\mathbf{R}\left(\mathbf{a}_m- \mathbf{b}_a -\mathbf{n}_a\right) du ds \label{eq:aaa}\\
{}^k \bm \beta_{k+1} &= \int_{t_k}^{t_{k+1}} {}^k_{u}\mathbf{R}\left(\mathbf{a}_m- \mathbf{b}_a -\mathbf{n}_a\right) du \label{eq:bbb}
\end{align}

To remove the dependencies of the above preintegrated measurements on the true biases, we linearize about the current bias estimates at time step $t_k$, ${\mathbf{b}}^\star_{a_k}$ and ${\mathbf{b}}^\star_{\omega_k}$.
Defining $\Delta \mathbf{b}= \mathbf{b} - {\mathbf{b}}^\star$, we have
{(noting that time indices are occasionally omitted to keep expressions concise, which however can be easily inferred from the context):}
\begin{align}
&{{^k_G\mathbf{R}} \left( ^G\mathbf{p}_{k+1} - {^G\mathbf p}_{k} - {^G\mathbf{v}_k\Delta T} +\frac{1}{2} {^G\mathbf g}\Delta T^2 \right)  \simeq  }  \label{eq:alpha}\\
&\hspace{0.2cm}{^k\bm\alpha_{k+1}}\left({\mathbf{b}}^\star_{\omega_k}, {\mathbf{b}}^\star_{a_k}\right)
+\frac{\partial \bm\alpha}{\partial  {\mathbf{b}}_\omega }\Big{|}_{{\mathbf{b}}^\star_{\omega_k}} \Delta {\mathbf{b}}_\omega
+\frac{\partial \bm\alpha}{\partial  {\mathbf{b}}_a }\Big{|}_{{\mathbf{b}}^\star_{a_k}} \Delta {\mathbf{b}}_a \nonumber\\[6pt]
&{}^k_G\mathbf{R}\left( ^G\mathbf{v}_{k+1} - {^G\mathbf v}_{k}+ {^G\mathbf g}\Delta T\right)  \simeq   \label{eq:beta}\\
&\hspace{0.2cm}{}^k\bm\beta_{k+1}\left({\mathbf{b}}^\star_{\omega_k}, {\mathbf{b}}^\star_{a_k}\right)
+\frac{\partial \bm\beta}{\partial  {\mathbf{b}}_\omega }\Big{|}_{{\mathbf{b}}^\star_{\omega_k}} \Delta {\mathbf{b}}_\omega
+\frac{\partial \bm\beta}{\partial {\mathbf{b}}_a }\Big{|}_{{\mathbf{b}}^\star_{a_k}} \Delta {\mathbf{b}}_a  \nonumber\\[6pt]
&{{^{k+1}_G\mathbf R}~{^{k}_G\mathbf R}^{\top}}
\simeq \mathbf{R}\left(\frac{\partial \mathbf R}{\partial  {\mathbf{b}}_\omega }\Big{|}_{{\mathbf{b}}^\star_{\omega_k}} \Delta {\mathbf{b}}_\omega \right){^{k+1}_k\mathbf R}\left({\mathbf{b}}^\star_{\omega_k} \right) 
\label{eq:theta}
\end{align}
Note that Equations \eqref{eq:alpha} and \eqref{eq:beta} are simple Taylor series expansions for our ${}^{k}{\bm \alpha}_{k+1}$ and ${}^{k}{\bm\beta}_{k+1}$ measurements, while Equation \eqref{eq:theta} models an additional  rotation induced due to a change of the linearization point (estimate) of the gyro bias \citep*{Forster2015RSS,Eckenhoff2016WAFR}.

The preintegrated measurement's mean values, ${}^k\breve{\bm\alpha}_{k+1}$, ${}^k\breve{\bm\beta}_{k+1}$, and ${}^{k+1}_k\breve{\bar{q}}$, must be computed for use in graph optimization.
It is important to note that
current preintegration methods~\citep*{Lupton2012TOR,Forster2015RSS,Ling2016ICRA} are all based on discrete integration of the measurement dynamics through Euler or midpoint integration. 
In particular, the discrete approximation used by \cite*{Forster2015RSS} in fact corresponds to a piecewise constant \textit{global acceleration} model 
{(expressed in the fixed global frame of reference), which may be easily violated in realistic navigation.}
By contrast, we here offer {\em closed-form} solutions for the measurement means under the assumptions of piecewise constant (local) \textit{measurements} 
and piecewise constant {\em local acceleration} 
{(expressed in local coordinates)}
which will be presented later in Section~\ref{sec:model2}.

\subsubsection{Computing preintegration mean:} \label{sec:compute-mean-model1}

Between two image times, $t_k$ and $t_{k+1}$, the IMU receives a series of inertial measurements.
We denote $\tau$ as the step at which an IMU measurement is received, and $\tau+1$ as the step of the $\textit{next}$ IMU reading.
The time associated with each of these steps is given by $t_{\tau}$ and $t_{\tau+1}$, respectively. 
The relative orientation between the interval, ${}^{k+1}_k\breve{\bar{q}}$, can be found using successive applications of the zeroth order quaternion integrator \citep*{Trawny2005_Q_TR}.
Based on the definitions of ${}^k{\bm \alpha}_{k+1}$ and ${^k\bm\beta}_{k+1}$ (see Equations \eqref{eq:aaa} and \eqref{eq:bbb}), 
we have the following continuous-time dynamics at every step $u$ with $t_u \in [t_\tau,t_{\tau+1}]$:
\begin{align}
{}^k\dot{\bm \alpha}_{u} &= {}^k{\bm \beta}_{u}  \label{eq:alpha_dot} \\
{}^k\dot{\bm \beta}_{u}  &= {}^k_{u}\mathbf {R} \left(\mathbf{a}_m-{\mathbf{b}}_a -\mathbf{n}_a \right)  \label{eq:beta_dot} 
\end{align}
From these governing differential equations, we formulate the following {\em linear} system that describes the evolution of the measurements by taking the expectation operation:
\begin{equation}
\begin{bmatrix}
{^k\dot{\breve{\bm \alpha}}}_{u}\\
{^k\dot{\breve{\bm \beta}}}_{u}
\end{bmatrix}= 
\begin{bmatrix}
\mathbf{0} & \mathbf{I} \\
\mathbf{0} & \mathbf{0} 
\end{bmatrix} \begin{bmatrix}
{^k\breve{\bm \alpha}}_{u} \\
{^k\breve{\bm \beta}}_{u}
\end{bmatrix}+ 
\begin{bmatrix}
\mathbf{0} \\
^{k}_{u}\breve{\mathbf{R}} 
\end{bmatrix}
(\mathbf{a}_m-{\mathbf{b}}^\star_{a_k})  
\label{eq:linear_system}
\end{equation}

Given  $ {\bm a}_m$ and $ {\bm \omega}_m$ sampled at time $t_{\tau}$ and assuming that these local IMU 
measurements are {\em piecewise constant} during $[t_\tau, t_{\tau+1}]$,
we analytically solve the above linear time-varying (LTV) system to obtain the updated preintegration mean values, which are computed as follows \citep*{Eckenhoff2018TR}:
{
\begin{align}
&\begin{bmatrix}
{^k\breve{\bm \alpha}}_{\tau+1} \\
{^k \breve{\bm\beta}}_{\tau+1}
\end{bmatrix}
=
\begin{bmatrix}
{}^k\breve{\bm \alpha}_{\tau} + {^k\breve{\bm \beta}}_{\tau}\Delta t +\mathbf{A}_{\tau} \hat{\mathbf a}\\ 
{}^k\breve{\bm \beta}_{\tau} + \mathbf{B}_{\tau} \hat{\mathbf a}
\end{bmatrix}
\label{eq:mean}
\\[5pt]
&
\scalemath{0.9}{
\mathbf{A}_{\tau} =
{}^{k}_{\tau+1}\breve{\mathbf{R}}\Big(\frac{\Delta t^2}{2}\mathbf{I}_{3 \times 3} + \frac{|\hat{\bm\omega}|\Delta t \textrm{cos}(|\hat{\bm\omega}|\Delta t)-\textrm{sin}(|\hat{\bm \omega}|\Delta t)}{|\hat{\bm\omega}|^3}\lfloor \hat{\bm\omega} \rfloor
}
\notag\\[1pt]
&
\scalemath{0.9}{
+ \frac{(|\hat{\bm\omega}|\Delta t)^2-2\textrm{cos}(|\hat{\bm\omega}|\Delta t)-2(|\hat{\bm\omega}|\Delta t)\textrm{sin}(|\hat{\bm\omega}|\Delta t)+2}{2|\hat{\bm\omega}|^4}\lfloor \hat{\bm\omega} \rfloor^2\Big)
} \label{eq:Atau}\\[7pt]
&
\scalemath{0.9}{
\mathbf{B}_{\tau} = {}^{k}_{\tau+1}\breve{\mathbf{R}}\Big(\Delta t \mathbf{I}_{3 \times 3} - \frac{1-\textrm{cos}(|\hat{\bm\omega}|(\Delta t))}{|\hat{\bm\omega}|^2}\lfloor \hat{\bm\omega} \rfloor
} \notag \\[1pt]
&\hspace{1cm}
\scalemath{0.9}{
+ \frac{(|\hat{\bm\omega}|\Delta t)-\textrm{sin}(|\hat{\bm\omega}|\Delta t)}{|\hat{\bm\omega}|^3}\lfloor \hat{\bm\omega} \rfloor^2\Big)
} \label{eq:Btau}
\end{align}
}where we have employed the definitions: $\hat{\bm \omega}= {\bm \omega}_m- {\mathbf{b}}^\star_{\omega_k}$ , $\hat{\mathbf a}= {\mathbf a}_m- {\mathbf{b}}^\star_{a_k}$, and $\Delta t= t_{\tau+1}- t_{\tau}$.
Clearly, 
these \textit{closed-form} expressions reveal the higher order affect of the angular velocity on the preintegrated measurements due to the evolution of the orientation over the IMU samping interval.

\subsubsection{Computing preintegration covariance:}

In order to derive the preintegrated measurement covariance, we first write the linearized \textit{measurement} error system as follows \citep*{Eckenhoff2018TR}:
\begingroup
\allowdisplaybreaks
\begin{align}
\scalemath{.96}{\begin{bmatrix} {}^{u}\dot{{\delta \bm \theta}}_k \\  \dot{\widetilde{\mathbf{b}}}_{\omega} \\ {}^k\dot{\delta\bm \beta}_{u} \\ \dot{\widetilde{\mathbf{b}}}_{a} \\  {}^k\dot{\delta {\bm \alpha}}_{u} \end{bmatrix}}
&\scalemath{.96}{= \begin{bmatrix} -\lfloor \hat{\bm \omega} \rfloor & -\mathbf{I} & \mathbf{0} & \mathbf{0} & \mathbf{0}\\
\mathbf{0}& \mathbf{0} & \mathbf{0} & \mathbf{0} & \mathbf{0} \\
-{}^k_{u} \breve{\mathbf{R}} \lfloor \hat{\mathbf{a}} \rfloor & \mathbf{0} & \mathbf{0} & -{}^k_{u} \breve{\mathbf{R}} & \mathbf{0} \\
\mathbf{0} & \mathbf{0} & \mathbf{0} & \mathbf{0} & \mathbf{0} \\   \mathbf{0} & \mathbf{0} & \mathbf{I} & \mathbf{0} & \mathbf{0} \end{bmatrix}
\begin{bmatrix} {}^{u}{\delta \bm \theta}_k \\ \widetilde{\mathbf{b}}_{\omega} \\ {}^k\delta {\bm \beta}_{u} \\ \widetilde{\mathbf{b}}_{a} \\  {}^k\delta {\bm \alpha}_{u} \end{bmatrix}} \notag\\
&\scalemath{.96}{+ \begin{bmatrix} -\mathbf{I} & \mathbf{0} & \mathbf{0}& \mathbf{0} \\ \mathbf{0} & \mathbf{I} & \mathbf{0} & \mathbf{0} \\ \mathbf{0} & \mathbf{0} & -{}^k_{u} \breve{\mathbf{R}}& \mathbf{0}\\
\mathbf{0} & \mathbf{0} & \mathbf{0} & \mathbf{I} \\
\mathbf{0} & \mathbf{0} & \mathbf{0} & \mathbf{0} \end{bmatrix} \begin{bmatrix} \mathbf{n}_{\omega} \\ \mathbf{n}_{\omega b} \\ \mathbf{n}_{a} \\ \mathbf{n}_{ab} \end{bmatrix}} \label{eq:model1-error-sys}\\
\Longleftrightarrow& ~~\dot{\mathbf{r}} = \mathbf{F}\mathbf{r} +\mathbf{G}\mathbf{n}
\end{align}
\endgroup
which is akin to the standard VINS error state propagation equations in a local frame of reference \citep*{Mourikis2007ICRA}.

It is important to note that in contrast to our previous work \citep*{Eckenhoff2016WAFR,Eckenhoff2017ICRA}, 
we here couple the preintegration bias and measurement evolution for improved accuracy.
Note also that the bias error terms in Equation~\eqref{eq:model1-error-sys}, $\widetilde{\mathbf{b}}_\omega$ and $\widetilde{\mathbf{b}}_a$,  describe the deviation of the bias over the interval due to the random-walk drift, rather than the error of the current bias estimate.
{The discrete state transition matrix $\bm \Phi (t_{\tau+1}, t_{\tau})$ can be  computed either analytically in closed-form  or numerically using Runge-Kutta methods based on the following continuous-time differential equation (see \citet*{Hesch2013TRO,Trawny2005_Q_TR}):}
\begin{align}
\dot{\bm \Phi}(t_u, t_{\tau}) &= \mathbf{F}(u)~{\bm \Phi}(t_u, t_{\tau}) \label{eq:statetrans1}\\
\bm \Phi (t_{\tau}, t_{\tau})&= \mathbf{I} \color{black} \label{eq:statetrans2}
\end{align}
The propagation of the measurement covariance, $\mathbf{P}$, 
{over the time interval $t_{\tau}\in [t_k, t_{k+1}]$},
takes the following form:
\begin{align}
&~~~~\mathbf{P}_k = \mathbf{0} \color{black} \\
&\mathbf{P}_{\tau+1} = \bm \Phi (t_{\tau+1}, t_{\tau})~\mathbf{P}_{\tau}~\bm \Phi (t_{\tau+1}, t_{\tau})^{\top} + \mathbf{Q}_{\tau}  \\
\mathbf{Q}_{\tau} &= \scalemath{.93}{\int_{t_\tau}^{t_{\tau+1}}  \bm \Phi (t_{\tau+1}, u)\mathbf{G}(u)\mathbf{Q}_c \mathbf{G}(u)^{\top} \bm \Phi (t_{\tau+1}, u)^{\top} du}  \label{eq:preintnoise}
\end{align}
where $\mathbf{Q}_c$ is the continuous-time IMU noise covariance. 
To keep presentation concise, the discrete-time noise covariance $\mathbf Q_\tau$,
can be computed similarly as in~\citep{Trawny2005_Q_TR}.

\subsubsection{Preintegration measurement residuals and Jacobians:} \label{sec:model1-res-jac}

For use in optimization,
we form the associated preintegration measurement cost and residual as follows:
\begin{align}
c_{IMU}(\mathbf{x}) &= \frac{1}{2}\left| \left| \mathbf{e}_{IMU} (\mathbf{x}) \right| \right|_{ \mathbf{P}_{k+1}^{-1}}^2
\label{eq:model1-res-cost}
\end{align}
\begin{align}
&\scalemath{0.95}{\mathbf{e}_{IMU}(\mathbf{x})=} \label{eq:model1-residual}\\
&~~~~\scalemath{0.95}{
\begin{bmatrix}
2\mathbf{vec}\left({}^{k+1}_G\bar{q} \otimes  {}^{k}_G\bar{q}^{-1} \otimes  {}^{k+1}_k\breve{\bar{q}}^{-1} \otimes \bar{q}_b \right) \\[4pt] \hdashline[2pt/2pt] \\[-8pt]
\mathbf{b}_{{\omega}_{k+1}}- \mathbf{b}_{{\omega}_{k}} \\[4pt] \hdashline[2pt/2pt] \\[-8pt]
\Bigg({{}^k_G\mathbf{R}\left({}^G\mathbf{v}_{k+1}-{}^G\mathbf{v}_{k}+ {}^G\mathbf{g}\Delta T \right)}  \\
{-\mathbf{J}_\beta \left(\mathbf{b}_{\omega_k}- {\mathbf{b}}^\star_{\omega_k} \right)-  \mathbf{H}_\beta \left(\mathbf{b}_{a_k}- {\mathbf{b}}^\star_{a_k} \right)- {}^k\breve{\bm \beta}_{k+1}} \Bigg) \\[4pt] \hdashline[2pt/2pt] \\[-8pt]
\mathbf{b}_{{a}_{k+1}}- \mathbf{b}_{{a}_{k}} \\[4pt] \hdashline[2pt/2pt] \\[-8pt]
\Bigg({{}^k_G\mathbf{R}\left({}^G\mathbf{p}_{k+1}-{}^G\mathbf{p}_{k}- {}^G\mathbf{v}_{k}\Delta T + \frac{1}{2}{}^G\mathbf{g}\Delta T^2 \right)} \\ 
{-\mathbf{J}_\alpha \left(\mathbf{b}_{\omega_k}- {\mathbf{b}}^\star_{\omega_k} \right)-  \mathbf{H}_\alpha \left(\mathbf{b}_{a_k}- {\mathbf{b}}^\star_{a_k} \right)- {}^k\breve{\bm \alpha}_{k+1}} \Bigg)
\end{bmatrix} 
}
\notag
\end{align}
where we have employed  $\bar{q}_b= \begin{bmatrix} \frac{\bm \theta}{||\bm \theta||}~\sin{\left(\frac{||\bm \theta||}{2}\right)} \\ \cos{\left(\frac{||\bm \theta||}{2}\right)} \end{bmatrix}$ 
and $ \bm \theta = \mathbf{J}_q \left( \mathbf{b}_{\omega_k} -  {\mathbf{b}}^\star_{\omega_k}\right)$.
In the above expressions, 
$\mathbf{J}_q, ~\mathbf{J}_\alpha,  ~\mathbf{J}_\beta, ~\mathbf{H}_\alpha,$ and $\mathbf{H}_\beta$, are the Jacobian matrices of the pertinent residuals with respect to the biases,
which are used to correct the measurements due to a change in the initial bias estimate $\mathbf b^\star$, 
thus compensating for the fact that preintegrated measurements have been linearized about $\mathbf{b}^\star_{\omega_k}$ and $\mathbf{b}^\star_{a_k}$ without having to recompute the required integrals whenever the bias estimates change 
(see Equations~\eqref{eq:alpha} and \eqref{eq:beta}).
In particular, using the fact that our preintegrated  measurement means are linear in the acceleration bias $\mathbf b_a$ (see Equation \eqref{eq:mean}), we have the following dynamics of its Jacobians
(see Equations~\eqref{eq:Atau} and \eqref{eq:Btau}):
{
\begin{align}
\begin{bmatrix}
\frac{\partial \bm \alpha}{\partial \mathbf b_a} \\[3pt]
\frac{\partial \bm \beta}{\partial \mathbf b_a}
\end{bmatrix} &=:
\begin{bmatrix}
\mathbf{H}_{\alpha}\left({\tau+1}\right) \\[3pt]
\mathbf{H}_{\beta}\left({\tau+1}\right)
\end{bmatrix} \notag\\
&=
\begin{bmatrix}
\mathbf{H}_{\alpha}\left({\tau}\right)+ \mathbf{H}_{\beta}\left({\tau}\right)\Delta t - \mathbf{A}_{\tau}\\[3pt]
\mathbf{H}_{\beta}\left({\tau}\right)-\mathbf{B}_{\tau}
\end{bmatrix}  
\end{align}
}Similarly, for the gyroscope bias Jacobians, we have:
{
\begin{align}
\begin{bmatrix}
\frac{\partial \bm \alpha}{\partial \mathbf b_\omega} \\[3pt]
\frac{\partial \bm \beta}{\partial \mathbf b_\omega}
\end{bmatrix} &=:
\begin{bmatrix}
\mathbf{J}_{\alpha}\left({\tau+1}\right) \\[3pt]
\mathbf{J}_{\beta}\left({\tau+1}\right)
\end{bmatrix} \notag\\
&=
\begin{bmatrix}
\mathbf{J}_{\alpha}\left({\tau}\right) + \mathbf{J}_{\beta}\left({\tau}\right) \Delta t +\frac{\partial \mathbf{A}_{\tau}\hat{\mathbf{a}}}{\partial \mathbf{b}_{\omega}}\\[3pt]
\mathbf{J}_{\beta}\left({\tau}\right)+\frac{\partial \mathbf{B}_{\tau}\hat{\mathbf{a}}}{\partial \mathbf{b}_{\omega}}
\end{bmatrix} 
\end{align}
Finally, the orientation Jacobian with respect to gyroscope bias can be found incrementally as:
\begin{align}
        \mathbf{J}_{q}(\tau+1) &=  {^{\tau+1}_{\tau}}\breve{\mathbf{R}}\mathbf{J}_{q}(\tau)+ \mathbf{J}_r\left(\hat{\bm \omega}\Delta t\right)\Delta t
\end{align}
where $\mathbf{J}_r\left(\cdot\right)$ is the right Jacobian of $SO(3)$ and is defined as \citep*{chirikjian2011stochastic}:
\begin{equation} \label{eq:right_jacob}
\scalemath{0.83}{
    \mathbf{J}_r(\phi)=\mathbf{I}_{3 \times 3}-\frac{1-\textrm{cos}(\parallel \bm \phi \parallel)}{\parallel \bm \phi \parallel^2}\lfloor\bm\phi\rfloor +\frac{\parallel \bm\phi \parallel-\textrm{sin}(\parallel \bm\phi \parallel)}{\parallel \bm\phi \parallel^3}\lfloor\bm\phi\rfloor^2
}
\end{equation}
}Moreover, the measurement Jacobians of these preintegrated measurements with respect to the error state~\eqref{eq:stateerror}, 
can also be analytically computed as shown in Appendix B.1, which are essential for batch optimization.
For the detailed derivations and closed-form expressions of the preintegrated measurements and Jacobians, 
the reader is referred to our companion technical report~\citep*{Eckenhoff2018TR}.

\subsection{Model 2: Piecewise Constant Local Acceleration} \label{sec:model2}

\begin{figure}
\centering
\includegraphics[height=5cm]{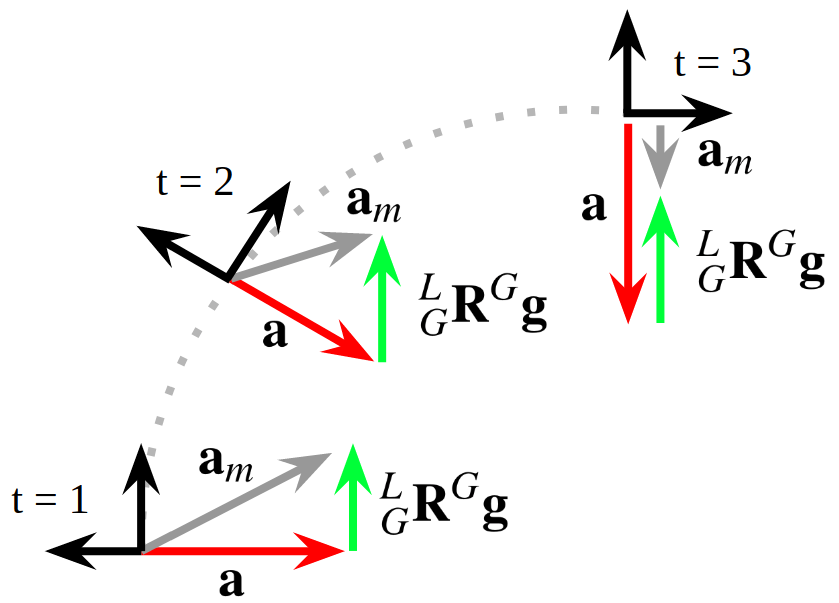}
\caption{An example of an IMU rotating about the gravity. It can be seen that the true local acceleration $\mathbf{a}$ (red) remains constant, while its local measurement $\mathbf{a}_m$ (grey) changes continuously due to the effect of gravity (green). \color{black}}
\label{fig:accel_breakout}
\end{figure}

The previous preintegration (Model 1) assumes that noiseless IMU measurements can be approximated as remaining constant over a sampling interval, which, however, might not always be a good approximation (see Figure \ref{fig:accel_breakout}).
In this section, we propose a new preintegration model that instead assumes piecewise constant \textit{true} local acceleration during the sampling time interval, which may better approximate motion dynamics in practice. 
To this end, we first  rewrite  Equations \eqref{eq:dp} and \eqref{eq:dv} as:
\begin{align}
^G\mathbf{p}_{k+1} &=
{}^G\mathbf {p}_{k} + {^G\mathbf{v}_k\Delta T}  +  {^G_k\mathbf{R}}\int_{t_k}^{t_{k+1}}\int_{t_k}^{s}{^{k}_{u}\mathbf{R}}\mathbf{a}~duds \label{eq:p-model2}\\
^G\mathbf{v}_{k+1} &=
{}^G\mathbf {v}_{k}+  {^G_k\mathbf{R}}\int_{t_k}^{t_{k+1}}{^{k}_{u}\mathbf{R}}\mathbf{a}~du \label{eq:v-model2}
\end{align}
Note that we have moved the effect of gravity back inside the integrals. We then define the following vectors:
\begin{align}
    \Delta p &= \int_{t_k}^{t_{k+1}}\int_{t_k}^{s}{^{k}_{u}\mathbf{R}}\mathbf{a}~duds \\
    \Delta v &= \int_{t_k}^{t_{k+1}}{^{k}_{u}\mathbf{R}}\mathbf{a}~du
\end{align}
which essentially are the true local position displacement and  velocity change during $[t_k, t_{k+1}]$, and yields:
\begin{align} \label{eq:print2del-p}
    \Delta \dot{p} &= \Delta v  \\
    \Delta \dot{v } &= {^{k}_{u}\mathbf{R}} \mathbf{a}
    \label{eq:print2delv}
\end{align}
In particular, between two IMU measurement times inside the preintegration interval, $\left[ t_{\tau}, t_{\tau+1} \right] \subset \left[ t_{k}, t_{k+1} \right]$, we assume that the $\textit{local}$ acceleration will be constant:
\begin{align}
    \forall t_u \in \left[ t_{\tau}, t_{\tau+1} \right], ~~~~\mathbf{a}(t_u)= \mathbf{a}(t_\tau)
\end{align}
Using this sampling model we can rewrite  \eqref{eq:print2delv} as:
\begin{align} \label{eq:del-v-new}
     \Delta \dot{v } &= {^{k}_{u}\mathbf{R}}\left(\mathbf{a}_m- \mathbf{b}_a - \mathbf{n}_a- {^{\tau}_{k}\mathbf{R}}{^{k}_{G}\mathbf{R}}{}^G\mathbf{g} \right)
\end{align}
We now write the relationship of the states at the beginning and end of the interval as (see Equations~\eqref{eq:p-model2} and \eqref{eq:v-model2}):
\begin{align}
 {^k_G\mathbf{R}} \left(^G\mathbf{p}_{k+1} -
{}^G\mathbf {p}_{k} - {^G\mathbf{v}_k\Delta T}\right)  &= \Delta p \\
 {^k_G\mathbf{R}}\left( ^G\mathbf{v}_{k+1} -
{}^G\mathbf {v}_{k}\right)  &= \Delta v
\end{align}
It is important to note that, since $\Delta p$ and $\Delta v$ are functions of both the biases \textit{and} the initial orientation,
we  perform the following linearization with respect to these states:
\begin{align}
&{^k_G\mathbf{R}} \left(^G\mathbf{p}_{k+1} -
{}^G\mathbf {p}_{k} - {^G\mathbf{v}_k\Delta T}\right) \simeq \Delta p \left({\mathbf{b}}^\star_{\omega_k}, {\mathbf{b}}^\star_{a_k}, {}^k_G\bar{q}^\star \right) \notag\\
&\hspace{0.1cm}
+\frac{\partial \Delta p}{\partial \mathbf{b}_\omega }\Big{|}_{{\mathbf{b}}^\star_{\omega_k}} \Delta \mathbf{b}_\omega
+\frac{\partial \Delta p}{\partial \mathbf{b}_a }\Big{|}_{{\mathbf{b}}^\star_{a_k}} \Delta \mathbf{b}_a
+\frac{\partial \Delta p}{\partial \Delta \bm \theta_k }\Big{|}_{{}^k_G\bar{q}^\star} \Delta \bm \theta_k  \notag\\[6pt]
&{^k_G\mathbf{R}}\left( ^G\mathbf{v}_{k+1} - {}^G\mathbf {v}_{k}\right)  \simeq \Delta v\left({\mathbf{b}}^{\star}_{\omega_k}, {\mathbf{b}}^{\star}_{a_k}, {}^k_G\bar{q}^\star \right) \\
&\hspace{0.1cm} 
+\frac{\partial \Delta v}{\partial \mathbf{b}_\omega }\Big{|}_{{\mathbf{b}}^\star_{\omega_k}} \Delta \mathbf{b}_\omega
+\frac{\partial \Delta v}{\partial \mathbf{b}_a }\Big{|}_{{\mathbf{b}}^\star_{a_k}} \Delta \mathbf{b}_a
+\frac{\partial \Delta v}{\partial \Delta \bm \theta_k }\Big{|}_{{}^k_G\bar{q}^\star} \Delta \bm \theta_k 
\end{align}
where $\Delta \bm \theta_k = 2\mathbf{vec}\left({}^k_G\bar{q} \otimes {}^k_G\bar{q}^{\star-1}\right)$ 
is the rotation angle change associated with the change of the  linearization point of quaternion ${}^k_G\bar{q}$.

\subsubsection{Computing preintegration mean:}

To compute the new preintegrated measurement mean values, 
we first determine the continuous-time dynamics of the expected preintegration vectors by taking expectations of Equations~\eqref{eq:print2del-p} and \eqref{eq:del-v-new}, given by:
\begin{align}
    \Delta \dot{\breve{p}} &= \Delta \breve{{v}} \label{eq:dp-mean-model2}\\
    \Delta \dot{\breve{{v}}} &= {^{k}_{u}\breve{\mathbf{R}}}\left(\mathbf{a}_m- \mathbf{b}^\star_{a_k} - {^{\tau}_{k}\breve{\mathbf{R}}}{^{k}_{G}{\mathbf{R}}^\star}{}^G\mathbf{g} \right)
    \label{eq:dv-mean-model2}
\end{align}
As in the case of Model~1 (see Section~\ref{sec:compute-mean-model1}),
we can formulate a linear system of the new preintegration measurement vectors and find the closed-from solutions.
Specifically, 
we can integrate these differential equations and obtain the solution similar to Equation~\eqref{eq:mean},
while using the new definition: $\hat{\mathbf{a}}= \mathbf{a}_m- {\mathbf{b}}^\star_{a_k} - {^{\tau}_{k}\breve{\mathbf{R}}}{^{k}_{G}{\mathbf{R}^\star}}{}^G\mathbf{g}$,
which serves as the estimate for the piecewise constant local acceleration over the sampling interval.

\subsubsection{Computing preintegration covariance:}

To compute the new preintegration measurement covariance, 
we first determine the differential equations for the corresponding preintegration measurement errors 
(see Equations~\eqref{eq:print2del-p}, \eqref{eq:del-v-new}, \eqref{eq:dp-mean-model2} and \eqref{eq:dv-mean-model2}):
{
\begin{align}
\Delta \dot{\tilde{p}}
&= \Delta v- \Delta \breve{v} = {\Delta \tilde{v}} \\
{\Delta \dot{\tilde{v}}}
&= {^{k}_{u}\breve{\mathbf{R}}}\left(\mathbf{I} + \lfloor {}^u\delta\bm\theta_k \rfloor \right) \Big(\mathbf{a}_m- {\mathbf{b}}^\star_{a_k}- \tilde{\mathbf{b}}_a \nonumber\\
&\hspace{1cm}- \left(\mathbf{I} - \lfloor {}^\tau\delta\bm\theta_k \rfloor \right){^{\tau}_{k}\breve{\mathbf{R}}}{^{k}_{G}{\mathbf{R}^\star}}{}^G\mathbf{g} -\mathbf{n}_a \Big)  \nonumber\\
&\hspace{1cm}- {^{k}_{u}\breve{\mathbf{R}}}\left(\mathbf{a}_m- \mathbf{b}^\star_{a_k} - {^{\tau}_{k}\breve{\mathbf{R}}}{^{k}_{G}{\mathbf{R}}^\star}{}^G\mathbf{g} \right) \nonumber\\
&=
-{}^k_{u}\breve{\mathbf{R}}\lfloor \hat{\mathbf{a}} \rfloor {}^u\delta\bm\theta_k 
-{}^k_{u}\breve{\mathbf{R}}\tilde{\mathbf{b}}_a \notag \\
&\hspace{0.4cm}
-{}^k_{u}\breve{\mathbf{R}}\lfloor{}^{\tau}\breve{\mathbf{g}} \rfloor{}^\tau\delta\bm\theta_k
-{}^k_{u}\breve{\mathbf{R}}\mathbf{n}_a 
\end{align}
where  ${}^\tau\breve{\mathbf{g}}$ represents the estimate for gravity in the sampled $\tau$ frame.}
It is important to notice that,
in the above expressions, we have used {two} angle errors: 
(i) ${}^u\delta\bm\theta_k$ corresponds to the active local IMU orientation error, and
(ii) ${}^\tau\delta\bm\theta_k$ corresponds to the cloned orientation error at the sampling time $t_\tau$.
In addition, the bias errors $\tilde{\mathbf{b}}$ describe the deviation of the bias from the starting value over the interval due to bias drift.
With this, we have the following time evolution of the full preintegrated measurement errors:
{
\begin{align}
    \begin{bmatrix}
    {}^u\delta\dot{\bm\theta}_k \\ 
    \dot{\tilde{\mathbf{b}}}_\omega \\
    \Delta{\dot{\tilde{v}}} \\
    \dot{\tilde{\mathbf{b}}}_a \\
    \Delta{\dot{\tilde{p}}} \\
    {}^\tau\delta\dot{\bm\theta}_k
    \end{bmatrix} &=
    \mathbf{F}
    \begin{bmatrix}
    {}^u\delta{\bm\theta}_k \\ 
    \tilde{\mathbf{b}}_\omega \\
    \Delta{{\tilde{v}}} \\
    \tilde{\mathbf{b}}_a \\
    \Delta{{\tilde{p}}} \\
    {}^\tau\delta{\bm\theta}_k
    \end{bmatrix} 
    + \begin{bmatrix}
     -\mathbf{I} & \mathbf{0} & \mathbf{0} & \mathbf{0}\\
     \mathbf{0} & \mathbf{I} & \mathbf{0} & \mathbf{0} \\
     \mathbf{0} &  \mathbf{0} & - \mathbf{}^k_{u}\breve{\mathbf{R}} & \mathbf{0} \\
     \mathbf{0} & \mathbf{0} & \mathbf{0} & \mathbf{I} \\
     \mathbf{0} & \mathbf{0} & \mathbf{0} & \mathbf{0}\\
     \mathbf{0} & \mathbf{0} & \mathbf{0}& \mathbf{0}
    \end{bmatrix}
    \begin{bmatrix}
    \mathbf{n}_\omega \\ \mathbf{n}_{\omega b} \\ \mathbf{n}_a \\ \mathbf{n}_{ab}
    \end{bmatrix} \notag \\
    \Longleftrightarrow ~~\dot{\mathbf{r}} &= \mathbf{F}\mathbf{r} +\mathbf{G}\mathbf{n}
    \label{eq:model2-prein-error-system}
\end{align}
}
where
{
\begin{align}
    &\mathbf{F} =
    &\scalemath{0.9}{
    \begin{bmatrix} 
    -\lfloor \hat{\bm \omega} \rfloor & -\mathbf{I} & \mathbf{0} &  \mathbf{0} &  \mathbf{0} & \mathbf{0} \\
            \mathbf{0} &  \mathbf{0} &  \mathbf{0} & \mathbf{0} &  \mathbf{0} & \mathbf{0} \\
            -{}^k_{u}\breve{\mathbf{R}}\lfloor \hat{\mathbf{a}} \rfloor & \mathbf{0} & \mathbf{0} & -{}^k_{u}\breve{\mathbf{R}} & \mathbf{0} & -{}^k_{u}\breve{\mathbf{R}}\lfloor {}^{\tau}\breve{\mathbf{g}}\rfloor & \\
            \mathbf{0} &  \mathbf{0} &  \mathbf{0} & \mathbf{0} &  \mathbf{0} & \mathbf{0} \\
        \mathbf{0} &  \mathbf{0} &  \mathbf{I} & \mathbf{0} &  \mathbf{0} & \mathbf{0}  \\
        \mathbf{0} &  \mathbf{0} &  \mathbf{0} & \mathbf{0} &  \mathbf{0} & \mathbf{0} 
    \end{bmatrix}
    }
\end{align}
}

In analogy to Equations \eqref{eq:statetrans1}, \eqref{eq:statetrans2}, and \eqref{eq:preintnoise}, 
we can determine the new state-transition matrix $\bm{\Phi}(t_{\tau+1}, t_\tau)$ and the new discrete noise covariance $\mathbf Q_\tau$.
{With that, we now propagate the measurement covariance over the time interval $t_{\tau}\in[t_k, t_{k+1}]$ as follows:
\begin{align}
\mathbf{P}_{k} &= \mathbf{0}\color{black} \label{eq:cov-prop-model2-1}\\
\mathbf{P}_{\tau+1} &= \bm{\Phi}(t_{\tau+1}, t_\tau)\mathbf{P}_{\tau}\bm{\Phi}(t_{\tau+1}, t_\tau)^{\top}+\mathbf{Q}_{\tau}  \label{eq:cov-prop-model2-2}\\ 
\mathbf{P}_{\tau+1} &= \bm\Gamma \mathbf{P}_{\tau+1} \bm\Gamma^{\top}  \label{eq:cov-prop-model2-3}
\end{align}
where $\bm\Gamma$ is the permutation matrix that allows us to replace the previous static orientation error ${}^\tau\delta{\bm\theta}_k$ by the new one ${}^{\tau+1}\delta{\bm\theta}_k$
simply by cloning the current local orientation error ${}^u\delta\bm\theta_k$ at the end of current sampling interval $t_u = t_{\tau+1}$
when moving from the current measurement time interval $[t_\tau,t_{\tau+1}]$ to the next one,}
and is given by:
{
\begin{align}
    \bm\Gamma &= \begin{bmatrix} \mathbf{I} & \mathbf{0} & \mathbf{0} & \mathbf{0} & \mathbf{0} & \mathbf{0} \\ 
        \mathbf{0} & \mathbf{I}  & \mathbf{0} & \mathbf{0} & \mathbf{0} & \mathbf{0} \\
        \mathbf{0} & \mathbf{0}  & \mathbf{I} & \mathbf{0} & \mathbf{0} & \mathbf{0} \\
        \mathbf{0} & \mathbf{0}  & \mathbf{0} & \mathbf{I} & \mathbf{0} & \mathbf{0} \\
         \mathbf{0} & \mathbf{0}  & \mathbf{0} & \mathbf{0} & \mathbf{I} & \mathbf{0}  \\
          \mathbf{I} & \mathbf{0}  & \mathbf{0} & \mathbf{0} & \mathbf{0} & \mathbf{0}
     \end{bmatrix}
\end{align}
}
The resulting preintegrated measurement covariance is then extracted from the top left 15$\times$15 block of $\mathbf{P}_{k+1}$
after the propagation with Equations~\eqref{eq:cov-prop-model2-1}-\eqref{eq:cov-prop-model2-3} over the entire preintegration interval $[t_k, t_{k+1}]$.

\subsubsection{Preintegration measurement residuals and Jacobians:}

{
As we linearize this preintegration with respect to the IMU biases and the initial orientation, it is important to compute the Jacobians with respect to these quantities. In particular, we note that the solution to the preintegration equation for Model 2 can be expressed as:
\begin{align}
    \Delta \breve{p}_{\tau+1} &= \Delta \breve{p}_{\tau} + \Delta \breve{v}_{\tau}\Delta t +  \mathbf{A}_{\tau}\left(\mathbf{a}_m- \mathbf{b}^{\star}_a - {}^{\tau}_{k}\breve{\mathbf{R}}^{k}_{G}\mathbf{R}^{\star}{}^G\mathbf{g} \right) 
    \notag\\
    \Delta \breve{v}_{\tau+1} &= \Delta \breve{v}_{\tau} + \mathbf{B}_{\tau}\left(\mathbf{a}_m- \mathbf{b}^{\star}_a - {}^{\tau}_{k}\breve{\mathbf{R}}^{k}_{G}\mathbf{R}^{\star}{}^G\mathbf{g} \right)
\end{align}
where $\mathbf{A}_{\tau}$ and $\mathbf{B}_{\tau}$ are defined the same as in Equations \eqref{eq:Atau} and \eqref{eq:Btau}.
Letting $\mathbf{O}_{\alpha}$ and $\mathbf{O}_{\beta}$ denote the Jacobians of the position and velocity preintegrated measurements with respect to the initial orientation, we have:
\begin{align}
\begin{bmatrix}
\mathbf{J}_{\alpha}\left({\tau+1}\right) \\[3pt]
\mathbf{J}_{\beta}\left({\tau+1}\right)
\end{bmatrix}
&=
\begin{bmatrix}
\mathbf{J}_{\alpha}\left({\tau}\right) + \mathbf{J}_{\beta}\left({\tau}\right) \Delta t \\[3pt]
\mathbf{J}_{\beta}\left({\tau}\right)
\end{bmatrix} \\
&\hspace{0.4cm}+
 \begin{bmatrix}
\frac{\partial \mathbf{A}_{\tau}\hat{\mathbf{a}}}{\partial \mathbf{b}_{\omega}}+\mathbf{A}_{\tau}\lfloor {}^{\tau}_{k}\breve{\mathbf{R}}^{k}_{G}\mathbf{R}^{\star}{}^G\mathbf{g} \rfloor \mathbf{J}_q\left(\tau\right)\\[3pt]
\frac{\partial \mathbf{B}_{\tau}\hat{\mathbf{a}}}{\partial \mathbf{b}_{\omega}} +\mathbf{B}_{\tau}\lfloor {}^{\tau}_{k}\breve{\mathbf{R}}^{k}_{G}\mathbf{R}^{\star}{}^G\mathbf{g} \rfloor \mathbf{J}_q\left(\tau\right)
\end{bmatrix}  \notag\\[4pt]
\begin{bmatrix}
\mathbf{H}_{\alpha}\left({\tau+1}\right) \\[3pt]
\mathbf{H}_{\beta}\left({\tau+1}\right)
\end{bmatrix}
&= \begin{bmatrix}
\mathbf{H}_{\alpha}\left({\tau}\right) + \mathbf{H}_{\beta}\left({\tau}\right) \Delta t -\mathbf{A}_{\tau} \\[3pt]
\mathbf{H}_{\beta}\left({\tau}\right) -\mathbf{B}_{\tau} 
\end{bmatrix} \\[4pt]
\begin{bmatrix}
\mathbf{O}_{\alpha}\left({\tau+1}\right) \\[3pt]
\mathbf{O}_{\beta}\left({\tau+1}\right)
\end{bmatrix}
&= \begin{bmatrix}
\mathbf{O}_{\alpha}\left({\tau}\right) + \mathbf{O}_{\beta}\left({\tau}\right) \Delta t \\[3pt]
\mathbf{O}_{\beta}\left({\tau}\right)
\end{bmatrix} \label{eq::m3_jac_init_ori}\\
&\hspace{0.4cm}-\begin{bmatrix} \mathbf{A}_{\tau} {}^{\tau}_{k}\breve{\mathbf{R}} \lfloor {}^k_G\mathbf{R}{}^{\star}{}^G\mathbf{g} \rfloor  \\[3pt]
\mathbf{B}_{\tau}{}^{\tau}_{k}\breve{\mathbf{R}} \lfloor {}^k_G\mathbf{R}^{\star}{}^G\mathbf{g} \rfloor
\end{bmatrix} \notag 
\end{align}
We note that Equation \eqref{eq::m3_jac_init_ori} reveals that only changes in the initial orientation \textit{perpendicular} to local gravity (${}^k\mathbf{g}$) will cause a change in the preintegrated measurement. As these directions of orientation are observable and thus are expected to have small errors, this highlights the fact that our linearization scheme about the initial orientation is appropriate.
}
At this point, using these Jacobians, we can write
the residual associated with the new preintegrated IMU measurement as follows:
\begin{align}  \label{eq:model2-res}
&\scalemath{0.95}{\mathbf{e}_{IMU}(\mathbf{x})=} \\
&~~~~\scalemath{0.95}{
 \begin{bmatrix}
2\mathbf{vec}\left({}^{k+1}_G\bar{q} \otimes  {}^{k}_G\bar{q}^{-1} \otimes  {}^{k+1}_k\breve{\bar{q}}^{-1} \otimes \bar{q}_b \right) \\[4pt] \hdashline[2pt/2pt] \\[-8pt]
\mathbf{b}_{{\omega}_{k+1}}- \mathbf{b}_{{\omega}_{k}} \\[4pt] \hdashline[2pt/2pt] \\[-8pt]
\Bigg({}^k_G\mathbf{R}\left({}^G\mathbf{v}_{k+1}-{}^G\mathbf{v}_{k} \right)  -\mathbf{J}_\beta \left(\mathbf{b}_{\omega_k}- \mathbf{b}^\star_{\omega_k} \right)-   \\
\mathbf{H}_\beta \left(\mathbf{b}_{a_k}- \mathbf{b}^\star_{{a_k}} \right)- \mathbf{O}_\beta~2\mathbf{vec}\left({}^k_G \bar{q} \otimes {}^k_G \bar{q}^{\star-1} \right)-\Delta \breve{v} \Bigg) \\[4pt] \hdashline[2pt/2pt] \\[-8pt]
\mathbf{b}_{{a}_{k+1}}- \mathbf{b}_{{a}_{k}} \\[4pt] \hdashline[2pt/2pt] \\[-8pt]
\Bigg({}^k_G\mathbf{R}\left({}^G\mathbf{p}_{k+1}-{}^G\mathbf{p}_{k}- {}^G\mathbf{v}_{k}\Delta T \right) -\mathbf{J}_\alpha \left(\mathbf{b}_{\omega_k}- \mathbf{b}^\star_{\omega_k} \right)-  \\
\mathbf{H}_\alpha \left(\mathbf{b}_{a_k}- \mathbf{b}^\star_{a_k} \right)- \mathbf{O}_\alpha~2\mathbf{vec}\left({}^k_G \bar{q} \otimes {}^k_G \bar{q}^{\star-1} \right) -\Delta \breve{p} \Bigg)
\end{bmatrix}  \notag
}
\end{align}
The resulting measurement Jacobians are necessary for an iterative solver,
which we analytically compute as shown in Appendix B.2.
 \section{Visual-Inertial Navigation}\label{sec:vins}

To demonstrate the applicability of the proposed {closed-form preintegration (CPI)} theory presented in the preceding section,
in this section, we develop two {sliding-window optimization-based} sensor fusion schemes for visual-inertial navigation systems (VINS) 
that utilize our inertial preintegration.

\subsection{Tightly-Coupled Indirect VIO} \label{sec:vio}

As an IMU-camera sensor suite moves through an unknown environment, 
visual feature keypoints can be extracted and tracked from the images to provide motion information about the platform.
In particular, the measurement function that maps the 3D position, ${}^G\mathbf{p}_f$, of a feature into the normalized uv-coordinates on the $j$-th camera's image plane at time step~$k$ takes the following form:
\begin{align}
    \mathbf{z}_{fjk}= \bm \Pi \left({}_I^{C_j}\mathbf{R}~{}_G^{k}\mathbf{R}\left({}^G\mathbf{p}_f -{}^G\mathbf{p}_{k} \right) + {}^{C_j} \mathbf{p}_I \right)+ \mathbf{n}_f \label{eq::normalized}
\end{align}
where ${}_I^{C_j}\mathbf{R}$ and ${}^{C_j} \mathbf{p}_I$ are the rigid IMU-to-camera extrinsic calibration parameters, $\mathbf{n}_f \sim \mathcal{N}(\mathbf{0}, \bm\Lambda_{fjk}^{-1} )$,
and $\bm \Pi(\cdot)$ is the perspective projection function given by~\citep{Hartley2000}:
\begin{align} \label{eq:perspective-proj}
    \bm \Pi \left( \begin{bmatrix} x \\ y \\ z \end{bmatrix} \right) = \begin{bmatrix} x/z \\ y/z \end{bmatrix}
\end{align}
The error (or residual) associated with this visual measurement is given by:
\begin{align}\label{eq:e_fjk}
    \mathbf{e}_{fjk}(\mathbf{x}) = \bm \Pi \left({}_I^{C_j}\mathbf{R}~{}_G^{k}\mathbf{R}\left({}^G\mathbf{p}_f -{}^G\mathbf{p}_k \right) + {}^{C_j} \mathbf{p}_I \right) -\mathbf{z}_{fjk} 
\end{align}
{Using all these visual measurements available in a sliding window} 
along with the preintegrated IMU measurements and marginalization prior, 
we solve the following optimization problem that tightly couples all available measurement residuals:
\begin{align} \label{opt:vio}
    \hat{\mathbf{x}} = &\mathop{\mathrm{argmin}}_{\mathbf{x}} \Big( 
    \left| \left|\mathbf{e}_{marg} \left(\mathbf{x} \right) \right| \right|_2^2+ \sum_{p \in \mathcal{P}} \left| \left| \mathbf{e}_{IMU} \left(\mathbf{x} \right) \right| \right|_{\mathbf{P}_p^{-1}}^2  \notag \\
    &\hspace{2cm} +\sum_{(f,j,k) \in \mathcal{C}} \left| \left| \mathbf{e}_{fjk}\left(\mathbf{x} \right) \right| \right|_{\bm \Lambda_{fjk}}^2 \Big) 
\end{align}
where $\mathcal{C}$ and $\mathcal{P}$ are the set of feature and preintegrated measurements, respectively, 
while $\mathbf{e}_{marg} \left(\mathbf{x} \right)$ is the residual of the marginal prior (see Equation \eqref{eq:margresidual}).
{We want to point out again that in practice we instead employ a (Huber or Cauchy) robust cost function on the last visual error term in Equation~\eqref{opt:vio}, while we here omit the detailed derivations of this standard treatment to keep  presentation concise, we do have a similar treatment in our ensuing loosely-coupled direct VINS (see Equation~\eqref{opt:direct-align}).}

\subsubsection{Inverse-depth representation:}

A well-known disadvantage of the above representation for features is that points at infinity are difficult to utilize.
To mitigate this issue, we instead employ an \textit{inverse-depth} representation~\citep*{Civera2008TRO}.
In particular, we represent a feature using the inverse coordinates in the camera frame where it was first observed.
Denoting $\{C_{a,i}\}$ the frame of reference of the ``anchoring'' camera, 
which is associated with the $i$-th camera frame and the anchoring time $a$, 
we have the following inverse-depth representation (see \citet*{Mourikis2007ICRA}):
\begin{align}
{}^{C_{a,i}}\mathbf{m}_{f} = \begin{bmatrix} \alpha \\ \beta \\ \rho \end{bmatrix} ~~~\Rightarrow~~~
{}^{C_{a,i}}\mathbf{p}_{f} = \frac{1}{\rho}\begin{bmatrix} \alpha \\ \beta \\ 1 \end{bmatrix} 
\end{align}
where we  also show the relationship between the inverse-depth representation of the feature ${}^{C_{a,i}}\mathbf{m}_{f}$ 
and the corresponding 3D position in the anchor frame ${}^{C_{a,i}}\mathbf{p}_{f}$.
The feature position in the $j$-th  camera frame at time step $k$ can be computed as follows:
\begin{align}
{}^{C_{k,j}} \mathbf{p}_f &= {}^{C_{k,j}}_{C_{a,i}} \mathbf{R}  \frac{1}{\rho}\begin{bmatrix} \alpha \\ \beta \\ 1 \end{bmatrix} + {}^{C_{k,j}}\mathbf{p}_{C_{a,i}} \label{eq:project}\\
{}^{C_{k,j}}_{C_{a,i}} \mathbf{R}&= {}^{C_j}_{I} \mathbf{R}~{}^{k}_{G}\mathbf{R}~{}^{a}_{G}\mathbf{R}^{\top}{}^{C_i}_{I} \mathbf{R}^{\top} \notag\\
{}^{C_{k,j}}\mathbf{p}_{C_{a,i}}& =  {}^{C_j}_{I} \mathbf{R}~{}^{k}_{G} \mathbf{R}\big({}^G\mathbf{p}_{a}+ {}_G^{a}\mathbf{R}^{\top} {}^{I}\mathbf{p}_{C_i}-{}^G\mathbf{p}_{k}\big)+ {}^{C_j}\mathbf{p}_I \notag
\end{align}
Note that due to the {projective geometry} of the perspective projection~\eqref{eq:perspective-proj}, $\bm \Pi(\mathbf{x})= \bm \Pi(\rho\mathbf{x})$,
 we can multiply both sides of Equation \eqref{eq:project} by $\rho$ and have the equivalent measurement model:
\begin{align}
\mathbf{z}_{fjk}&= \bm \Pi \left( \mathbf h\right) + \mathbf{n}_f \label{eq:z_fjk}\\
\mathbf h = \begin{bmatrix} h_1 \\h_2 \\ h_3 \end{bmatrix} &:= \rho {}^{C_{k,j}} \mathbf{p}_f  = {}^{C_{k,j}}_{C_{a,i}} \mathbf{R}  \begin{bmatrix} \alpha \\ \beta \\ 1 \end{bmatrix} + \rho{}^{C_{k,j}}\mathbf{p}_{C_{a,i}} 
\label{eq:inverse-depth-meas-h}
\end{align}
The measurement Jacobians of this inverse-depth model can be found in Appendix C.
Note that this measurement model is numerically stable and can handle points at infinity, 
thus allowing for the gain of feature direction information from these far-off feature points.
 \subsection{Loosely-Coupled Direct VINS}

\begin{figure}
\centering
\includegraphics[width=0.48\textwidth]{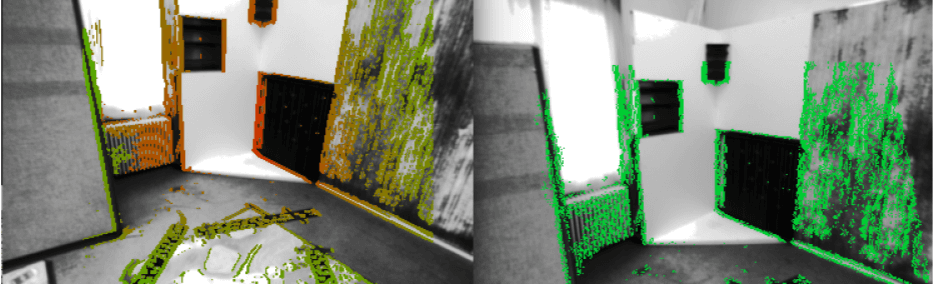}
\caption{Visualization of selected depth map pixels with a large intensity gradient (left).
Keyframe pixels are projected onto the query frame as a result of the optimized direct alignment of the frame-to-frame relative transformation (right).}
\label{fig:directviz}
\end{figure}

To further validate the proposed {closed-form preintegration} theory,
in the following, by leveraging our prior work~\citep*{Eckenhoff2017ICRA},
we develop a \textit{loosely-coupled} VINS algorithm based on direct image alignment and IMU preintegration.
In particular, we estimate the relative frame-to-frame motion through direct alignment of image pixels. 
These relative-motion constraints then allow us to efficiently perform loop closure without 
explicitly detecting/tracking (or matching) features.

Consider the case where we wish to directly align a current frame $C_2$ against a keyframe $C_1$ (see Figure \ref{fig:directviz}).
Finding the optimal transformation can be formulated as an optimization problem over the total (warped) pixel intensity difference (i.e., photometric error):
\begin{align} \label{opt:direct-align}
\scalemath{0.9} {
    {}_{C_1}^{C_2} \breve{\mathbf{T}} = \mathop{\mathrm{argmin}}_{ {}_{C_1}^{C_2}  {\mathbf{T}}} \sum_{f} \gamma \left( 
    \underbrace{
    \frac{1}{\sigma_r^2} 
    \left(  
    \underbrace{I_{C_2} \left( {}_{C_1}^{C_2}  {\mathbf{T}}~ {{}^{C_1}\mathbf{p}_f} \right) - I_{C_1} \left({}^{C_1}\mathbf{p}_f \right) }_{e_f}
    \right)^2 
    }_{v_f}
    \right) 
}
\end{align}
where ${}_{C_1}^{C_2}{\mathbf{T}}$ is the transformation between the two camera frames parameterized by the relative quaternion ${}_{C_1}^{C_2} \bar{q}$ and relative position ${}^{C_1} \mathbf{p}_{C_2}$, while $I_{C_i}(\cdot)$ returns the intensity of a given point projected into the image frame, and $\gamma(\cdot)$ is the Huber cost.
The pixel's position in the keyframe, ${}^{C_1} \mathbf{p}_f$ can be found via an online or stereo pair depth map computation.
This position is treated as a noisy parameter in the residual allowing for computation of the residual sigma, $\sigma_r$, with the summation being over all pixels $f$ with valid depth estimates and high gradients along the epipolar line.
The Huber cost function $\gamma(\cdot)$ with parameter $k$ is defined as \citep*{Eade2013TR}:
\begin{align}
  \gamma(r)= 
\begin{cases}
    r,& \text{if } r < k^2\\
    2k\sqrt{r}-k^2,      & \textrm{otherwise}
\end{cases}
\end{align}
The purpose of the Huber cost is to down-weight large residuals which occur naturally in image alignment due to occlusions, and has been used extensively in the literature (e.g., \citet*{Engel2014ECCV}).

Note that the covariance of each residual $\sigma_r^2$ encodes the uncertainty due to errors in the intensity measurements as well as the disparity map:
\begin{align}
\sigma_r^2= 2\sigma_{int}^2+\Big(\frac{\partial e_f}{\partial d}\Big)^2\sigma_d^2
\end{align}
 where $\sigma_{int}^2$ denotes the covariance of the intensity reading, $\frac{\partial e_f}{\partial d}$ is the Jacobian of the residual $e_f$~\eqref{opt:direct-align} with respect to the measured disparity $d$, and $\sigma_d^2$ is the covariance of the disparity measurement.
In the case of a depth map computed from a stereo pair as considered in this work, 
we define $\mathbf{t}$ as the pixel coordinates, $z$ as the pixel depth, and $b$ as the baseline between the stereo pair.
The Jacobian $\frac{\partial e_f}{\partial d}$ can be calculated using the chain rule of differentiation as follows (see Equation~\eqref{opt:direct-align}):
\begin{align}
\frac{\partial e_f}{\partial d} =& \frac{\partial I_{C_2}}{\partial \mathbf{t}}\frac{\partial \mathbf{t}}{\partial {}^{C_2} \mathbf{p}_f}
\frac{\partial {}^{C_2} \mathbf{p}_f}{\partial {}^{C_1} \mathbf{p}_f}\frac{\partial {}^{C_1} \mathbf{p}_f}{\partial z}\frac{\partial z}{\partial d}  \notag\\
=&
\begin{bmatrix} I_{C_{2_x}} &  I_{C_{2_y}} \end{bmatrix}
\begin{bmatrix} \frac{f_x}{^{C_2}\mathbf{p}_{f_j}(3)} & 0 &
-\frac{f_x{}^{C_2}\mathbf{p}_{f_j}(1)}{^{C_2}\mathbf{p}_{f_j}(3)^2} \\
0 & \frac{f_y}{^{C_2}\mathbf{p}_{f_j}(3)} &
-\frac{f_y{}^{C_2}\mathbf{p}_{f_j}(2)}{^{C_2}\mathbf{p}_{f_j}(3)^2}\end{bmatrix} \notag\\
&\times~ {}^{C_2}_{C_1}\mathbf{R}\frac{^{C_1}\mathbf{p}_{f_j}}{z}\frac{-f_x b}{d^2} 
\end{align}
where
$I_{C_{2_x}}$ and $I_{C_{2_y}}$ are the image gradients in the $x$ and $y$ directions respectively, while $f_x$ and $f_y$ are the focal lengths of the camera. 

The covariance of the pixel disparity, $\sigma_d^2$, is obtained based on the observation that 
this disparity is the maximum likelihood estimate for a single measurement graph, with the residual being the difference in intensity between the pixel in the left, $I_{C_1L}$, and right, $I_{C_1R}$, images in the keyframe stereo pair, which can be formulated as follows:
{\begin{align}
\breve{d} = \mathop{\mathrm{argmin}}_d 
\frac{1}{\sigma_{rd}^2} \Big( \underbrace{ I_{C_1L}(v,u)- I_{C_1R}(v,u-d)}_{e_d} \Big)^2
\end{align}}where the covariance associated with this residual can be found as $\sigma_{rd}^2= 2\sigma_{int}^2$, and
comes from uncertainty in the intensity readings.
The covariance on our disparity estimate can then be approximated as:
\begin{align}
\sigma_{d}^2= \left(\frac{\partial e_d}{\partial d}^2\frac{1}{\sigma_{rd}^2} \right)^{-1} 
= \sigma_{rd}^2\left(\frac{1}{I_{C_1R_x}}\right)^2
\end{align}
where $I_{C_1R_x}$ is the $x$-gradient of the pixel in the right image which is selected as the match.

{Once we have determined the photometric error covariance $\sigma_r^2$,}
we now solve the direct alignment problem~\eqref{opt:direct-align} using the Levenberg-Marquadt method.
In particular, at each iteration we solve the following normal equation:
\begin{align}
    &\left(\left(\sum w_f \mathbf{J}_f^{\top}\mathbf{J}_f\right) + \lambda \textrm{diag}\left(\sum w_f \mathbf{J}_f^{\top}\mathbf{J}_f\right) \right) \delta ~{}_{C_1}^{C_2}{\mathbf{T}} \notag\\
    &= -\sum  w_f \mathbf{J}_f^{\top} e_f\left( {}_{C_1}^{C_2}\breve{\mathbf{T}} \right) 
\end{align}
where $\lambda$ is the damping parameter, and $e_f({}_{C_1}^{C_2}\breve{\mathbf{T}})$ is the residual due to the $f$-th pixel in the alignment, evaluated at the current estimate (linearization point) for the relative transformation, ${}_{C_1}^{C_2}\breve{\mathbf{T}}$.
The weight $w_f$ is computed at each iteration as follows:
\begin{align}
w_f
&= \frac{\partial \gamma(v_f)}{\partial v_f}\frac{1}{\sigma_f^2} \\
\frac{\partial \gamma(v_f)}{\partial v_f}
&= \begin{cases}
1,& \text{if } v_f < k^2\\
\frac{k}{\sqrt{v_f}},      & \text{otherwise}
\end{cases}
\end{align}
where $v_f$ is the raw cost fed into the Huber norm (see Equation~\eqref{opt:direct-align}), and $k$ is a design parameter. 
The Jacobian matrix $\mathbf{J}_f$ is of the direct alignment measurement residual with respect to the state, computed as:
\begin{align}
\mathbf{J}_f&=
\begin{bmatrix} I_{C_{2_x}} &  I_{C_{2_y}} \end{bmatrix} 
\begin{bmatrix} \frac{f_x}{^{C_2}\mathbf{p}_{f}(3)} & 0 &
-\frac{f_x{}^{C_2}\mathbf{p}_{f}(1)}{(^{C_2}\mathbf{p}_{f}(3))^2} \\
0 & \frac{f_y}{^{C_2}\mathbf{p}_{f}(3)} &
-\frac{f_y{}^{C_2}\mathbf{p}_{f}(2)}{(^{C_2}\mathbf{p}_{f}(3))^2}\end{bmatrix} \notag\\[8pt]
&~~~~\times\begin{bmatrix} \lfloor {}^{C_2} \mathbf{p}_{f} \rfloor & -{}^{C_2}_{C_1} \mathbf{R} \end{bmatrix}
\end{align}

After optimization, we will be left with a Gaussian distribution on our estimated relative \textit{camera} pose.
We can then transform this into a distribution on the relative {\em IMU} pose (denoted $k$ and $j$ for the keyframe and query frame IMU states respectively) using covariance propagation:
\begin{align}
    {}_{C_1}^{C_2}\mathbf{T} &= {}_{C_1}^{C_2}\breve{\mathbf{T}}\boxplus {}_{C_1}^{C_2}\delta{\mathbf{T}}, ~~
    \mathrm{where}~~{}_{C_1}^{C_2}\delta{\mathbf{T}} \sim \mathcal{N}\left(\mathbf{0}, \bm \Sigma_c  \right) \\[10pt]
    {}_{k}^{j}\mathbf{T} &= {}_{k}^{j}\breve{\mathbf{T}}\boxplus {}_{k}^{j}\delta{\mathbf{T}}, ~~
    \mathrm{where}~~{}_{k}^{j}\delta{\mathbf{T}} \sim \mathcal{N}\left(\mathbf{0}, \bm \Sigma_i  \right) \\
        \bm \Sigma_i &=\frac{\partial {}_{k}^{j}\delta{\mathbf{T}}}{\partial{}_{C_1}^{C_2}\delta{\mathbf{T}}} \bm \Sigma_c\frac{\partial {}_{k}^{j}\delta{\mathbf{T}}}{\partial{}_{C_1}^{C_2}\delta{\mathbf{T}}}^{\top}
\end{align}
where $\bm \Sigma_c = (\sum w_f \mathbf{J}_f^{\top}\mathbf{J}_f)^{-1} $ is the covariance of the zero-mean alignment error. 
From this, the relative pose measurement that connects the IMU keyframe and query frame has the following residual:
\begin{align} \label{eq:direct-res}
\mathbf{e}_{d}(\mathbf{x}) = \begin{bmatrix} 2\mathbf{vec}\left({}_G^{j}\bar{q} \otimes {}_G^{k}\bar{q}^{-1} \otimes {}_{k}^{j}\breve{\bar{q}}^{-1} \right) \\[8pt]
{}_G^{k} \mathbf{R}\left({}^G \mathbf{p}_{j}-{}^G \mathbf{p}_{k} \right) - {}^k\breve{\mathbf{p}}_{j} \end{bmatrix}
\end{align}
whose Jacobians with respect to the state are provided in Appendix D,
which will be used during graph optimization. 

Using this visual measurement residual, along with the preintegrated IMU measurements,
we have the following optimization problem for the loosely-coupled direct VINS, 
which can be solved analogously as in Equation~\eqref{opt:vio}:
\begin{align}
    \scalemath{0.95}{
    \hat{\mathbf{x}} = \mathop{\mathrm{argmin}}_{\mathbf{x}} \sum_{d \in \mathcal{D}} \left| \left| \mathbf{e}_{d} \left(\mathbf{x} \right) \right| \right|_{\bm \Sigma_i^{-1}} ^2+ \sum_{p \in \mathcal{P}} \left| \left| \mathbf{e}_{IMU} \left(\mathbf{x} \right) \right| \right|_{\mathbf{P}_p^{-1}}^2
    }
    \label{opt:direct-vins}
\end{align}
where $\mathcal{D}$ and $\mathcal{P}$ are the set of direct alignment relative pose and preintegrated measurements, respectively.
Note that 
this direct image alignment allows for computationally efficient incorporation of large-scale loop closures 
due to the direct compression of intensity residuals into a single informative relative motion measurement.
 \section{Monte-Carlo Simulation Analysis} \label{sec:sim}

\begin{figure}    \centering
    \includegraphics[width=0.48\textwidth]{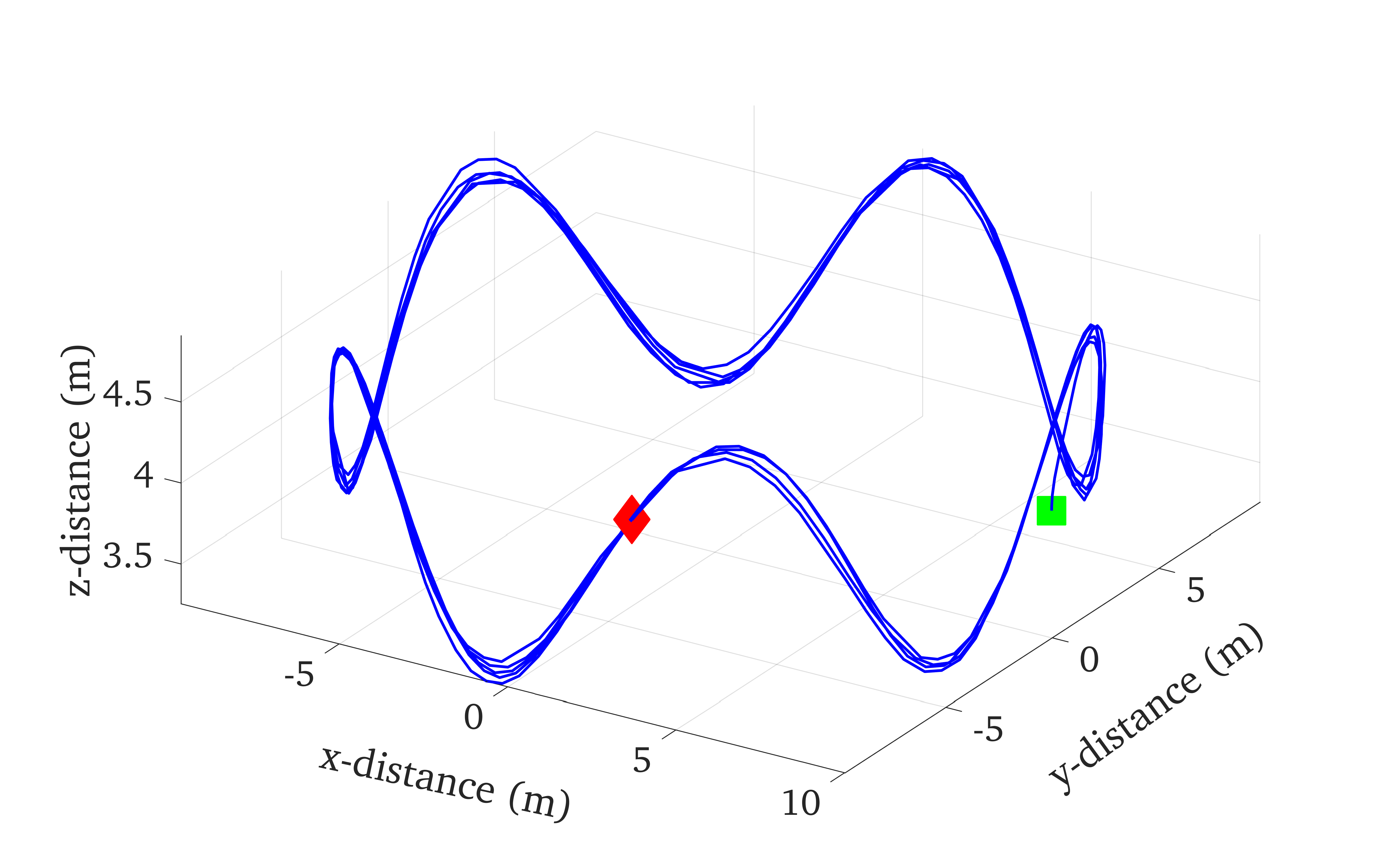}
    \caption{The ground truth trajectory of a MAV flying in a circle sinusoidal path generated in the Gazebo simulator. The total trajectory length is 307 meters with an average velocity of 6.13 m/s. Start and end positions are denoted with a green square and red diamond, respectively.}
    \label{fig:pathsim}
\end{figure}

To validate the proposed {closed-form preintegration} theory, 
we first perform extensive Monte-Carlo simulations in various conditions in terms of sampling rates and motion dynamics.
In particular, to better model the motion dynamics of a physical system, we leverage the open-source Gazebo simulator of a micro air vehicle (MAV) \citep*{Koenig2004IROS} which allows for direct realistic  simulation and collection of true inertial and pose data (constrained by the physical MAV motion).
The simulated datasets were generated as follows: (i) the MAV was commanded to follow a series of waypoints after takeoff, (ii) the ground truth of 100 Hz inertial and pose information was recorded, (iii) 80 synthetic stereo visual feature measurements (uv-coordinates) were created for each camera frame using the true pose information at a static 10Hz frequency.
Following the commonly-used IMU model~\citep*{Trawny2005_Q_TR}, the true inertial measurements were corrupted with an additive discrete bias and white noise using the noise parameters from the VI-Sensor~\citep*{Nikolic2014ICRA}, while the features' uv-coordinates were corrupted with an additive white noise to each axis with one pixel standard deviation.

\begin{table} \caption{Analysis of the effect of different IMU frequencies on estimation accuracy. 
Note that each frequency run has a slightly different trajectory, and thus only the relative spread within a given frequency should be considered.}
\begin{center}
\scalebox{1.0}{
\begin{tabular}{|c|c|c|c|c|c|c|} \hline
    & \multicolumn{2}{|c|}{\textbf{MODEL-1}} & \multicolumn{2}{|c|}{\textbf{MODEL-2}} & \multicolumn{2}{|c|}{\textbf{DISCRETE}} \\\hline
    Units & m & deg & m & deg & m & deg \\\hline
    100 Hz & 0.096 & 0.327 & \ws{0.093} & \ws{0.300} & 0.107 & 0.328 \\\hline
    200 Hz & 0.051 & 0.204 & \ws{0.049} & \ws{0.179} & 0.058 & 0.207 \\\hline
    400 Hz & 0.033 & 0.107 & \ws{0.033} & \ws{0.101} & 0.035 & 0.109 \\\hline
    800 Hz & \ws{0.030} & 0.085 & 0.030 & \ws{0.085} & 0.031 & 0.086 \\\hline
\end{tabular}
}
\end{center}
\label{table:imu-freq}
\end{table}

In our tests, 
we used the popular GTSAM \citep*{Dellaert2012tr} framework to construct, optimize, and marginalize our graph using the included fixed-lag smoother.
To ensure a fair comparison, we evaluate our preintegration methods against the state-of-art {\bf discrete} preintegration~\citep*{Forster2015RSS,Forster2017TRO}, by using the on-manifold preintegrator class within the GTSAM implementation to compute the required measurement means, bias Jacobians, and covariances.
We constructed all graphs side by side, ensuring that the measurements inserted are exactly the same, and thus fair to all methods.
For simplicity we used the tightly-coupled indirect features in the graph, in which features are automatically marginalized out after three seconds (that is, no map was created for loop closures, and thus, the system is a VIO system).
Note also that we initialized all systems to the ground-truth pose, and with zero bias.
Figure~\ref{fig:pathsim} shows one example of the true simulated trajectory generated using Gazebo in our simulations. 

\begin{figure*}    \centering
    \begin{subfigure}[b]{\textwidth}
        \centering
        \includegraphics[width=.99\textwidth]{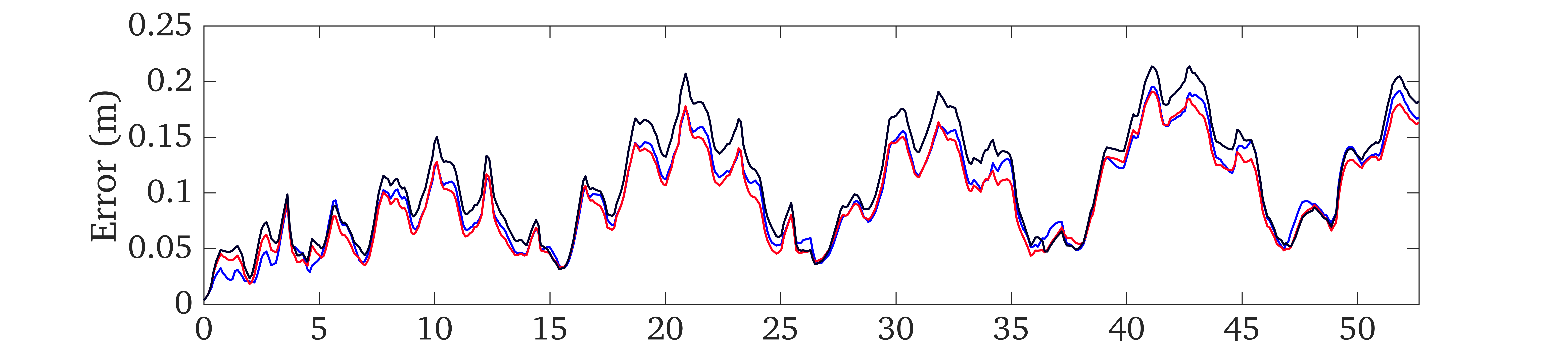}
    \end{subfigure}        \vspace{1em}
    \begin{subfigure}[b]{\textwidth}
        \centering
        \includegraphics[width=.99\textwidth]{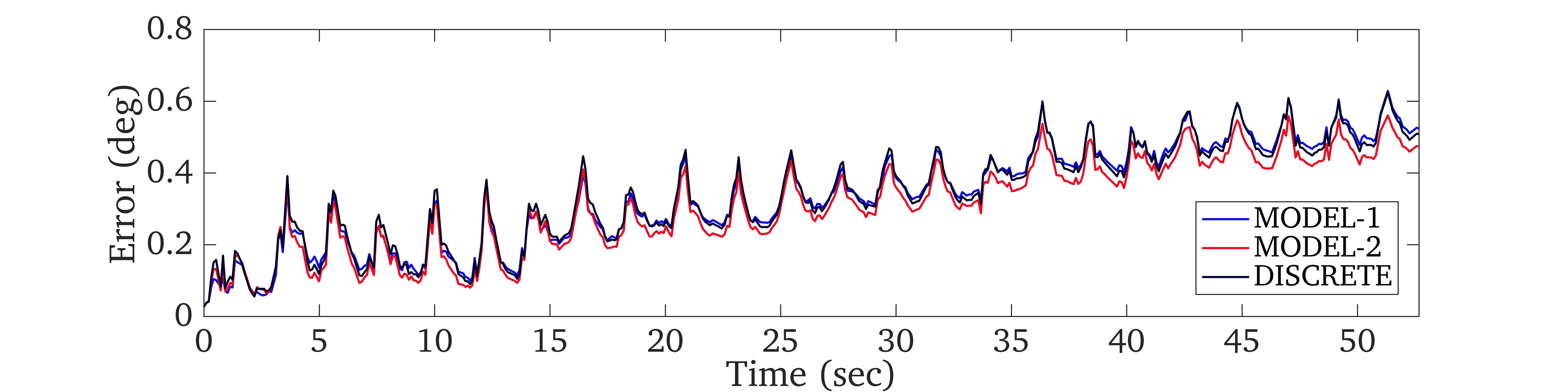}
    \end{subfigure}
        \caption{Monte-Carlo simulation results averaged over 50 runs: (top) position RMSE, and (bottom) orientation RMSE. In this test, physically-realistic synthetic data was generated using a Gazebo MAV simulator. It is clear that the proposed closed-form preintegration outperforms the state-of-the-art discrete approach~\citep*{Forster2015RSS,Forster2017TRO}.     }
    \label{fig:rmsesim}
\end{figure*}

The Monte-Carlo simulation comparison results of root mean squared error (RMSE) averaged over 50 runs are shown in  Figure~\ref{fig:rmsesim}. 
Evidently, the proposed preintegration using piecewise constant local acceleration model (i.e., {\bf Model 2}) is slightly better  with the RMSE (averaged over all time steps and all runs) of 0.093 meters and 0.300 degrees 
than that using the piecewise constant measurement model (i.e., {\bf Model 1}) with the RMSE of 0.096 meters and 0.327 degrees.
More importantly, both methods are shown to outperform the discrete state-of-the-art method~\citep*{Forster2015RSS,Forster2017TRO}, 
which has the RMSE of 0.107 meters and 0.328 degrees.
It is important to point out that the superior performance (though by a small margin in this MAV test, with larger improvement margins expected for higher dynamics not constrained by MAV motion) endowed by the proposed closed-form preintegration using the new inertial models over the discrete one does not incur extra computational overhead during graph-based VINS optimization.

Furthermore, we investigate the effect that the IMU frequency has on the relative performance of the preintegration methods under consideration.
Using the same simulation setup, the MAV was commanded to follow the trajectory with different Gazebo simulation frequencies.
It is important to note that since we are using a physical simulation, the true trajectory will vary from frequency to frequency since the controller will perform differently, however, this is acceptable since we are looking at the relative performance within a given frequency.
Table~\ref{table:imu-freq} shows the averaged RMSE results of different IMU frequencies.
It can be seen that the proposed closed-form preintegration methods have greater impact when the frequency of the IMU is lower (i.e., significantly better performance); while at higher frequencies, the above methods become less distinguishable.
This implies that the proposed closed-form preintegration \color{black} methods are better suited for applications with limited IMU frequency, which is often the case for low-cost MEMS sensors.

 \section{Real-World Experimental Validations} \label{sec:exp1}

\subsection{Tightly-Coupled Indirect VIO}

In our tightly-coupled VIO system, we use stereo vision due to its superior estimation performance as compared to monocular systems \citep*{Paul2017ICRA}. Stereo correspondences allow for accurate triangulation of features regardless of vehicle motion, making them robust to maneuvers  troubling monocular systems such as hovering. In addition, stereo allows for a direct reading of the scale, which is highly informative to the estimator.

When a pair of stereo images arrive, we perform KLT tracking \citep*{Baker2004IJCV} of FAST \citep*{Rosten2010PAMI} features that have been extracted in an uniform grid over the image.
Stereo correspondence information is known by initializing new features in the left image and KLT tracking them into the right.
The set of stereo tracks from the current image is then tracked temporally forward at each future time step, while also ensuring to initialize new feature tracks if the number of active tracks falls under our desired active feature threshold.
To reject outliers we perform 8-point RANSAC between both the temporal and stereo left-to-right matches.
We have found that this frontend provides a good balance between track longevity, computational speed, and accuracy.
If an active feature is successfully tracked, the normalized image coordinates are added as measurements associated with that feature.
{To robustify our system to outliers, we utilized the Cauchy loss function for all image measurements.}

Inspired by \cite*{LeuteneggerIJRR2014}, we maintain a sliding window of IMU states in the estimator that consists of two sub-windows. The first, denoted as the inertial window, contains the full 15 DOF IMU state and refers to the most recent imaging times. The second window, called the pose window, contains a set of pose-only clones (that is, only the orientation and position are maintained).
At every imaging time we create a new corresponding IMU node. The IMU readings collected over the interval are preintegrated to both predict the new state and to form a preintegrated IMU measurement between the previous and new state. 

After tracking, we formulate the sliding-window batch optimization (i.e., BA) problem using all features with a sufficient number of tracks as well as all nodes in the inertial and pose windows. The measurements contained in this graph are: (i) the prior, (ii) the visual measurements for the active features, and (iii) the preintegration factors between the inertial window states (see Equation~\eqref{opt:vio}).
We use the Ceres Solver with an elimination ordering that takes advantage of the sparsity of the problem through the Schur Complement~\citep*{ceres-solver, Kummerle2011ICRA}.

If the inertial window has reached its maximum length, we flag the oldest state's velocity and biases for marginalization. If the pose window also reaches its maximum length, we add both the oldest pose and all features it has seen into the marginal state list.
Performing marginalization yields a new prior factor that has absorbed the old prior, the marginalized feature measurements, and the oldest preintegration factor. 
The oldest IMU state whose velocity and biases have been marginalized is then moved into the pose window. 
 \subsubsection{EuRoC MAV dataset:}

\begin{figure*}
    \centering
    \begin{subfigure}[b]{0.48\textwidth}
        \centering
        \includegraphics[height=5cm]{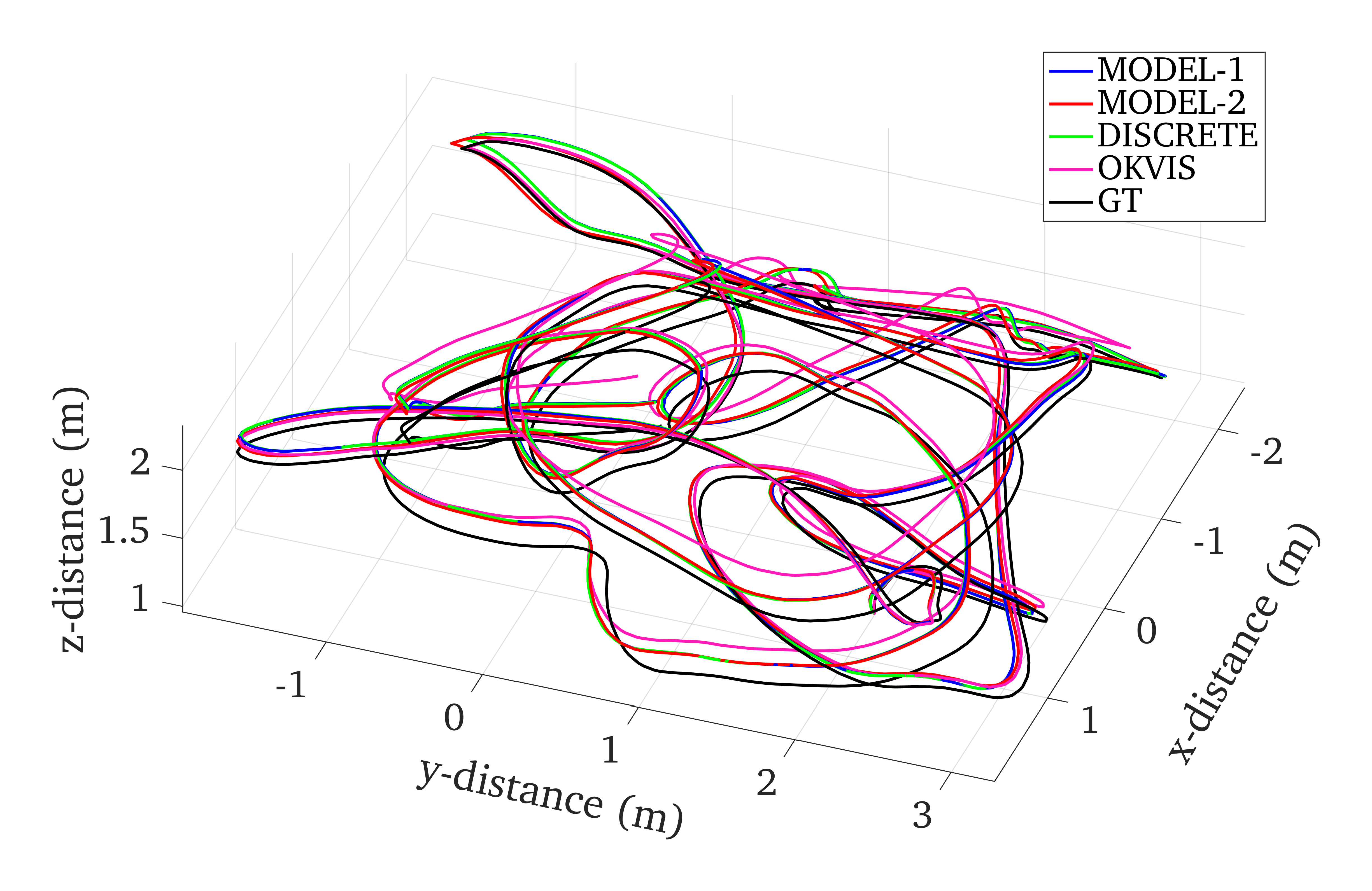}
        \caption{Average trajectory estimates for ``V1\_02\_med''.}
        \label{fig:patheth1}
    \end{subfigure}
    \hfill
    \begin{subfigure}[b]{0.48\textwidth}
        \centering
        \includegraphics[height=5cm]{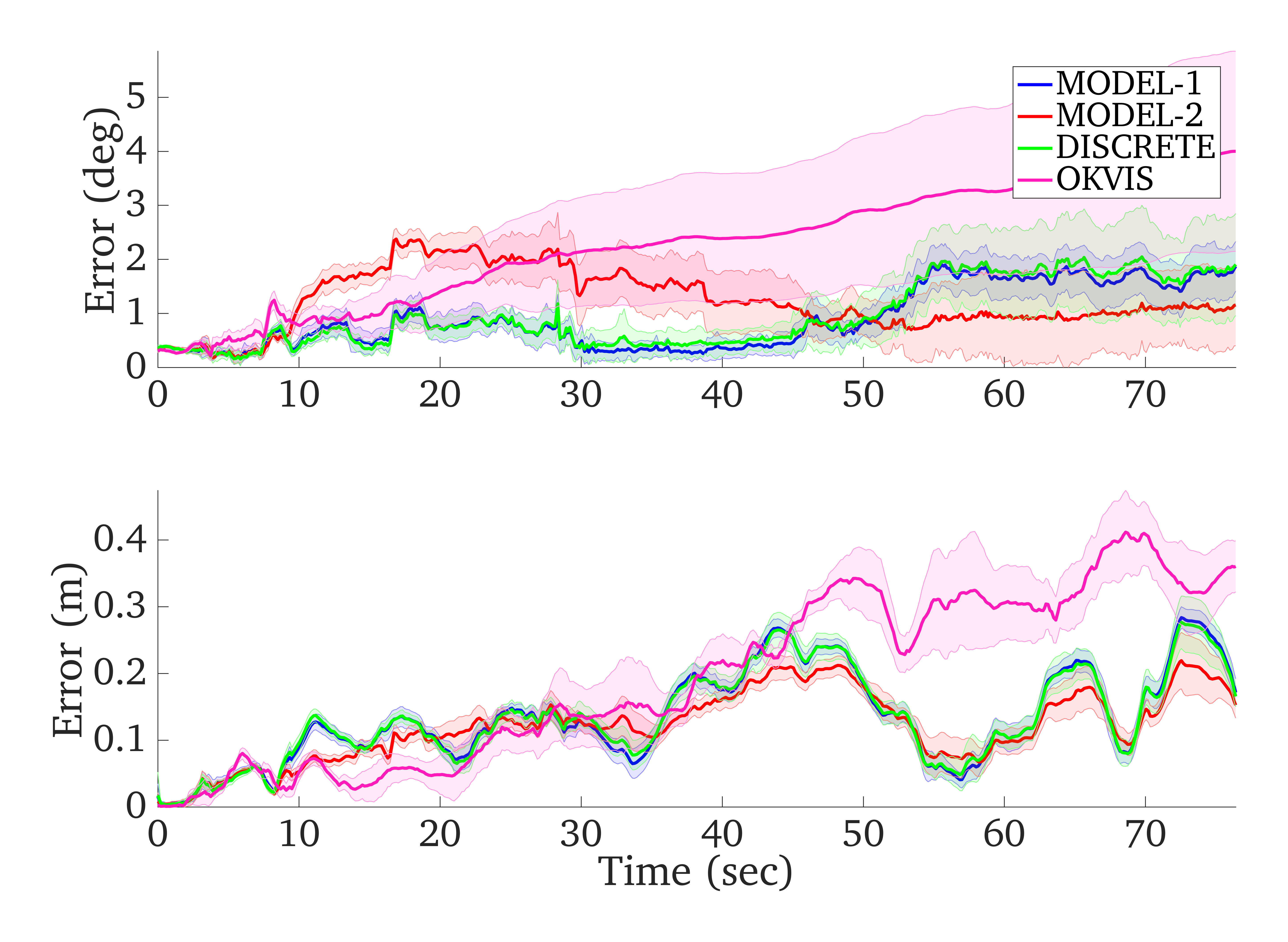}
        \caption{Average position and orientation RMSE for ``V1\_02\_med''.}
        \label{fig:rmseeth1}
    \end{subfigure}
    \newline
    \centering
    \begin{subfigure}[b]{0.48\textwidth}
        \centering
        \includegraphics[height=5cm]{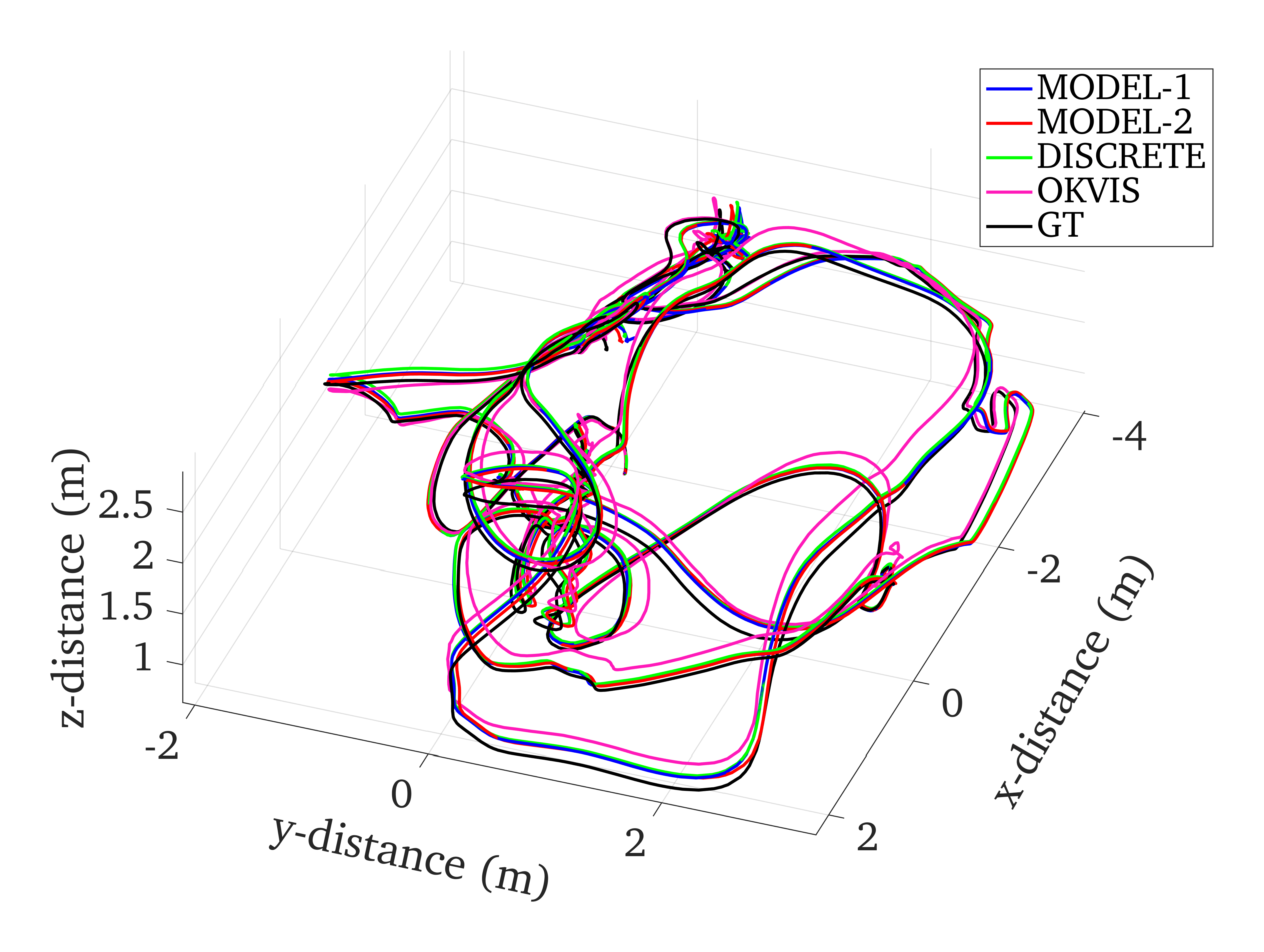}
        \caption{Average trajectory estimates for ``V2\_02\_med''.}
        \label{fig:patheth2}
    \end{subfigure}
    \hfill
    \begin{subfigure}[b]{0.48\textwidth}
        \centering
        \includegraphics[height=5cm]{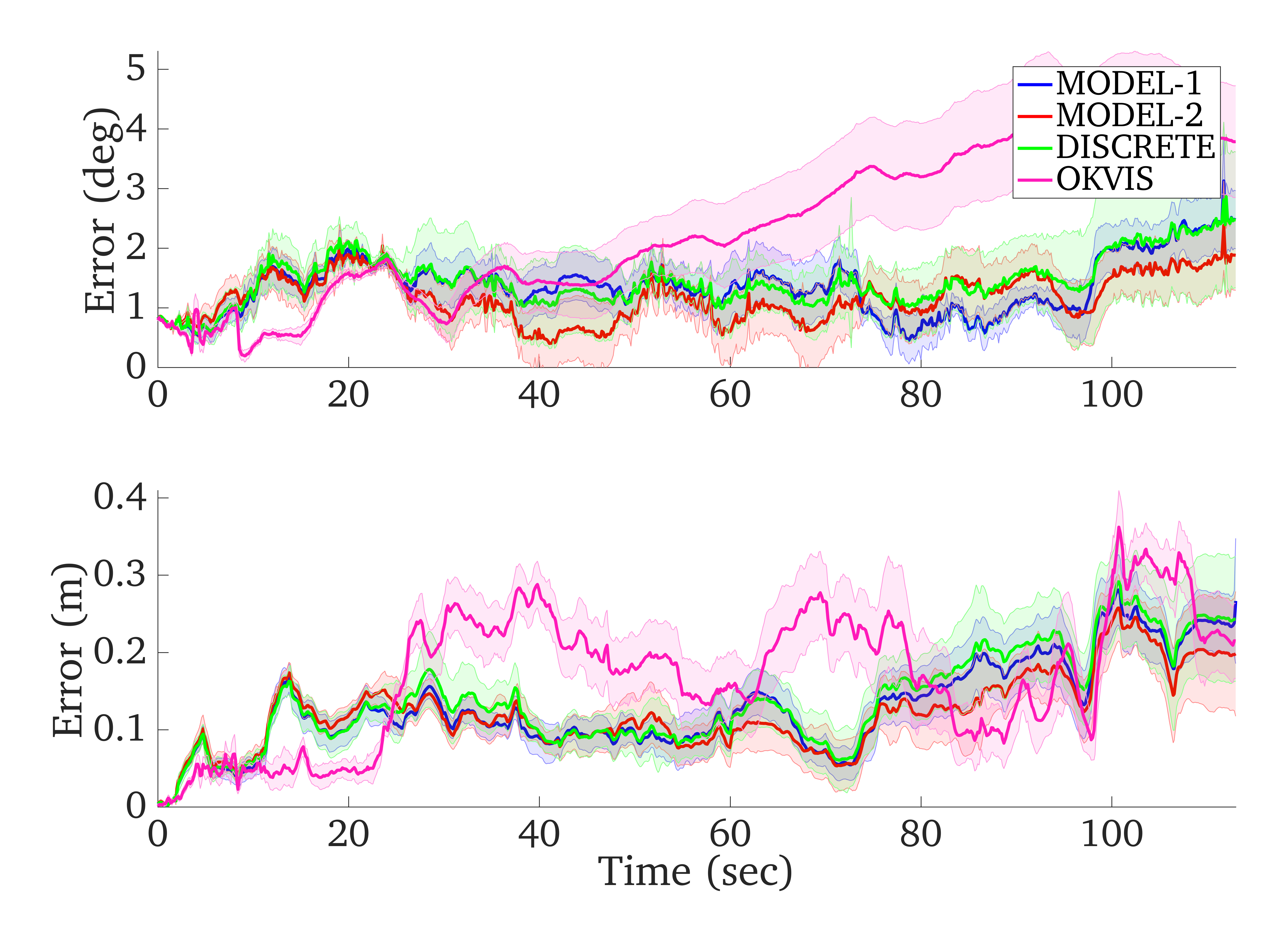}
        \caption{Average position and orientation RMSE for ``V2\_02\_med''.}
        \label{fig:rmseeth2}
    \end{subfigure}    \caption{Average trajectory and RMSE error over ten runs for the ``V1\_02\_med'' (top) and ``V2\_02\_med'' (bottom) sequences of the proposed tightly-coupled indirect VIO system. The one-sigma bound on the mean error is also shown and can be interpreted as the repeatability of the system (due to some randomness occurred in visual tracking). Note that this is not the same as estimator uncertainty and instead shows the variance of the VIO systems. The total trajectory lengths are 80 and 88 meters, respectively.}
\end{figure*}

\begin{table*} \caption{Average absolute RMSE results of the tightly-coupled indirect VINS for the EuRoC MAV sequences averaged over 10 runs. All systems were initialized with the ground truth state. The smallest position and orientation errors have been highlighted.}
\begin{center}
\scalebox{1.0}{
\begin{tabular}{|c|c|c|c|c|c|c|c|c|} \hline
& \multicolumn{2}{|c|}{\textbf{MODEL-1}} & \multicolumn{2}{|c|}{\textbf{MODEL-2}} & \multicolumn{2}{|c|}{\textbf{DISCRETE}} & \multicolumn{2}{|c|}{\textbf{OKVIS}} \\\hline
Units & m & deg & m & deg & m & deg & m & deg \\\hline
V1\_01\_easy & 0.2522 & 2.749 & \wi{0.2160} & 2.503 & 0.2547 & 2.781 & 0.2356 & \wi{2.458} \\\hline
V1\_02\_med  & 0.1342 & \wi{0.942} & \wi{0.1214} & 1.215 & 0.1344 & 1.001 & 0.1996 & 2.321 \\\hline
V1\_03\_diff & 0.1101 & 0.880 & \wi{0.0953} & \wi{0.809} & 0.1012 & 0.830 & 0.1830 & 3.498 \\\hline
V2\_01\_easy & 0.1429 & 1.069 & \wi{0.1426} & 1.148 & \wi{0.1426} & 1.118 & 0.1806 & \wi{0.973}  \\\hline
V2\_02\_med  & 0.1297 & 1.390 & \wi{0.1223} & \wi{1.135} & 0.1375 & 1.450 & 0.1695 & 2.334 \\\hline
V2\_03\_diff & 0.2982 & 2.159 & \wi{0.2800} & \wi{1.769} & 0.3055 & 2.052 & 0.3483 & 8.327 \\\hline\hline
MH\_01\_easy & 0.1817 & 1.398 & \wi{0.1653} & 1.761 & 0.2050 & 1.321 & 0.2523 & \wi{0.728} \\\hline
MH\_02\_easy & 0.1533 & 0.691 & \wi{0.1498} & \wi{0.525} & 0.1564 & 0.599 & 0.2523 & 0.728\\\hline
MH\_03\_med  & 0.2993 & 1.024 & \wi{0.2627} & 0.968 & 0.2800 & \wi{0.840} & 0.3193 & 1.903  \\\hline
MH\_04\_diff & 0.3312 & \wi{0.849} & 0.3515 & 0.974 & 0.3488 & 0.852 & \wi{0.2145} & 1.022 \\\hline
MH\_05\_diff & 0.3939 & \wi{0.692} & 0.3971 & 0.715 & \wi{0.3835} & 0.809 & 0.5432 & 0.738 \\\hline
\end{tabular}
}
\end{center}
\label{table:rmseethruns}
\end{table*}

We compared our tightly-coupled indirect VIO system with a state-of-the-art open-source VINS -- that is, the Open Keyframe-based Visual-Inertial SLAM (OKVIS)~\citep*{LeuteneggerIJRR2014}, although several different VINS methods were recently introduced (e.g., \cite*{Bloesch2017IJRR}).
We performed this comparison on the EuRoC MAV dataset~\citep*{Burri2016IJRR}, which has become the standard method for evaluating VINS algorithms and provides 20hz stereo pairs with a 200hz MEMS ADIS16448 IMU.
Our tightly-coupled preintegration-based system was run with inertial and pose sliding windows of 6 and 8 with a maximum of 300 extracted features.
Stereo-OKVIS was run with 4 and 6 inertial and keyframes with 300 features.
These parameters were selected to ensure real-time performance with both systems having minimal dropped frames.
It should be noted that depending on the tuning parameters used in the VINS algorithms, their performance may vary (e.g., see~\cite*{Delmerico2018ICRA}).
Note also that our VIO system uses a sliding window of poses as well as the inertial window connected with preintegrated measurements but does {\em not} keep any kind of map (to allow intra-window loop closures), 
while OKVIS employs a set of keyframes where mapped points are maintained.
Nevertheless, to provide a direct comparison, we initialize both systems with the true orientation, biases, velocity, and position such that no post-processing yaw alignment is needed.
Note also that due to some randomness that may occur during the visual tracking frontend (e.g., RANSAC-based outlier rejection), variations in the VINS results can be observed even if running the same algorithm on the same sequences.
To limit this variability of the algorithm, we perform 10 runs on the real-world sequences and average the results.

The ``V1\_02\_med'' and ``V2\_02\_med'' average trajectories can be seen in Figures \ref{fig:patheth1} and \ref{fig:patheth2} where we plot the estimated trajectories of our VIO and OKVIS along with the ground truth.
Note that it is understood that the performance of VINS algorithms may vary even if re-running on the same sequences due to some randomness in visual tracking (e.g., RANSAC); and thus, we repeated the test for 10 times and averaged the results in order to better evaluate the relative performance of the compared approaches.
Figures \ref{fig:rmseeth1} and \ref{fig:rmseeth2} show the averaged RMSE results of our VIO algorithm based on the proposed closed-form preintegration with the two models as compared to OKVIS,
which were computed at every time step and then averaged over all the runs;
while the averaged RMSE results are shown in Table \ref{table:rmseethruns}.
To show the repeatability/variability between runs, we also plot the standard deviation of the runs, noting that this should not be confused with the estimator uncertainty bounds commonly found in the literature.

\begin{figure} \centering
\hspace*{-1.2cm}
\includegraphics[width=0.85\columnwidth]{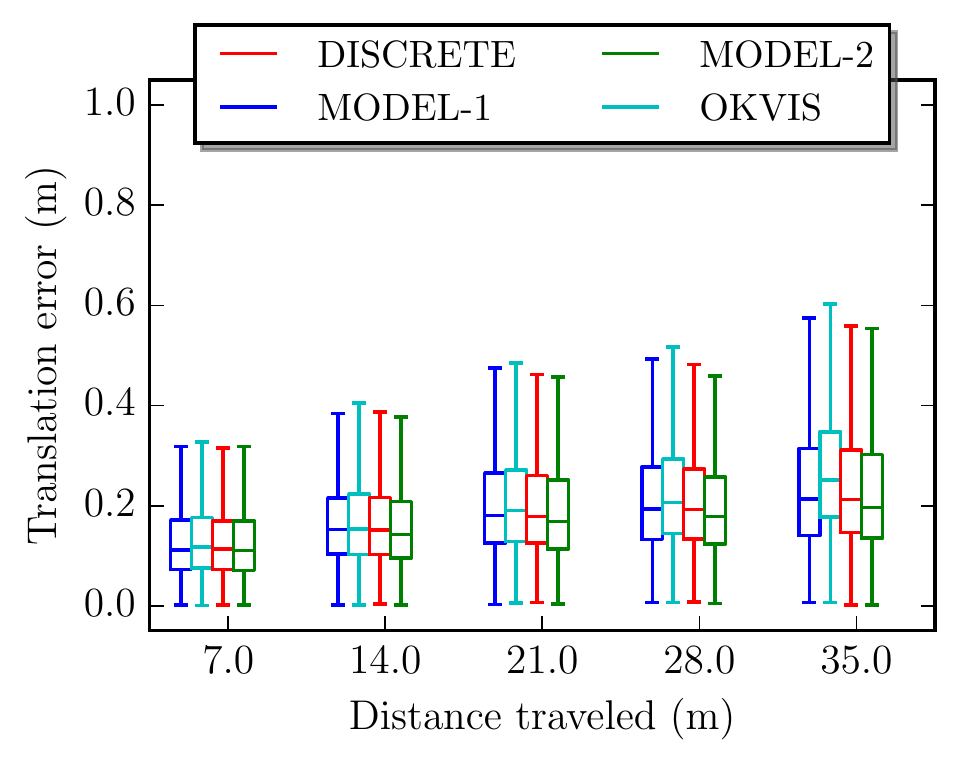}
\caption{
Boxplot of the odometric translation error statistics for the tightly-coupled indirect system evaluated over all of the EuRoC MAV sequences.
Errors were computed using the odometry metric over trajectory segments of \{7,14,21,28,35\} meters in length.
The middle box spans the first and third quartiles, while the whiskers are the upper and lower limits.
}
\label{fig:eth_rel_error}
\end{figure}

We additionally evaluated the trajectories of the proposed models using the odometry error metric \citep*{Zhang18IROS}.
As compared to the absolute RMSE value, this metric splits the trajectory into small segments of predetermined lengths, aligns the start of each segment to the ground truth, and then computes the error of the ending pose of the segment in respect to the ground truth.
This allows for insights of how drift is a function of distance.
Following the method proposed by \citet*{Zhang18IROS}, each of the ten runs performed by each model on the EuRoC MAV sequences were evaluated and the total odometric error over all sequences was computed.
Figure \ref{fig:eth_rel_error} and Table \ref{table:eth_rel_error}, show the resulting odometric error for trajectory segments of \{7,14,21,28,35\} meters.

These results clearly demonstrate that our VIO system can offer competitive performance to OKVIS; that is, we see instances where our method outperforms OKVIS, while in others OKVIS is superior.
Between the two proposed preintegration models, for these experiments, Model 2 offers the best performance.
We note that the discrete preintegration method tends to perform with lower accuracy compared to the two proposed models (although not in all cases), thereby validating the proposed preintegration models.

\begin{table} \caption{
Mean odometric translation errors for the tightly-coupled indirect system evaluated over all of the EuRoC MAV sequences.
Errors were evaluated over trajectory segments of \{7,14,21,28,35\} meters in length.
All errors are in meters.
}
\begin{center}
\scalebox{1.0}{
\begin{tabular}{|c|c|c|c|c|} \hline
& \textbf{MODEL-1} & \textbf{MODEL-2} & \textbf{DISCRETE} & \textbf{OKVIS} \\\hline
7 m & 0.137 & \wi{0.135} & 0.136 &  0.142  \\\hline
14 m & 0.177 & \wi{0.169} & 0.177 &  0.185  \\\hline
21 m & 0.220 & \wi{0.209} & 0.217 &  0.226  \\\hline
28 m & 0.229 & \wi{0.216} & 0.226 &  0.245  \\\hline
35 m & 0.247 & \wi{0.238} & 0.246 &  0.287  \\\hline
\end{tabular}
}
\end{center}
\label{table:eth_rel_error}
\end{table}

\subsubsection{UD indoor datasets:}

\begin{figure*} \centering
\begin{subfigure}[b]{0.48\textwidth}
\centering
    \includegraphics[width=0.48\textwidth]{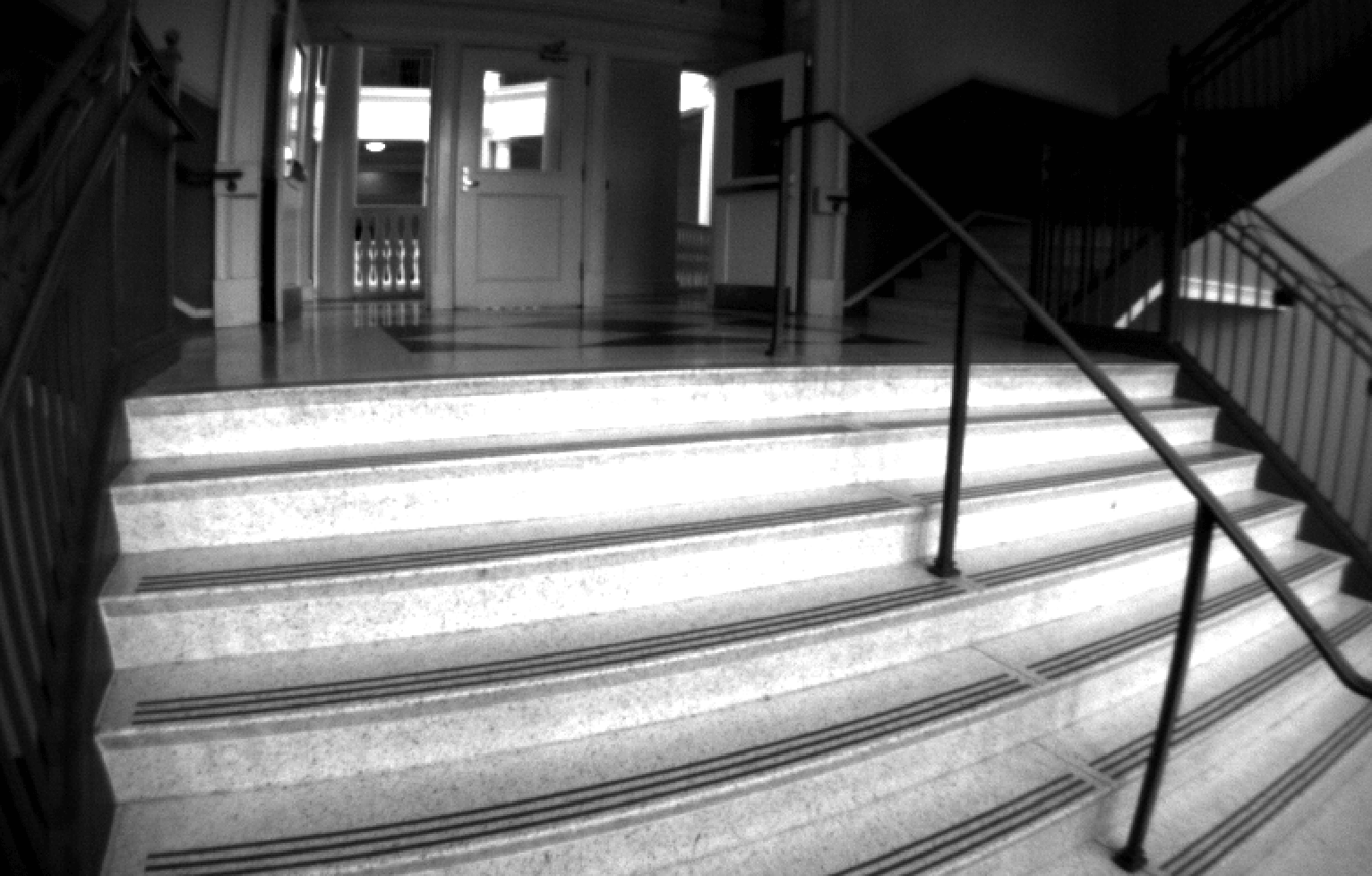}
    \includegraphics[width=0.48\textwidth]{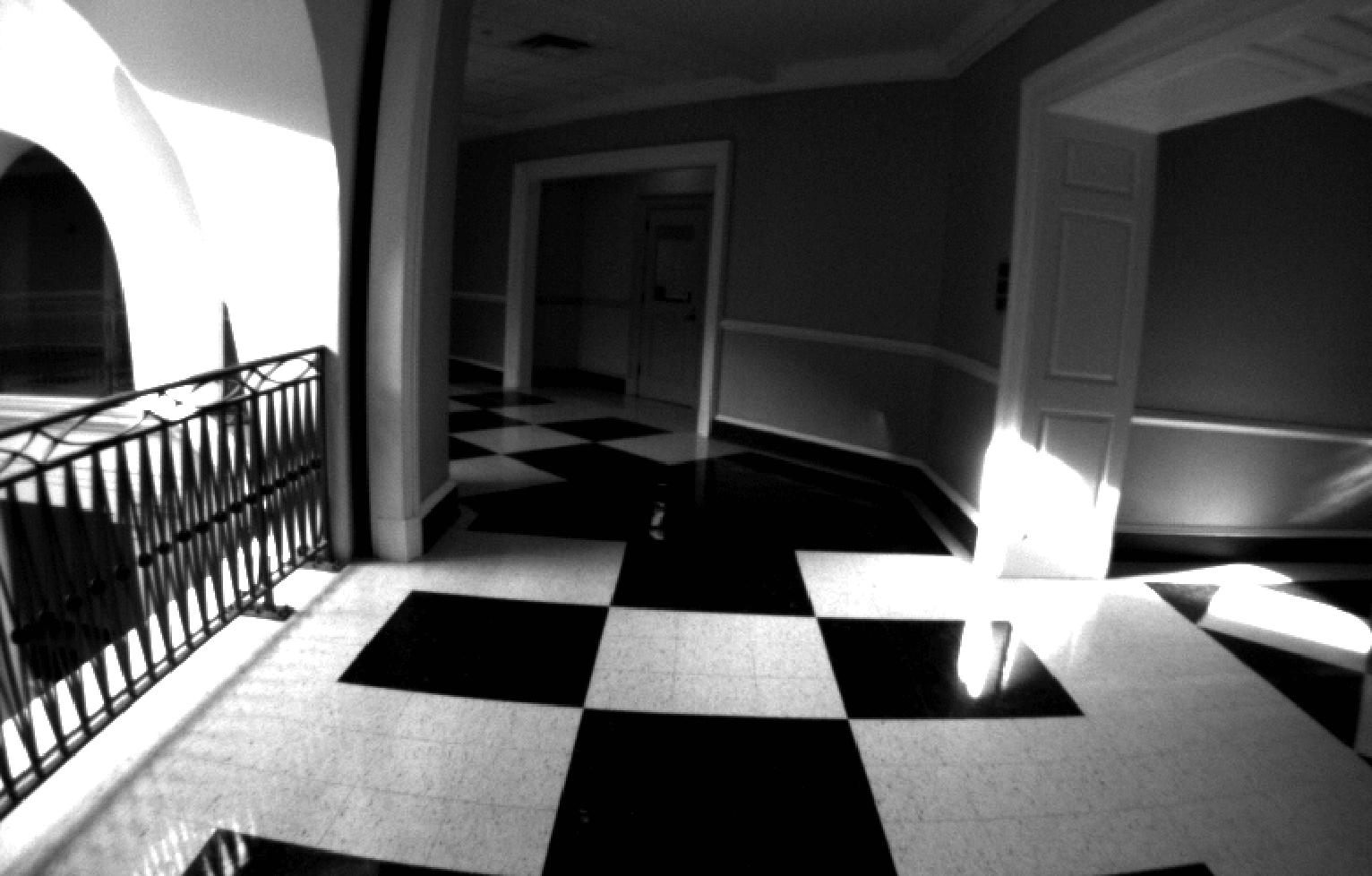}
    \includegraphics[height=4.5cm]{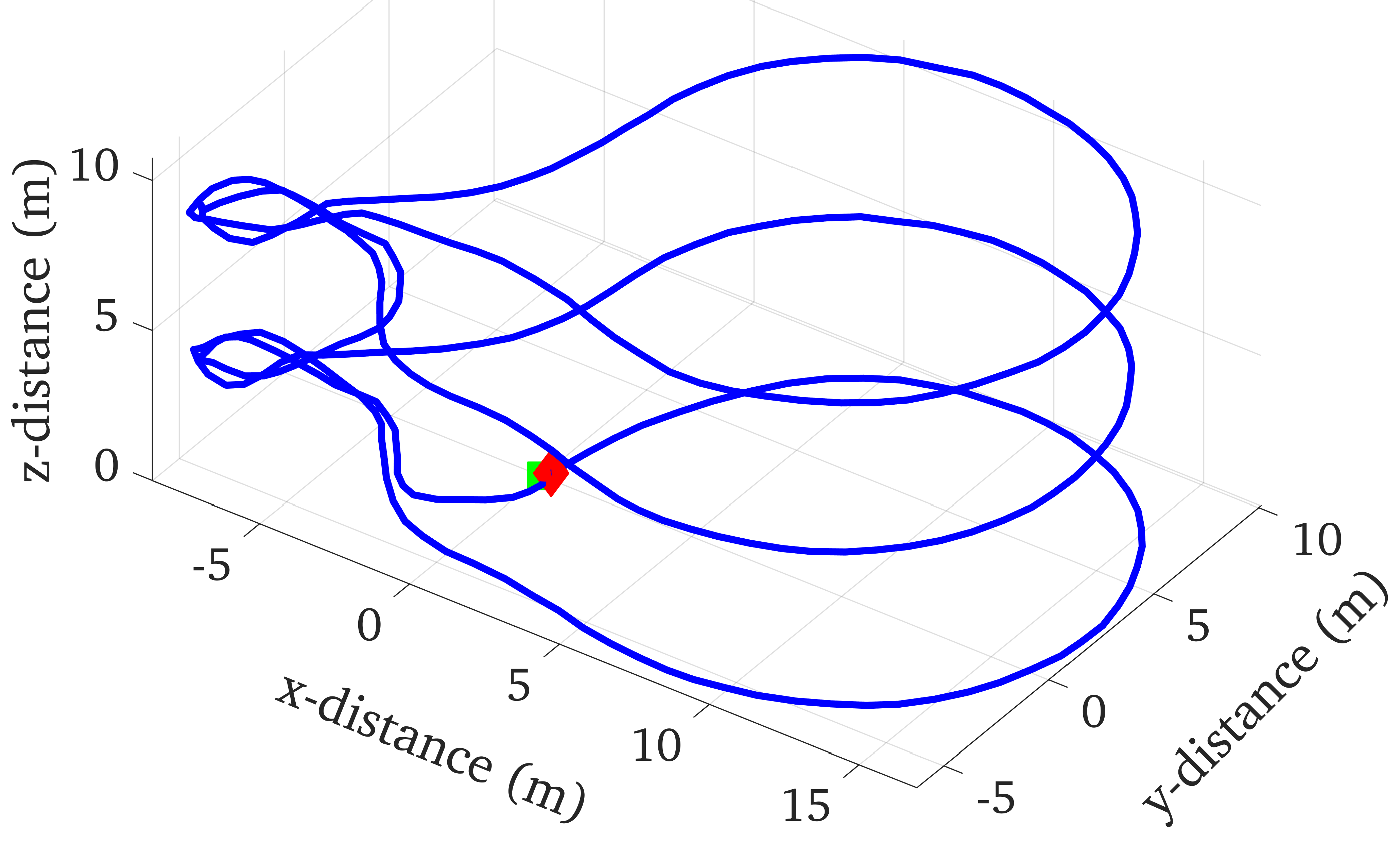}
    \caption{Example images (top) and 3D trajectory (bottom).}
    \label{fig:gore_iso}
\end{subfigure}    
\begin{subfigure}[b]{0.48\textwidth}
    \centering
    \includegraphics[height=7cm]{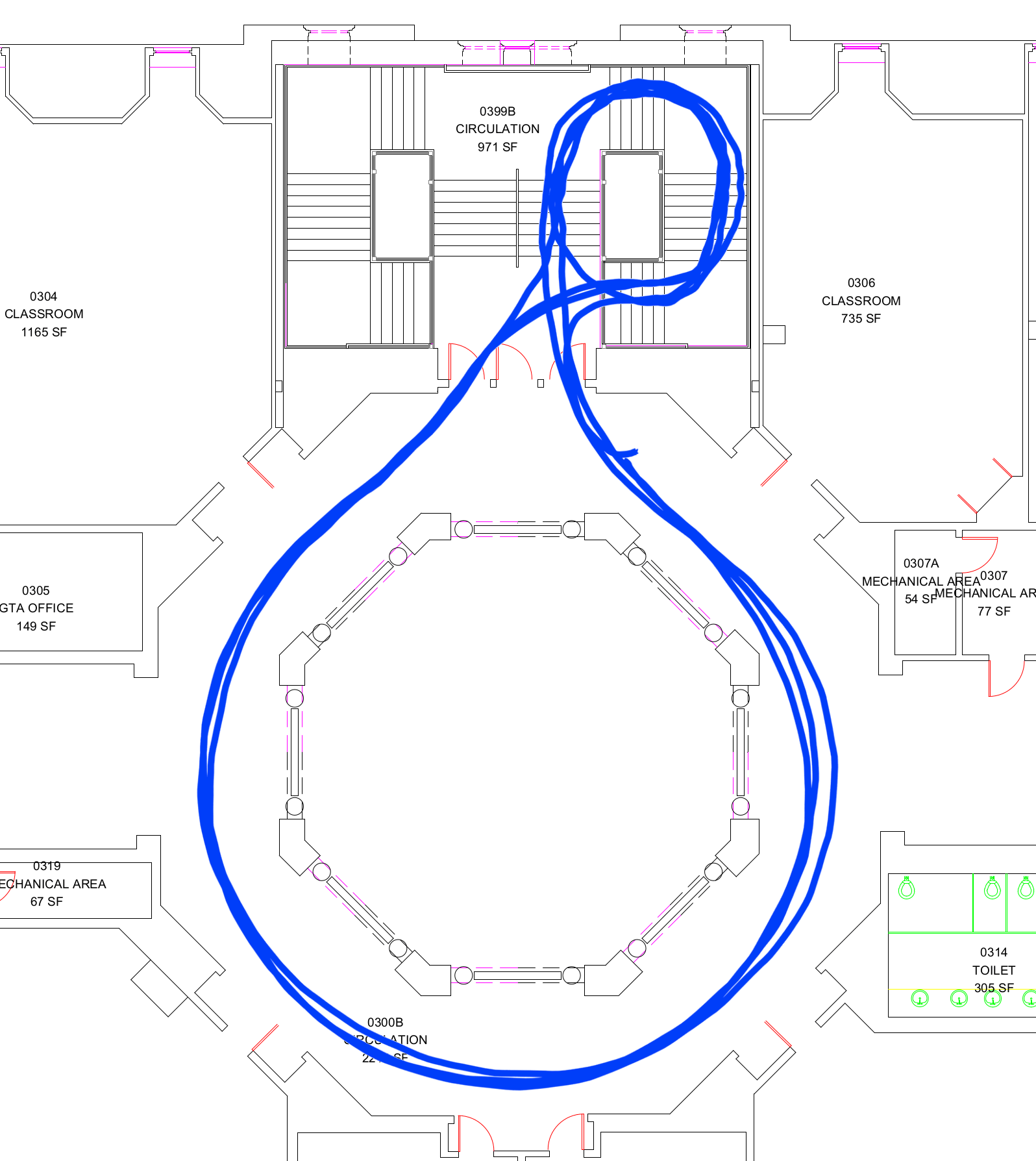}
    \caption{Projection of the estimated trajectory onto the floor plan. }
    \label{fig:gore_top}
\end{subfigure}
\caption{The trajectory estimates of the indoor experiment performed in the UD Gore Hall. Two example images from the dataset can be seen in (a), while the starting and ending locations are shown by a green square and red diamond in the plot, respectively. Note that the three floors have similar layouts, and thus only one floor plan is shown in plot (b).}
\end{figure*}

We further performed relatively large-scale (as compared to the EuRoC MAV dataset) indoor experiments in two buildings at the University of Delaware (UD) using our hand-held VI-Sensor with an IMU frequency of 400 Hz. 
In these experiments, because no ground truth was available, we initialized the system by keeping the device stationary for 
a short period of time (e.g., 2 seconds) 
so that the initial orientation and biases could be found, while the position and velocities were initialized as zero. To account for poor calibration of the sensor suite, both the IMU-to-camera spatial calibration parameters as well as the camera \textit{intrinsics} were estimated online. This was done by adding these quantities into the state and using the $\textit{raw}$ image coordinates as measurements, while expressing these as a function of the normalized pixel coordinates \eqref{eq::normalized} as well as the camera intrinsics \citep*{Li2014}.
The first indoor experiment was performed in the UD Gore Hall, in which the trajectory starts on the first floor, traverses the staircase to the third floor, and returns to the bottom floor, making a loop on each level.
To evaluate the estimation performance the trajectory returns to the original starting location.
The 3D trajectory estimate is shown in Figure \ref{fig:gore_iso} while its projection onto the building floor plan is shown in Figure \ref{fig:gore_top}.
We ran each preintegration model ten times across the 228 meter long dataset and averaged the results.
Model 1 had an ending error of 0.763 m (0.33$\%$ of the path), Model 2 had an ending error 0.747 m (0.33$\%$), discrete preintegration achieved 0.765 (0.34$\%$), and OKVIS achieved an ending error of 0.762 m (0.33$\%$) showing the improvement due to closed-form preintegration.

\begin{figure} \centering
\begin{subfigure}[b]{.5\textwidth}
\centering
    \includegraphics[width=0.48\textwidth]{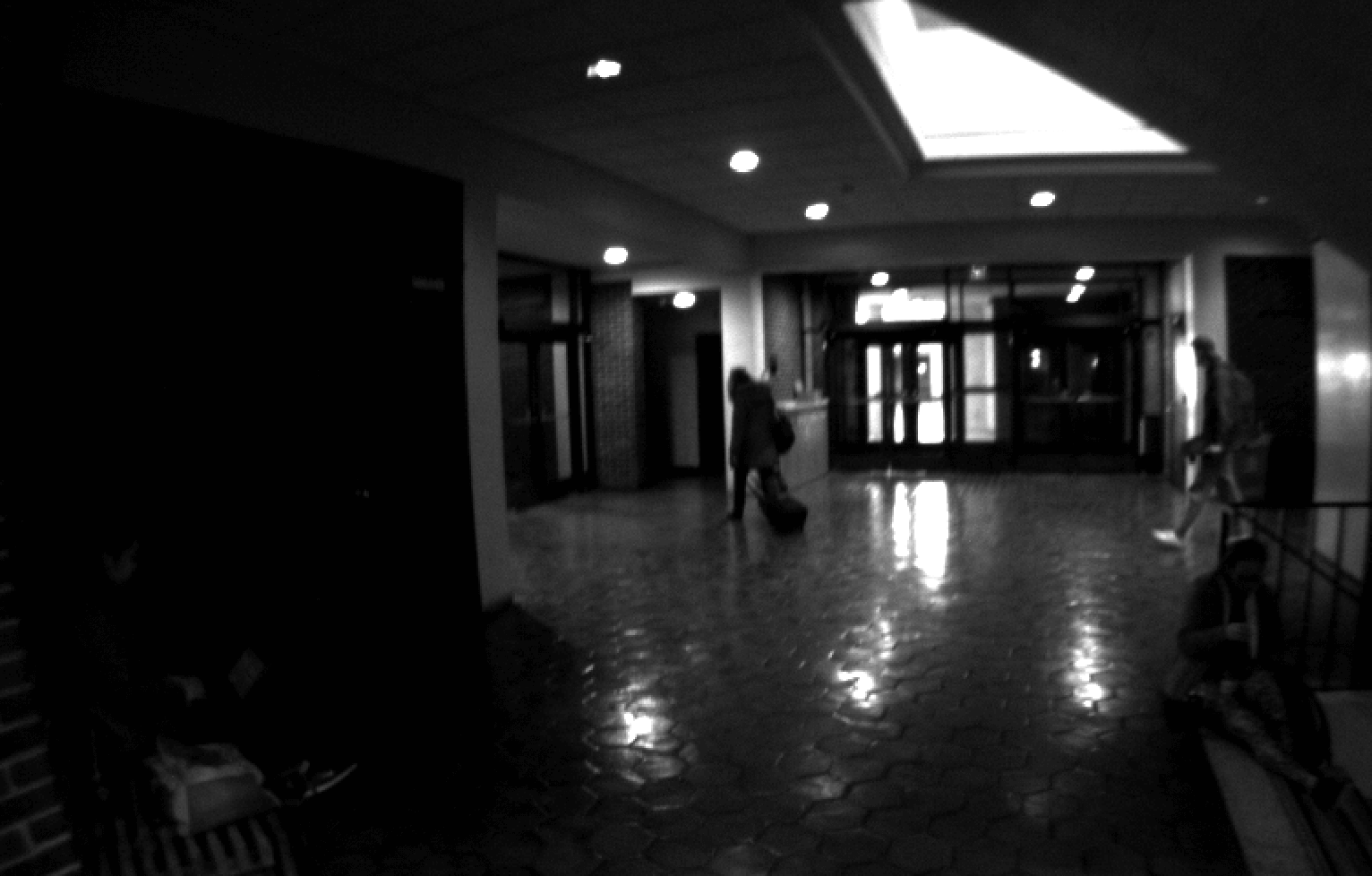}
    \includegraphics[width=0.48\textwidth]{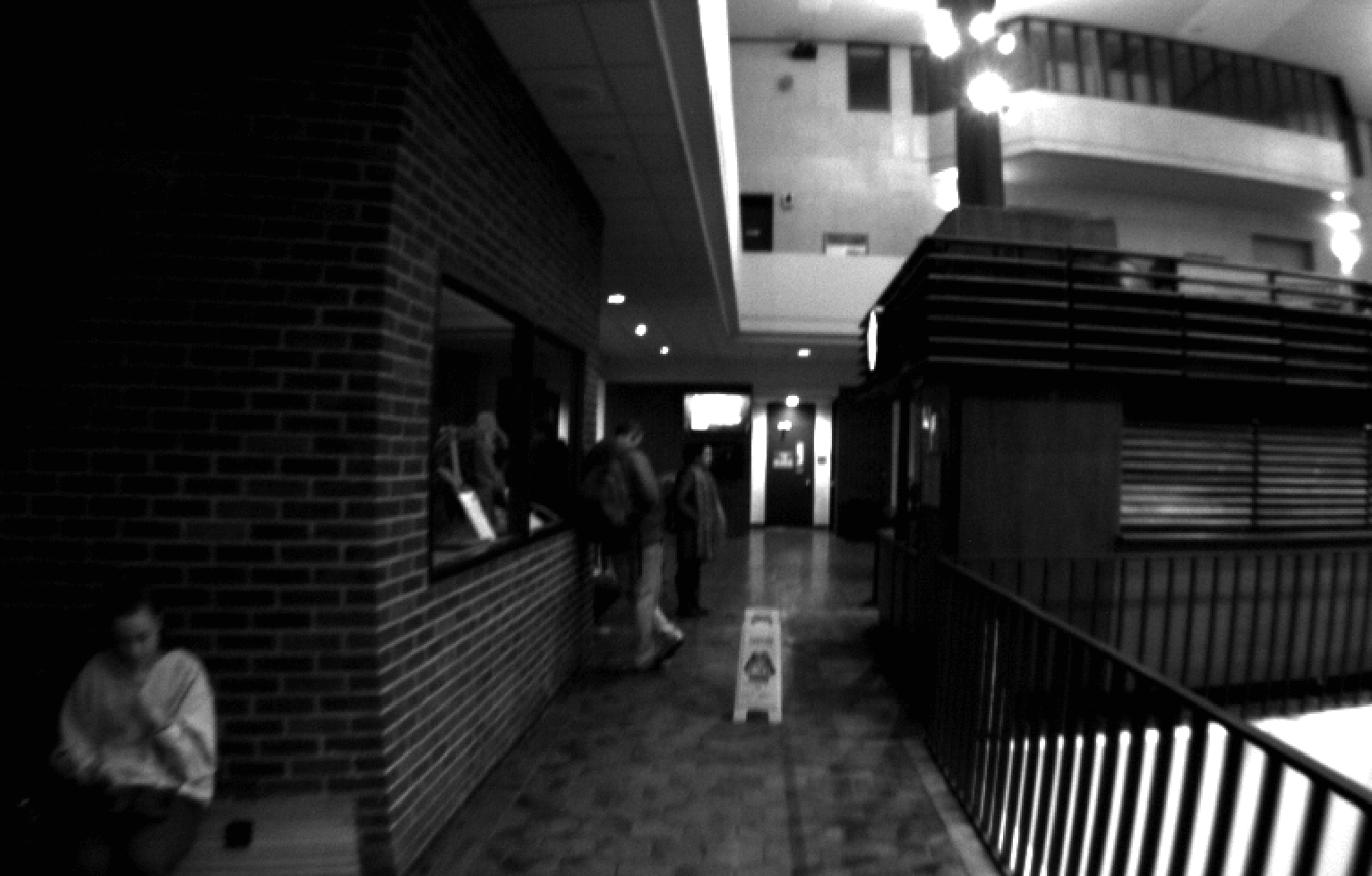}
    \caption{Two sample images seen during the experiment.}
    \label{fig:smith-image}
\end{subfigure} \\
\begin{subfigure}[b]{.5\textwidth}
\centering
    \includegraphics[width=0.9\textwidth]{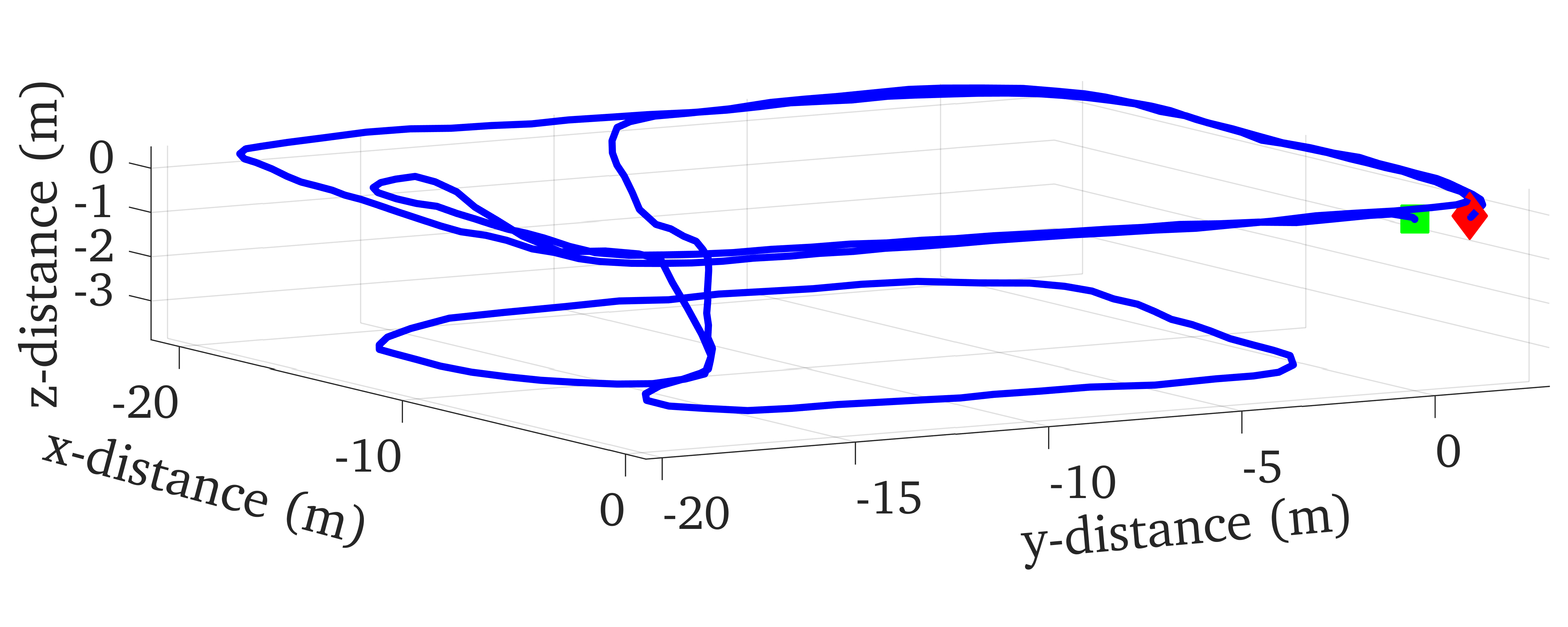}\vspace{-1em}
    \caption{View of the estimated 3D trajectory.}
    \label{fig:smith_iso}
\end{subfigure} \\
\vspace{1em}
\begin{subfigure}[b]{.5\textwidth}
    \centering
    \includegraphics[width=.8\textwidth]{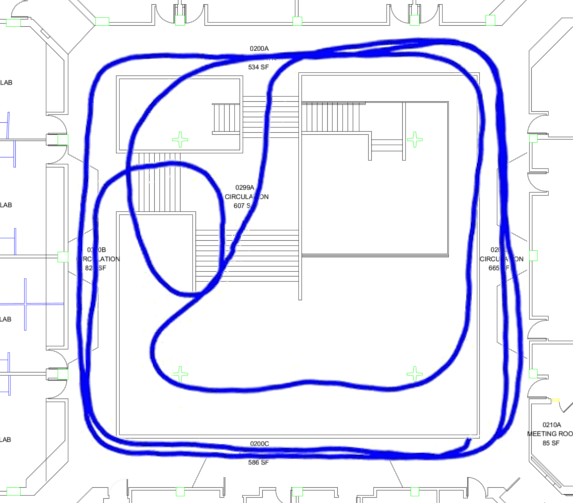}
    \caption{Projection of the estimated trajectory onto the floor plan.} \label{fig:smith_top}
\end{subfigure}
\caption{The results of the indoor experiment performed in the UD Smith Hall. Two example left camera images are shown in (a). In plot (b), the starting and ending locations of  the trajectory estimates are shown as a green square and red diamond, respectively.  Note that the trajectory shown in (c) occurs on both the second and first floors.}
\end{figure}

The second indoor experiment was conducted in the UD Smith Hall.
Starting on the second floor, we traversed along a rectangular wall before descending the stairs, looping around the first floor, then returning up the stairs, looping one and a half times around the upper level before returning to the starting position.
Model 1 had an ending error of 0.632 m (0.28$\%$ of the path), Model 2 had an ending error 0.788 m (0.35$\%$), discrete preintegration achieved 0.768 m (0.34$\%$), and OKVIS achieved 1.699 m (0.75$\%$) over the 230 meter trajectory.
Note that this scenario was more challenging than the first experiment, primarily due to the fact that
during this test, there were people walking around, lighting conditions were varying, and some parts of the environment lacked good features to detect and track (see Figure~\ref{fig:smith-image}).

The 3D trajectory estimate and its projection onto the floor plan are show in Figures~\ref{fig:smith_iso} and~\ref{fig:smith_top}, respectively. 
These results clearly demonstrate that our VIO systems using the proposed closed-form preintegration are able to perform accurate 3D motion tracking 
in relatively large-scale complex environments.

 \subsection{Loosely-Coupled Direct VINS} \label{sec:exp2}

When a stereo pair arrives, as in the preceding indirect VIO, we perform the proposed closed-form preintegration from the previous IMU state to the current state.
We then check the list of stored keyframes for a suitable candidate for direct image alignment,
based on a field-of-view constraint between the candidate and the current image.
If no such acceptable candidate is found, a new keyframe is created from the previous image pair 
and its depth map is computed using the OpenCV function \texttt{StereoSGBM}.
In particular, in order to perform course-to-fine alignment, the depth map is computed for multiple image pyramid levels.
Starting at the coarsest image level, we perform iterative image alignment, using the larger levels to further refine the course image alignment transform.
This image alignment optimization was implemented in a CUDA kernel for GPU acceleration, thus allowing for the system to achieve real-time performance.
After convergence, we recover the relative-pose constraint and add it as a factor to our direct-VINS graph.
Note that as compared to our indirect VIO method, we do not perform marginalization and thus allow later incorporation of loop closures.
To handle this increase of computational complexity and allow for real-time performance, 
we leverage the iSAM2 incremental smoothing implementation within the GTSAM framework \citep*{Kaess2012IJRR,Dellaert2012tr}.
However, the proposed framework is by no means optimal and can be further refined, for example, by more intelligently selecting keyframes.

 \subsubsection{EurocMav dataset:}

\begin{table*}[t]
\caption{Average absolute RMSE results for the EurocMav sequences over ten runs using the proposed direct VINS algorithm. 
All systems were initialized with the ground truth state. The smallest position and orientation errors have been highlighted.}
\begin{center}
\scalebox{1.}{
\begin{tabular}{|c|c|c|c|c|c|c|} \hline
& \multicolumn{2}{|c|}{\textbf{MODEL-1}}
& \multicolumn{2}{|c|}{\textbf{MODEL-2}}
& \multicolumn{2}{|c|}{\textbf{DISCRETE}} \\\hline
Units & m & deg & m & deg & m & deg \\\hline
V1\_01\_easy & \wj{0.2445} & \wj{2.218} & 0.2482 & 2.246 & 0.2530 & 2.223 \\\hline
V1\_02\_med  & 0.1598 & 1.767 & \wj{0.1309} & \wj{1.483} & 0.1763 & 1.899 \\\hline
V1\_03\_diff & \wj{0.0990} & \wj{1.180} & 0.1030 & 1.279 & 0.1030 & 1.234 \\\hline
V2\_01\_easy & \wj{0.1627} & 2.089 & 0.1940 & 1.956 & 0.1664 & \wj{1.533} \\\hline
V2\_02\_med  & 0.1809 & 2.530 & \wj{0.1665} & 2.527 & 0.1688 & \wj{2.309} \\\hline
V2\_03\_diff & 0.9337 & 6.187 & \wj{0.8927} & 5.425 & 1.0137 & \wj{4.998} \\\hline\hline
MH\_01\_easy & \wj{0.2947} & 2.270 & 0.3277 & \wj{2.148} & 0.3217 & 2.226 \\\hline
MH\_02\_easy & \wj{0.1882} & 1.650 & 0.2136 & 1.582 & 0.2008 & \wj{1.483} \\\hline
MH\_03\_med  & 0.2330 & 2.096 & 0.2295 & 2.121 & \wj{0.2288} & \wj{2.092} \\\hline
MH\_04\_diff & 0.4792 & \wj{2.513} & 0.4867 & 2.627 & \wj{0.4724} & 2.562 \\\hline
MH\_05\_diff & \wj{0.2884} & \wj{1.664} & 0.3014 & 1.722 & 0.2946 & 1.700 \\\hline
\end{tabular}
}
\end{center}
\label{table:rmseethdirect}
\end{table*}

To validate our direct VINS approach, we perform tests on the same EurocMav sequences as before,
which allow for direct comparison to a ground-truth trajectory \citep*{Burri2016IJRR}.
The results of the proposed direct VINS using two different preintegration models are shown in Table~\ref{table:rmseethdirect}.
Clearly, 
in scenarios in which a large amount of loop closures are present (e.g., ``V1\_03\_diff''), this system can outperform the tightly-coupled VIO system (see Section~\ref{sec:vio} and Table~\ref{table:rmseethruns}). 
However, when such loop closures are not available, the loosely-coupled systems suffer from larger drifts, as can be seen from the result of ``V2\_01\_easy''.

In these experiments, both proposed models tended to offer improved performance as compared to discrete preintegration (although not in all cases), while providing similar levels of performance to each other, with each having trajectories where they outperform the other.
In addition, the proposed direct VINS is sensitive to the tuning parameters, which in this experiment were chosen as {\em identical} across all sequences, rather than finding an optimal set per scenario. This led to situations such as ``V2\_03\_diff'', in which some of the runs yield incorrect loop closures despite our system attempting to reject these, greatly corrupting the resulting trajectory estimates. 
However, as the purpose of this work is to show the accuracy of the proposed preintegration, instead of the robustness of the utilized front-ends, these results along with the previous simulation results strongly suggest that our preintegration models can be, and should be, used when designing graph-based VINS.

 \section{Conclusions and Future Work} \label{sec:concl}

In this paper, we have analytically derived closed-form inertial preintegration 
and successfully applied it to graph-based visual-inertial navigation systems (VINS).
In particular, we advocate two new preintegration models for the evolution of IMU measurements across sampling intervals.
In the first, we assume that the inertial measurements remain piecewise constant;
while in the second, we incorporate a piecewise constant \textit{local} acceleration model into the preintegration framework.
We have validated through extensive Monte-Carlo simulations that both models outperform the state-of-the-art \textit{discrete} preintegration.
Furthermore, we have utilized this closed-form preintegration theory and developed two different VINS algorithms 
primarily to show the advantages of the proposed preintegration.
In the first, we formulated an indirect (feature-based), tightly-coupled, sliding-window optimization based VIO system 
that offers competitive (if not better) performance to a state-of-the-art graph-based VINS algorithm.
The second VINS method was developed instead based on loosely-coupled direct image alignment with the proposed preintegrations, 
allowing for efficient incorporation of informative loop closures.

In the future, we will integrate the proposed closed-form preintegrations to aided inertial navigation systems with other aiding sources (e.g., LiDAR).
We also seek to further robustify our VINS to handle more challenging scenarios (e.g., ultra-fast motion and highly-dynamic scenes).

\section*{Funding}
This work was partially supported by the University of Delaware College of Engineering,
the NSF [grant number IIS-1566129], the DTRA [grant number HDTRA1-16-1-0039], and Google Daydream.

{
\vspace{0.5cm}
\bibliography{libraries/related_works,libraries/library_extra,libraries/library,libraries/rpng}
}

{
\allowdisplaybreaks

\section*{Appendix B: Preintegration Measurement Jacobians}

\subsection*{B.1. Model 1 Measurement Jacobians}

We first partition the preintegrated measurement residual as follows:
\begin{align}
\mathbf{e}_{IMU}(\mathbf{x})= \begin{bmatrix} \mathbf{e}_{\theta}^\top & \mathbf{e}_{b_\omega}^\top & \mathbf{e}_{v}^\top & \mathbf{e}_{b_a}^\top & \mathbf{e}_p^\top \end{bmatrix}^\top
\end{align}
The measurement Jacobian with respect to each element of the error state vector can be found by perturbing the measurement function for the corresponding element.
For example, the relative-rotation measurement residual $\mathbf{e}_{\theta}$ is perturbed by a small change in gyro bias around the current estimate,
i.e., $\mathbf{b}_{\omega_k} - {\mathbf b}^{\star}_{\omega_k} = \hat{\mathbf{b}}_{\omega_k}+ \delta{\mathbf{b}}_{\omega_k} - {\mathbf b}_{\omega_k}^\star$, which yields (see Equation~\eqref{eq:model1-residual}):
\begin{align}
\mathbf{e}_{\theta} &= 
\scalemath{.9}{
2\mathbf{vec}\left( {}^{k+1}_G\hat{\bar q} \otimes {{}^{k}_G\hat{\bar q}}^{-1} \otimes {{}^{k+1}_{k} \breve{\bar{q}} }^{-1} \otimes \begin{bmatrix} \frac{\mathbf{J}_q (\hat{\mathbf{b}}_{\omega_k}+ \delta{ \mathbf{b}}_{\omega_k}-\mathbf{ b}^\star_{\omega_k})}{2} \\ 1
\end{bmatrix}\right) } \notag \\
 &=: 2\mathbf{vec}\left(   \hat{\bar q}_{r} \otimes \begin{bmatrix} \frac{\mathbf{J}_q (\hat{\mathbf{b}}_{\omega_k}+ \delta{\mathbf{b}}_{\omega_k}-\mathbf{ b}^{\star}_{\omega_k})}{2} \\ 1
 \end{bmatrix}\right) \notag\\
 &= 2\mathbf{vec}\left(\mathcal{L}(\hat{\bar q}_{r}) \begin{bmatrix} \frac{\mathbf{J}_q (\hat{\mathbf{b}}_{\omega_k}+ \delta{ \mathbf{b}}_{\omega_k}-\mathbf{b}^\star_{\omega_k})}{2} \\ 1
 \end{bmatrix}\right) \notag\\
&= \scalemath{.9}{
2\mathbf{vec}\left(
\begin{bmatrix}
\hat{q}_{r,4} \mathbf{I}_{3 \times 3} - \lfloor \hat{\mathbf{q}}_r \rfloor && \hat{\mathbf{q}}_r \\ -\hat{\mathbf{q}}_r^{\top} && \hat{q}_{r,4} \end{bmatrix}\begin{bmatrix} \frac{\mathbf{J}_q (\hat{\mathbf{b}}_{\omega_k}+ \delta{ \mathbf{b}}_{\omega_k}-\mathbf{ b}^\star_{\omega_k})}{2} \\ 1
\end{bmatrix}\right)  }\notag\\
&=  \scalemath{.9}{
(\hat{q}_{r,4}\mathbf{I}_{3 \times 3} - \lfloor \hat{\mathbf{q}}_r \rfloor)\mathbf{J}_q (\hat{\mathbf{b}}_{\omega_k}+ \delta{ \mathbf{b}}_{\omega_k}-\mathbf{b}^\star_{\omega_k}) + {\rm other~terms} }
\end{align}
As a result, the Jacobian with respect to a perturbance in bias can be read out as:
\begin{align}
\frac{\partial \mathbf{e}_{\theta} }{\partial  \delta{ \mathbf{b}}_{\omega_k}} &=  (\hat{q}_{r,4} \mathbf{I}_{3 \times 3} - \lfloor \hat{\mathbf{q}}_r \rfloor)\mathbf{J}_q 
\end{align}
Proceeding analogously, the Jacobian with respect to $^{k+1}\delta\bm \theta_G$ can be found as follows:
\begin{align}
\mathbf{e}_{\theta} 
&= 2\mathbf{vec}\left(  \begin{bmatrix} \frac{{}^{k+1}\delta\bm \theta_G}{2} \\ 1 \end{bmatrix} \otimes {}^{k+1}_G \hat{\bar q}\otimes {{}^{k}_G\hat{\bar q}}^{-1} \otimes {{}^{k+1}_k \bar{q} }^{-1} \otimes \hat{\bar q}_{b}\right)  \notag \\ 
 &=: 2\mathbf{vec}\left(  \begin{bmatrix} \frac{{}^{k+1}\delta\bm \theta_G}{2} \\ 1 \end{bmatrix} \otimes \hat{\bar{q}}_{rb}\right)  \notag \\
 &= 2\mathbf{vec}\left(\mathcal{R}(\hat{q}_{rb}) \begin{bmatrix} \frac{{}^{k+1}\delta\bm \theta_G}{2} \\ 1 \end{bmatrix}\right) \notag\\
&= 2\mathbf{vec}\left(
\begin{bmatrix}
\hat{q}_{rb,4} \mathbf{I}_{3 \times 3} + \lfloor \hat{\mathbf{q}}_{rb} \rfloor && \hat{\mathbf{q}}_{rb} \\ -\hat{\mathbf{q}}_{rb}^{\top} && \hat{q}_{rb,4} \end{bmatrix}\begin{bmatrix} \frac{1}{2}{{}^{k+1}\delta\bm \theta_G} \\ 1 \end{bmatrix}\right) \notag\\
&= (\hat{q}_{rb,4}\mathbf{I}_{3 \times 3} + \lfloor \hat{\mathbf{q}}_{rb} \rfloor) {{}^{k+1}\delta\bm \theta_G} + {\rm ~other~terms} \notag \\
\Rightarrow&  \frac{\partial \mathbf{e}_{\theta}  }{\partial  ^{k+1}\delta \bm\theta_G} = 
\hat{q}_{rb,4}\mathbf{I}_{3 \times 3} + \lfloor \hat{\mathbf{q}}_{rb} \rfloor
\end{align}
Similarly, the Jacobian with respect to  $^{k}\delta\bm \theta_G$ is computed by:
\begin{align}
\mathbf{e}_{\theta} 
&= 2\mathbf{vec}\left(  {}^{k+1}_G\hat{\bar q} \otimes {{}^{k}_G\hat{\bar q}}^{-1} \otimes \begin{bmatrix} -\frac{{}^{k}\delta\bm \theta_G}{2} \\ 1 \end{bmatrix}\otimes {{}^{k+1}_k\bar{q} }^{-1} \otimes \hat{\bar q}_{b}\right)  \notag\\
&=: 2\mathbf{vec}\left( \hat{q}_{n} \otimes \begin{bmatrix} -\frac{{}^{k}\delta\bm \theta_G}{2} \\ 1 \end{bmatrix}\otimes { \hat{q}_{mb} }^{-1} \right) \notag\\
 &= 2\mathbf{vec}\left(\mathcal{L}(\hat{q}_{n})\mathcal{R}({\bar{q}_{mb} }^{-1})\begin{bmatrix} -\frac{{}^{k}\delta\bm \theta_G}{2} \\ 1 \end{bmatrix}\right)  \notag\\
& =  2\mathbf{vec}\Big( \begin{bmatrix}
\hat{q}_{n,4}\mathbf{I}_{3\times 3} - \lfloor \hat{\mathbf{q}}_n \rfloor & \hat{\mathbf{q}}_n \\ -\hat{\mathbf{q}}_n^{\top} & \hat{q}_{n,4}
\end{bmatrix} \times \notag\\ 
&~~~~\begin{bmatrix}\bar{q}_{mb,4} \mathbf{I}_{3\times 3} - \lfloor \bar{\mathbf{q}}_{mb} \rfloor & -{\mathbf{q}}_{mb} \\ {\mathbf{q}}_{mb}^{\top} & \bar{q}_{mb,4} \end{bmatrix} \begin{bmatrix} -\frac{{}^{k}\delta\bm \theta_G}{2} \\ 1 \end{bmatrix} \Big)   \notag \\
& = -((\hat{q}_{n,4}\mathbf{I}_{3\times 3} - \lfloor \hat{\mathbf{q}}_n \rfloor)({q}_{mb,4}\mathbf{I}_{3\times 3} - \lfloor {\mathbf{q}}_{mb} \rfloor) \notag\\ & ~~~~ + \hat{\mathbf{q}}_n {\mathbf{q}}_{mb}^{\top}){}^{k}\delta\bm \theta_G + {\rm~other~terms} \notag \\
\Rightarrow &~~ \frac{\partial \mathbf{e}_{\theta} }{\partial  ^{k} \delta \bm\theta_G} = \notag\\ 
& -((\hat{q}_{n,4}\mathbf{I}_{3\times 3} - \lfloor \hat{\mathbf{q}}_n \rfloor)(\bar{q}_{mb,4}\mathbf{I}_{3\times 3} - \lfloor {\mathbf{q}}_{mb} \rfloor) +\hat{\mathbf{q}}_n \bar{\mathbf{q}}_{mb}^{\top}) 
\end{align}
where we have defined several intermediate quaternions, $\hat{\bar{q}}_r, \hat{\bar{q}}_{rb}, \hat{\bar{q}}_n,$ and $\hat{\bar{q}}_{mb}$, for ease of notation.
We compute the Jacobians of the remaining preintegrated measurements as follows~\citep*{Eckenhoff2018TR}:
\begin{align}
\frac{\partial \mathbf{e}_{b_\omega}}{\partial \delta \mathbf{b}_{\omega_{k}}} &= -\mathbf{I} \\
\frac{\partial \mathbf{e}_{b_\omega}}{\partial \delta \mathbf{b}_{\omega_{k+1}}} &= \mathbf{I} \\
\frac{\partial \mathbf{e}_v }{\partial ^{k} \delta \bm\theta_G} &= \left\lfloor {}^{k}_{G}\hat{\mathbf{R}} ({^G\hat{\mathbf{v}}_{k+1}} - {^G\hat{\mathbf{v}}_{k}} + {^G{\mathbf{g}}}\Delta t) \right\rfloor \\
\frac{\partial \mathbf{e}_v }{\partial  \delta{\mathbf{b}}_{\omega_k}} &= -\mathbf{J}_{\beta} \\
\frac{\partial \mathbf{e}_v }{\partial ^{G}{\delta \mathbf{v}}_{k}} &= - {^{k}_{G}\hat{\mathbf{R}}} \\
\frac{\partial \mathbf{e}_v }{\partial ^{G}\delta{\mathbf{v}}_{k+1}} &= {^{k}_{G}\hat{\mathbf{R}}} \\
\frac{\partial \mathbf{e}_v }{\partial  \delta \mathbf{b}_a} &= -\mathbf{H}_{\beta} \\
\frac{\partial \mathbf{e}_{b_a}}{\partial \delta \mathbf{b}_{a_{k}}} &= -\mathbf{I} \\
\frac{\partial \mathbf{e}_{b_a}}{\partial \delta \mathbf{b}_{a_{k+1}}} &= \mathbf{I} \\
\frac{\partial \mathbf{e}_p}{\partial ^{k}\delta\bm \theta_G}
&= \scalemath{0.9}{\left\lfloor ^{k}_{G}\hat{\mathbf{R}}  \left( {^{G}\hat{\mathbf{p}}_{k+1}} - {^{G}\hat{\mathbf{p}}_{k}} - {^G\hat{\mathbf{v}}_k} \Delta t+\frac{1}{2}{^G\mathbf{g}}\Delta t^2 \right) \right\rfloor}  \notag\\
\frac{\partial \mathbf{e}_p}{\partial \delta \mathbf{b}_{\omega_k}} & =-\mathbf{J}_{\alpha}  \\
\frac{\partial \mathbf{e}_p}{\partial ^{G}\delta{\mathbf{v}}_{k}}
& = -^{k}_{G}\hat{\mathbf{R}} \Delta t \\
\frac{\partial \mathbf{e}_p}{\partial \delta{\mathbf{b}}_{a_k}} & =-\mathbf{H}_{\alpha} \\
\frac{\partial \mathbf{e}_p}{\partial ^{G}\delta{\mathbf{p}}_{k}}
& = - {^{k}_{G}\hat{\mathbf{R}}} \\
\frac{\partial \mathbf{e}_p}{\partial ^{G}\delta{\mathbf{p}}_{k+1}} 
&=  {^{k}_G\hat{\mathbf{R}}} \\
\end{align}

\subsection*{B.2. Model 2 Measurement Jacobians}

For Model 2, the orientation measurement Jacobians remain the same as in Model 1.
For the remaining measurement Jacobians, we compute them in the same way as in Model 1 and are given by (see Equation~\eqref{eq:model2-res}):
\begin{align}
\frac{\partial \mathbf{e}_{b_\omega}}{\partial \delta \mathbf{b}_{\omega_{k}}} &= -\mathbf{I} \\
\frac{\partial \mathbf{e}_{b_\omega}}{\partial \delta \mathbf{b}_{\omega_{k+1}}} &= \mathbf{I} \\
\frac{\partial \mathbf{e}_{v}}{\partial {}^k\delta \bm \theta_G}
&= \scalemath{.9}{ \lfloor {}^k_G\mathbf{R} \left({}^G\hat{\mathbf{v}}_{k+1}-{}^G\hat{\mathbf{v}}_{k} \right)\rfloor - \mathbf{O}_\beta \left(\tilde q_4\mathbf{I}+\lfloor \tilde{\mathbf{q}} \rfloor \right)  } \\ 
\frac{\partial \mathbf{e}_{v}}{\partial \delta \mathbf{b}_{\omega_k}}
&= -\mathbf{J}_\beta \\
\frac{\partial \mathbf{e}_{v}}{\partial {}^G \delta \mathbf{v}_{k}}
&= -{}^k_G\hat{\mathbf{R}} \\
\frac{\partial \mathbf{e}_{v}}{\partial {}^G \delta \mathbf{v}_{k+1}}
&= {}^k_G\hat{\mathbf{R}}  \\
\frac{\partial \mathbf{e}_{v}}{\partial \delta \mathbf{b}_{a_{k}}}
&= -\mathbf{H}_\beta \\
\frac{\partial \mathbf{e}_{b_a}}{\partial \delta \mathbf{b}_{a_{k+1}}} &= \mathbf{I} \\
\frac{\partial \mathbf{e}_{b_a}}{\partial \delta \mathbf{b}_{a_{k}}} &= -\mathbf{I}  \\
\frac{\partial \mathbf{e}_{p}}{\partial {}^k\delta \bm \theta_G}
&= \lfloor {}^k_G\mathbf{R} \left({}^G\hat{\mathbf{p}}_{k+1}-{}^G\hat{\mathbf{p}}_{k} - {}^G\hat{\mathbf{v}}_{k} \Delta T\right)\rfloor \notag\\ 
&~~~~  -\mathbf{O}_\alpha \left(\tilde q_4\mathbf{I}+\lfloor \tilde{\mathbf{q}} \rfloor \right) \\
\frac{\partial \mathbf{e}_{p}}{\partial \delta \mathbf{b}_{\omega_k}}
&= -\mathbf{J}_\alpha  \\
\frac{\partial \mathbf{e}_{p}}{\partial {}^G\delta \mathbf{v}_{k}}
&= -{}^k_G\hat{\mathbf{R}}\Delta T \\
\frac{\partial \mathbf{e}_{p}}{\partial \delta \mathbf{b}_{a_{k}}}
&= -\mathbf{H}_\alpha \\
\frac{\partial \mathbf{e}_{p}}{\partial {}^G \delta\mathbf{p}_{k}}
&= -{}^k_G\hat{\mathbf{R}} \\
\frac{\partial \mathbf{e}_{p}}{\partial {}^G \delta \mathbf{p}_{k+1}}
&= {}^k_G\hat{\mathbf{R}}
\end{align}
where $[ \tilde{\mathbf{q}}^{\top} ~ \tilde{q}_4 ]^{\top} = {}^k_G \hat{\bar{q}} \otimes {}^k_G \bar{q}^{\star-1}$

 \section*{Appendix C: Inverse-Depth Measurement Jacobians}

We denote $a$ and $i$  the anchoring time step and the associated anchoring camera frame, respectively. 
Consider the case where we receive an image of the same feature at step $k$ from camera~$j$.
This measurement can be divided into three categories:
(i)~when the measurement refers to both the anchoring time and camera that the inverse depth is being represented in;
(ii)~when the measurement refers to the same anchoring time, but a different camera;
(iii)~when the anchoring time and measurement time are distinct. 

In case (i), we have (see Equation~\eqref{eq:inverse-depth-meas-h}):
\begin{align}
\mathbf{h} &= \begin{bmatrix} \alpha \\ \beta \\ 1 \end{bmatrix}
\end{align}
Then the measurement Jacobians are computed by (see Equations~\eqref{eq:e_fjk}, \eqref{eq:z_fjk} and \eqref{eq:inverse-depth-meas-h}):
\begin{align}
    \frac{\partial \mathbf{e}_{fjk}}{\partial \alpha} &= \mathbf{H}_{proj}(0,0,2,1) \\
    \frac{\partial \mathbf{e}_{fjk}}{\partial \beta} &= \mathbf{H}_{proj}(0,1,2,1) \\
    \frac{\partial \mathbf{e}_{fjk}}{\partial \rho} &= \mathbf{0} \\
    \mathbf{H}_{proj} &= \begin{bmatrix} \frac{1}{h_3} & 0 & \frac{-h_1}{(h_3)^2} \\
    0 & \frac{1}{h_3} &\frac{-h_2}{(h_3)^2} \end{bmatrix}
\end{align}
where $\mathbf{H}_{proj}(i,j,k,l)$ refers to the block matrix of size $(k,l)$ with starting index $(i,j)$.

In case (ii)
where $k$ refers to the same imaging time but a different camera (such as a stereo partner), 
we have (see Equation~\eqref{eq:inverse-depth-meas-h}):
\begin{align}
    \mathbf{h} &= {}_{C_i}^{C_j}\mathbf{R}\begin{bmatrix} \alpha \\ \beta \\ 1 \end{bmatrix}+ \rho{}^{C_j}\mathbf{p}_{C_i}
\end{align}
Because in this case the transformation parameters are rigid and known, we need only the derivatives with respect to the unknown feature parameterization:
\begin{align}
    \frac{\partial \mathbf{e}_{fjk}}{\partial \alpha} &= \Big[\mathbf{H}_{proj}{}_{C_i}^{C_j}\mathbf{R}\Big](0,0,2,1) \\
    \frac{\partial \mathbf{e}_{fjk}}{\partial \beta} &= \Big[\mathbf{H}_{proj}{}_{C_i}^{C_j}\mathbf{R}\Big](0,1,2,1) \\
    \frac{\partial \mathbf{e}_{fjk}}{\partial \rho} &= {}^{C_j}\mathbf{p}_{C_i}
\end{align}

In case (iii) where instead the measurement refers to a different time, we can write out the rigid transformation between the anchor and new current camera frame as follows:
\begin{align}
    {}^{C_{k,j}}\mathbf{p}_f &= {}_{C_{a,i}}^{C_{k,j}}\mathbf{R}{}^{C_{a,i}}\mathbf{p}_f+ {}^{C_{k,j}}\mathbf{p}_{C_{a,i}} \notag \\
    &= \frac{1}{\rho}{}_{I}^{C_j}\mathbf{R}{}_{G}^{k}\mathbf{R}{}_{a}^{G}\mathbf{R}{}_{C_i}^{I}\mathbf{R}\begin{bmatrix} \alpha \\ \beta \\ 1 \end{bmatrix}  +  \notag\\
    &~~~~ {}^{C_{k,j}}_G\mathbf{R}\left({}^G\mathbf{p}_{C_{a,i}} - {}^G\mathbf{p}_{C_{k,j}} \right) \notag\\
                    &= \frac{1}{\rho}{}_{I}^{C_j}\mathbf{R}{}_{G}^{k}\mathbf{R}{}_{a}^{G}\mathbf{R}{}_{C_i}^{I}\mathbf{R}\begin{bmatrix} \alpha \\ \beta \\ 1 \end{bmatrix} + \notag\\
    &~~~~ {}^{C_j}_{I}\mathbf{R}{}^{k}_{G}\mathbf{R}\left({}^G\mathbf{p}_{a}+ {}^{G}_{a}\mathbf{R}{}^{I}\mathbf{p}_{C_i} - {}^G\mathbf{p}_{k} - {}^{G}_{k}\mathbf{R}{}^{I}\mathbf{p}_{C_j}\right)  \notag\\
                        &= {}_{I}^{C_j}\mathbf{R}{}_{G}^{k}\mathbf{R}{}_{a}^{G}\mathbf{R}{}_{C_i}^{I}\mathbf{R}\left(\frac{1}{\rho}\begin{bmatrix} \alpha \\ \beta \\ 1 \end{bmatrix}-  {}^{C_i}\mathbf{p}_{I} \right) + \notag\\
    &~~~~    {}_{I}^{C_j}\mathbf{R}{}_{G}^{k}\mathbf{R}({}^G\mathbf{p}_{a}-  {}^G\mathbf{p}_{k} ) + {}^{C_j}\mathbf{p}_{I} 
\end{align}
With this, we have:
\begin{align}
    \mathbf{h} =& {}_{I}^{C_j}\mathbf{R}{}_{G}^{k}\mathbf{R}{}_{a}^{G}\mathbf{R}{}_{C_i}^{I}\mathbf{R}\left(\begin{bmatrix} \alpha \\ \beta \\ 1 \end{bmatrix}-  \rho {}^{C_i}\mathbf{p}_{I} \right) +  \notag\\
    &  \rho{}_{I}^{C_j}\mathbf{R}{}_{G}^{k}\mathbf{R}\left({}^G\mathbf{p}_{a}- {}^G\mathbf{p}_{k} \right) + \rho{}^{C_j}\mathbf{p}_{I}
\end{align}
We can then take the derivative with respect to each variable:
\begin{align}
    \frac{\partial \mathbf{e}_{fjk}}{\partial  {}^a\delta \bm \theta_G} &= \scalemath{0.9}{-\mathbf{H}_{proj}{}_{I}^{C_j}\mathbf{R}{}_{G}^{k}\mathbf{R}{}_{a}^{G} \mathbf{R}\left\lfloor {}_{C_i}^{I}\mathbf{R}\left(\begin{bmatrix} \alpha \\ \beta \\ 1 \end{bmatrix}-  \rho {}^{C_i}\mathbf{p}_{I} \right) \right\rfloor } \\
                    \frac{\partial \mathbf{e}_{fjk}}{\partial {}^G \mathbf{p}_{a}} &= \mathbf{H}_{proj}\rho{}_{I}^{C_j}\mathbf{R}{}_{G}^{k}\mathbf{R} \\
                \frac{\partial \mathbf{e}_{fjk}}{\partial  {}^k\delta \bm \theta_G} &=  \mathbf{H}_{proj} {}_{I}^{C_j}\mathbf{R}\Bigg\lfloor {}_{G}^{k}\mathbf{R}{}_{a}^{G}\mathbf{R}{}_{C}^{I_i}\mathbf{R}\left(\begin{bmatrix} \alpha \\ \beta \\ 1 \end{bmatrix}-\rho {}^{C_i}\mathbf{p}_{I} \right) + \notag\\ 
    &~~~~\rho{}_{G}^{k}\mathbf{R}\left({}^G\mathbf{p}_{a}- {}^G\mathbf{p}_{k} \right) \Bigg\rfloor \\ 
                \frac{\partial \mathbf{e}_{fjk}}{\partial {}^G \mathbf{p}_{k}} &= -\mathbf{H}_{proj}\rho{}_{I}^{C_j}\mathbf{R}{}_{G}^{k}\mathbf{R} \\
                \frac{\partial \mathbf{e}_{fjk}}{\partial \alpha} &=  \Big[\mathbf{H}_{proj}{}_{I}^{C_j}\mathbf{R}{}_{G}^{k}\mathbf{R}{}_{a}^{G}\mathbf{R}{}_{C_i}^{I}\mathbf{R}\Big](0,0,2,1) \\ 
     \frac{\partial \mathbf{e}_{fjk}}{\partial \beta} &=  \Big[\mathbf{H}_{proj}{}_{I}^{C_j}\mathbf{R}{}_{G}^{k}\mathbf{R}{}_{a}^{G}\mathbf{R}{}_{C_i}^{I}\mathbf{R}\Big](0,1,2,1) \\ 
     \frac{\partial \mathbf{e}_{fjk}}{\partial \rho} &=  \mathbf{H}_{proj}(-{}_{I}^{C_j}\mathbf{R}{}_{G}^{k}\mathbf{R}{}_{a}^{G}\mathbf{R}{}_{C_i}^{I}\mathbf{R}{}^{C_i}\mathbf{p}_{I} + \notag \\ 
     &~~~~ {}_{I}^{C_j}\mathbf{R}{}_{G}^{k}\mathbf{R}\left({}^G\mathbf{p}_{a}- {}^G\mathbf{p}_{k} \right) + {}^{C_j}\mathbf{p}_{I} )
\end{align}

 \section*{Appendix D: Relative-Pose Measurement Jacobian}

Recall that in Equation~\eqref{eq:direct-res}, $j$ denotes the query image and $k$ is the keyframe. 
We partition the relative-pose residual $\mathbf{e}_d$ into the relative-orientation residual $\mathbf{e}_\theta$   and the relative-position residual $\mathbf{e}_p$.
The Jacobians with respect to the states can be found by perturbation in the same way as before.
\begin{align}
    \mathbf{e}_{\theta} &= 2\textbf{vec}\left( \begin{bmatrix} \frac{{}^j\bm \delta \theta_G}{2} \\ 1 \end{bmatrix}\otimes{}_G^j\hat{\bar}{q} \otimes {}_G^k\hat{\bar{q}}^{-1} \otimes {}_k^j\breve{\bar{q}}^{-1}  \right) \notag \\
        &= 2 \textbf{vec}\left(
    \scalemath{0.95}{
    \mathcal{R}\left({}_G^j\hat{\bar}{q} \otimes {}_G^k\hat{\bar{q}}^{-1} \otimes {}_k^j\breve{\bar{q}}^{-1} \right)  \begin{bmatrix} \frac{{}^j\delta\bm \theta_G}{2} \\ 1 \end{bmatrix}
    }
    \right) \notag\\ 
    &= \left(\bar{q}_{r,4}\mathbf{I} + \lfloor \mathbf{q}_{r} \rfloor \right) {}^j\delta \bm \theta_G + \cdots \notag\\
    \Rightarrow~~\frac{\partial \mathbf{e}_\theta}{\partial {}^j\delta\bm \theta_G} &= \left(\bar{q}_{r,4}\mathbf{I} + \lfloor \mathbf{q}_{r} \rfloor \right)
\end{align}
Similarly, we perturb the quaternion estimate of the keyframe to compute the corresponding Jacobian as:
\begin{align}
    \mathbf{e}_{\theta} &=  
     2\textbf{vec}\left({}_G^j\hat{\bar}{q} \otimes {}_G^k\hat{\bar{q}}^{-1} \otimes  \begin{bmatrix} \frac{-{}^k\delta\bm \theta_G}{2} \\ 1 \end{bmatrix} \otimes {}_k^j\breve{\bar{q}}^{-1}  \right)  \notag\\
    & =2\textbf{vec}\left(\mathcal{L} \left({}_j^k\hat{\bar{q}}\right) \mathcal{R} \left({}_j^k\breve{\bar{q}}\right)^{\top}\begin{bmatrix} \frac{-{}^k\delta\bm \theta_G}{2} \\ 1 \end{bmatrix} \right) = \notag\\
    & \scalemath{0.9}{-\left( \left({}_k^j\hat{\bar{q}}_4 \mathbf{I} - \lfloor{}_k^j\hat{\mathbf{q}} \rfloor\right)
    \Big({}_k^j\breve{\bar{q}}_4 \mathbf{I} - \lfloor{}_k^j\breve{\mathbf{q}} \rfloor\right) 
    + {}_k^j\hat{\mathbf{q}}{}_k^j\breve{\mathbf{q}}^{\top} \Big){}^k\delta\bm \theta_G + \cdots } \notag\\ 
       &\Rightarrow  \frac{\partial \mathbf{e}_\theta}{\partial {}^k\delta\bm \theta_G} \!\! = \!\!
   \scalemath{0.9}{-\left( \left({}_k^j\hat{\bar{q}}_4 \mathbf{I} - \lfloor{}_k^j\hat{\mathbf{q}} \rfloor\right)\left({}_k^j\breve{\bar{q}}_4 \mathbf{I} - \lfloor{}_k^j\breve{\mathbf{q}} \rfloor\right)+ {}_k^j\hat{\mathbf{q}}{}_k^j\breve{\mathbf{q}}^{\top} \right) } \notag
\end{align}
Again by following a similar procedure, we can find the Jacobians of the relative-position residual with respect to the state as follows:
\begin{align}
    \frac{\partial \mathbf{e}_p}{\partial {}^G \delta \mathbf{p}_j} &= {}_G^k\hat{\mathbf{R}}  \\
     \frac{\partial\mathbf{e}_p}{\partial {}^G \delta \mathbf{p}_k}&= -{}_G^k\hat{\mathbf{R}}  \\
       \frac{\partial\mathbf{e}_p}{\partial {}^k\delta\bm \theta_G}&= \lfloor {}_G^k\hat{\mathbf{R}} \left({}^G \mathbf{p}_j- {}^G \mathbf{p}_k \right) \rfloor
\end{align}

 }

\end{document}